\documentclass[lettersize,journal]{IEEEtran}
\usepackage{times}
\usepackage{amsmath}
\usepackage{amssymb}
\usepackage{amsfonts}
\usepackage{array}
\usepackage{epstopdf}
\usepackage{graphicx}
\usepackage{caption}
\usepackage{booktabs}
\usepackage{setspace}

\usepackage{xcolor, colortbl}
\usepackage{comment}
\usepackage[en-US]{datetime2}
\usepackage{caption}
\usepackage{svg}
\usepackage{commath}
\usepackage{algorithm}
\usepackage[noend]{algpseudocode}
\usepackage{mathtools}
\usepackage{enumerate}
\usepackage{array}
\usepackage{ltablex} 
\usepackage{threeparttable}
\usepackage{multirow}
\usepackage{makecell}
\usepackage{balance}
\usepackage{cite}
\usepackage{pifont}
\usepackage{hyperref}

\usepackage[labelformat=simple]{subfig}
\usepackage{subcaption}

\newcommand{\xmark}{\ding{55}}

\begin{document}
\title{Leveraging Self-Supervised Learning for Scene Classification in Child Sexual Abuse Imagery}
\author{IEEE Publication Technology Department
\thanks{}}

\author{
Pedro~H.~V.~Valois,
João Macedo,
Leo S. F. Ribeiro,
Jefersson A. dos~Santos,
and~Sandra~Avila
\thanks{
P.~H.~V.~Valois, L.~S.~F.~Ribeiro, and S.~Avila are with the Instituto de Computação, Universidade Estadual de Campinas (UNICAMP), Campinas, São Paulo, Brazil. 
J.~Macedo is with the Departamento de Ciência da Computação, Universidade Federal de Minas Gerais (UFMG), and with Departamento de Polícia Federal (DPF), Belo Horizonte, Minas Gerais, Brazil. 
J.~A.~dos~Santos is with the Department of Computer Science, University of Sheffield, United Kingdom. Corresponding author: S. Avila (e-mail: sandra@ic.unicamp.br).
}}

\markboth{}%
{How to Use the IEEEtran \LaTeX \ Templates}

\maketitle

\begin{abstract}
\noindent Crime in the 21st century is split into a virtual and real world. However, the former has become a global menace to people's well-being and security in the latter. The challenges it presents must be faced with unified global cooperation, and we must rely more than ever on automated yet trustworthy tools to combat the ever-growing nature of online offenses. Over 10 million child sexual abuse reports are submitted to the US National Center for Missing \& Exploited Children every year, and over 80\% originate from online sources. Therefore, investigation centers cannot manually process and correctly investigate all imagery. In light of that, reliable automated tools that can securely and efficiently deal with this data are paramount. In this sense, the scene classification task looks for contextual cues in the environment, being able to group and classify child sexual abuse data without requiring to be trained on sensitive material. The scarcity and limitations of working with child sexual abuse images lead to self-supervised learning, a machine-learning methodology that leverages unlabeled data to produce powerful representations that can be more easily transferred to downstream tasks. This work shows that self-supervised deep learning models pre-trained on scene-centric data can reach 71.6\% balanced accuracy on our indoor scene classification task and, on average, 2.2 percentage points better performance than a fully supervised version. We cooperate with Brazilian Federal Police experts to evaluate our indoor classification model on actual child abuse material. The results demonstrate a notable discrepancy between the features observed in widely used scene datasets and those depicted on sensitive materials. 
\end{abstract}

\section{Introduction}
With the popularization of the internet, the modern world has been deemed to exist without physical borders, undeniably changing how we obtain information, communicate, work, consume, and socialize, but also providing an environment for misuse and abuse. New vicious forms of crime include spreading malicious software, global networks of abuse imagery, drug trafficking, illegal organ supply chains, and even artificial intelligence (AI) enabled blackmail. As crime develops and changes in the age of connected societies, so must policing, for ``the world is becoming a single jurisdiction'' \cite{jahankhani2020policing} and efforts to fight crime in the 21st century must be made internationally. The spread of Child Sexual Abuse Materials (CSAM) is one such crime that has reached alarming proportions in the digital age. In the United States alone, the National Center for Missing and Exploited Children (NCMEC) reports 
a record 32~million reports of suspected CSAM in 2022 in the US\footnote{\url{https://www.missingkids.org/cybertiplinedata}}. 
The availability and sharing of such harmful content online not only exacerbate the trauma inflicted upon the victims but also create a significant burden on law enforcement agents who have to inspect thousands of files, leading to emotional strain when done \text{manually \cite{bursztein2019rethinking, Dalins2018LayingFoundationsEffective}}.

In this paper, we look into the challenges of automating Child Sexual Abuse Imagery (CSAI) detection to present a new under-explored dimension of the problem --- Indoor Scene Classification --- and thoroughly study how state-of-the-art foundation models \cite{Zhou2023ComprehensiveSurveyPretrained} perform under this task, both on publicly available datasets and CSAI directly. With partner law enforcement agents, we design and present new class splits on public scene datasets, allowing for evaluating different methods and hyperparameter tuning without accessing CSAI beyond the final classification test carried out by the~agents.

As expected, access to CSAI is heavily restricted, with possession and handling of such materials forbidden in most of the world. Even in a study in partnership with law enforcement (such as ours), access is restricted even to trained agents. This makes for one of the biggest challenges in the automation of CSAI recognition: researchers have to design methods that can be trained and evaluated with little access to the final test data. This is the main reason why hash-based techniques have been the most common methods used to handle CSAI, with PhotoDNA being the most well-known reference in this category for over a decade \cite{LEE2020301022, PhotoDNA_2011} and Apple's recent NeuralHash \cite{Apple2021CSAMDetectionTechnical} trying to improve upon easily broken hashes with a deep learning solution. However, these methods are only ever helpful in detecting the reappearance of existing CSAI but not new materials being created and shared daily. 

Solutions that use visual clues to detect novel CSAI have not been absent but have focused on two main dimensions of what is expected of these materials from the outside: age estimation and nudity detection \cite{Dalins2018LayingFoundationsEffective, macedo2018benchmark, VITORINO2018303, gangwar2021attm, castrillon2018evaluation, chaves2020improving}. The recent study from \cite{laranjeira2022faact} has found, however, that these are not the only aspects that can be explored without direct data access; they highlight how, in the interviews with law enforcement agents performed by \cite{Kloess2019ChallengesCSAM, Kloess2021Challenges2CSAM} and \cite{Yiallourou2017DetectionImagesContaining}, the illumination, gestures, facial expressions, and scene context are just as valuable to the classification of the material as legal or not.

With this expanded view of the CSAI automation problem and the goal of adding to the tools available to law enforcement, we introduce the first Indoor Scene Classification method designed for use on CSAI; in developing  this task, we yield both a tool for the triage of suspect materials but also a better understanding of how well the knowledge of methods trained with public scene datasets transfer (or not) to CSAI and how this can be improved. The choice of specifically Indoor Scenes comes from talks with partner agents (the crime is seldom recorded in public spaces) and the findings of other studies. For instance, detecting multiple images of a child's room potentially indicates the presence of children's images in a database, even though they are not naked or undergoing any sexually explicit activity \cite{bursztein2019rethinking}. Indoor Scene Classification is also attractive for its feasibility
with reliable, abuse-free material available in large quantities for training. 

Given that this is a novel task, we present a thorough investigation of the performance of foundation models, specifically models trained through self-supervision~\cite{jing2020self}, when fine-tuned on Indoor Scene Classification as a downstream task.

\begin{figure*}[h]
    \centering
    \includegraphics[width=\textwidth]{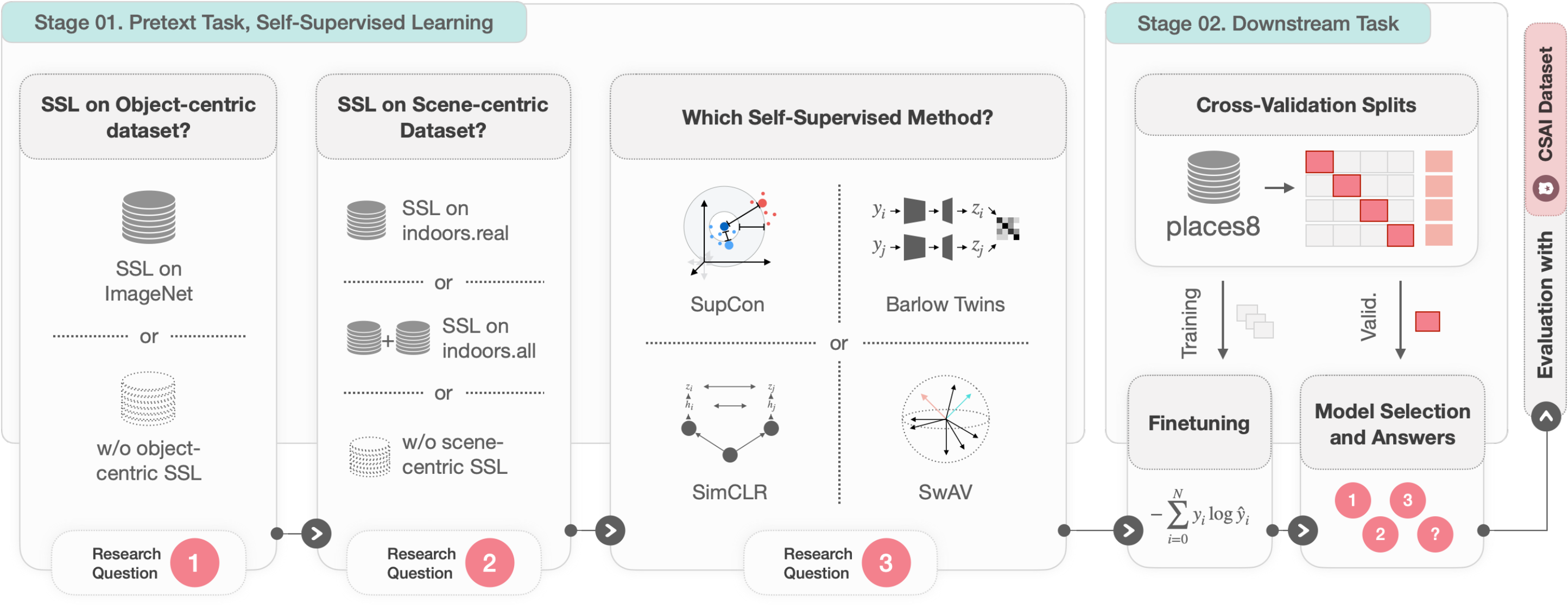}
    \caption{Methodology pipeline. Self-supervised methods are often used in a common two-stage training protocol: (1) the pretext task --- or pre-training stage --- and (2) the downstream task --- or fine-tuning stage. The pretext task uses unlabeled data and runs the SSL technique, while the downstream task uses labeled data from our downstream task. Our downstream stage is fixed, so we consider the pretext task through the lenses of the three research questions depicted (Section~\ref{sec:Methodology}).}
    \label{fig:full-methodology-schema}
\end{figure*}

We aim to study the performance of self-supervised methods on our Indoor Scene Classification task and apply the best model on CSAI following the experimental pipeline at Fig.~\ref{fig:full-methodology-schema}. Our goal is to use the Places8 dataset --- created with law enforcement guidance --- to approximate performance in CSAI. Because sensitive data is restricted in its access, we do not hope to use it for more than one evaluation with a single~model.

We henceforth consider the problem through the lenses of three research questions, the first two about the data used to pre-train the model and the last about the method:

\begin{enumerate}[Q1]
    \item \textbf{Does performing SSL on an object-centric dataset help or hurt downstream performance?}~Following the findings \cite{zhou2017places} that showed the combination of training on scene and object-centric datasets was beneficial to both tasks, we consider how SSL on ImageNet (an object-centric dataset) impacts the downstream performance on Places8. This is in addition to our comparison against a supervised baseline model pre-trained on ImageNet and fine-tuned on Places8.
    \item \textbf{How does SSL with scene-centric datasets impact performance?}~We investigate both how combining object and scene-centric training influences performance and whether including synthetic scenes (Indoors.all) improves the downstream performance.
    \item \textbf{Which well-known SSL method performs best on our task?}~We consider four alternative successful SSL formulations: SwAV \cite{swav}, SimCLR \cite{simclr}, Barlow Twins~\cite{zbontar2021barlow}, and SupCon \cite{scl}.
\end{enumerate}

With these questions and the experiments that will answer each in mind, our analysis can be summarized by the model variants yielded from the combination of the choices enlisted through the questions. When comparing these models, we use Bayesian estimation --- replacing the common t-test --- for ranking and contrasting. Bayesian Estimation Supersedes the T-test (BEST) \cite{kruschke2013bayesian} offers a nuanced and comprehensive view of experimental results compared to traditional t-tests. In our particular case, using 5-fold cross-validation breaks the independence assumption of the traditional t-test, leading to unreliable results. Also, we measure the advantage of using SSL over supervised techniques, which is accurately done using Bayesian~methods.

The task of self-supervision --- learning without labels --- is driven by the desire for more general representations that can be easily fine-tuned for downstream tasks unrelated to the original dataset, a goal well aligned with our CSAI classification needs. We do find, however, that because most self-supervised models are trained on object-centric datasets (namely ImageNet \cite{deng2009imagenet}), knowledge is not as easily transferable to a scene-centric task and that there are benefits to be had from doing self-supervision with scenes in addition to objects. Our set of contributions is then multi-faceted, from the introduction of indoor scene classification to its application on CSAI data:

\begin{enumerate}
    \item We introduce the novel task of Indoor Scene Classification for CSAI by tailoring data from classic datasets on Scene Classification to yield models useful for CSAI without explicitly training on CSAI data;
    \item Our thorough evaluation of self-supervised models shows both their usefulness for Indoor Scene Classification but also how further self-supervision on actual indoor scenes improves their downstream performance;
    \item We present the first evaluation of an Indoor Scene Classification model on CSA imagery and the first discussion on the domain gap between these images and public indoor scene photographs.
\end{enumerate}

The remaining sections are organized as follows. In Sections~\ref{sec:RelatedWork:SSL} and \ref{sec:RelatedWork:CL}, we review self-supervised and contrastive learning frameworks under deep learning. In Sections \ref{sec:RelatedWork:Scenes} and \ref{sec:RelatedWork:CSAM}, we bring the most recent works on scene classification and CSAI investigation. In Section \ref{sec:Methodology}, we describe the datasets, general pipelines used for pretraining, and fine-tuning models used in this work. Then, in Section \ref{sec:Results}, we present the results obtained for our downstream task and real CSAI data. Finally, in Section \ref{sec:Discussion}, we point out the contributions, challenges, and possible routes for future work.

\section{Related Work}
\label{sec:RelatedWork}
\subsection{Self-supervised Learning}
\label{sec:RelatedWork:SSL}

Self-supervised learning (SSL) is a machine-learning approach for building models using unlabeled data. Using unannotated data reduces data annotation costs while finding a purpose to the massive amounts of unlabeled images and texts published daily worldwide \cite{goyal2021self,Ericsson2022SelfSupervisedRepresentationLearning,Gui2023SurveySelfsupervisedLearning}. SSL follows a pattern similar to supervised transfer learning with a pre-training step called pretext task, designed to extract features from unlabeled data. In computer vision, common pretext tasks are grayscale colorization \cite{larsson2017colorization}, jigsaw puzzle, cutout reconstruction, image inpainting, resolution upscaling, foreground object segmentation, clustering, temporal order verification, visual-audio correspondence verification, and contrastive representation comparison.

Afterward, the resulting model is fine-tuned on specific labeled downstream tasks without large amounts of data as a requirement. There are no limitations to the domain of the downstream task. However, it has been empirically shown that ideally the pretext and downstream tasks should comprise images with similar properties, such as textures, shapes, and color gamut~\cite{Gui2023SurveySelfsupervisedLearning,ericsson2021well}. It has been demonstrated that self-supervised models outperform their supervised counterparts for ResNet-50 on image classification, detection, and segmentation in multiple benchmarks \cite{ericsson2021well}. Therefore, SSL shows real advantages over supervised learning, given its higher robustness and generally better performance, especially for CSAI data, which has less than a hundred thousand images in the dataset~\cite{cole2022whenscl}.

\subsection{Contrastive Learning}
\label{sec:RelatedWork:CL}

Regular self-supervision cannot use multiple images to solve one task simultaneously, adding further complexity to model pre-training and leading to limited generalization capacity.

Contrastive learning, on the other hand, was thus developed as a new paradigm for self-supervised pre-training~\cite{jaiswal2021survey}. It is closely related to metric learning, an unsupervised learning sub-field that aims to construct a specialized distance metric from weakly supervised data \cite{bellet2013survey}.

In the end-to-end contrastive learning framework, two encoders are trained simultaneously with contrastive loss. This method focuses on creating a model that, ideally, compares two samples numerically (e.g., two images), therefore grouping similar images while pushing different ones apart. Within an unlabeled dataset, there is no prior distance measurement. Thus, any augmented version is labeled as ``similar'' (or positive) for an anchor image sample, while all other samples are labeled as ``different'' (or negative). The first methods to propose this kind of instance discrimination were IntDisc~\cite{intdisc}, CMC \cite{cmc}, and Deep InfoMax \cite{deepinfomax}. However, the approach by \cite{simclr} was notoriously successful and demonstrated the importance of hard positive samples by introducing a set of 10~data augmentations from which transformations are~sampled.

Despite its state-of-the-art performance on several downstream tasks, the end-to-end training is computationally expensive. It does not always lead to the best downstream performance \cite{ericsson2021well}. Thus, many other approaches were developed, such as MoCo~\cite{moco, mocov2}, SwAV \cite{swav}, and Barlow Twins \cite{zbontar2021barlow} to deal with batch size sensitivity; BYOL \cite{byol} and SimSiam~\cite{simsiam} to overcome the need for negative sampling by changing the instance discrimination pretext task for self-distillation \cite{Gui2023SurveySelfsupervisedLearning}; SupCon~\cite{scl} to make use of labels in SSL when those are available.

\subsection{Scene Classification}
\label{sec:RelatedWork:Scenes}

Humans analyze their environment and important events efficiently, but machines have long stayed caught up in this topic. Scenes are understood as semantically coherent views of real-world environments containing objects, textures, and a spatially separated background \cite{henderson1999highsceneperception, epstein2005cortical}. Scene classification is a well-studied field with a particular interest in surveillance, autonomous driving, and robotics navigation, but it is still lacking in performance \cite{quattoni2009recognizing}.

Scene classification is closely related to object and texture classification but with multiple objects, textures, backgrounds, and intricate relations between them. Thus, one could imagine it as multiple local classifications happening simultaneously to provide a context and a set of items or actions~\cite{zeng2021scenesurvey}. However, handcrafting such solution is unfeasible. Then, current approaches~\cite{goyal2021self} use deep learning and scene representation to solve this problem.

Neural Networks used for scene classification are typically convolutional nets (CNN) with popular architectures such as VGG or ResNet. However, CNNs are not usable in this field without being trained on large amounts of labeled data~\cite{zeng2021scenesurvey}, meaning most works start by transferring learning \cite{ericsson2021well,jaiswal2021survey} from models trained on ImageNet \cite{deng2009imagenet} or Places~\cite{zhou2017places}. However, transfer learning is usually limited by the similarity between the source and target domains, meaning that ImageNet pre-trained models are, on average, less effective than Places pre-trained versions~\cite{zhou2014learning}. 

Additionally, \cite{qiu2021scene} used graph nets and image inpainting to show that a scene constitutes minor and essential objects. Only the essential ones matter for the final scene classification, while minor elements can be removed without impact. However, the number of minor objects is usually much higher and can be regarded as noise.

After selecting a pre-trained model, most works develop specialized fine-tuning routines to extract the most scene-centric features in the downstream dataset~\cite{zeng2021scenesurvey}. GAP-CNN replaces fully connected layers for global average pooling (GAP) to focus on class-specific regions while using fewer parameters~\cite{zhou2016learning}; Hierarchical LSTM uses four Long-Short Term Memory (LSTM) modules in an attempt to capture spatial dependencies \cite{zuo2016learning}; DL-CNN proposes dictionary learning (DL) layers to obtain sparse representations and reduce the total number of parameters; finally, the Spatially Unstructured layer was designed to help cope with layout deformations, and scale changes in scene classification \cite{hayat2016spatial}. Fisher Vector (FV-CNN) \cite{cimpoi2015deep} and Mixture of Factor Analyzers Fisher Vector (MFAFVNet) \cite{li2017deep} use Improved FV while Multi-scale Orderless Pooling (MOP-CNN) \cite{gong2014multi} and Semantic Descriptor with Objectness (SDO) \cite{cheng2018scene} employ a Vector of Locally Aggregated Descriptors (VLAD) to cluster local features. However, such an approach heavily increases the number of images to be~processed. 

Moreover, many studies \cite{durand2016weldon, laranjeira2019modeling, chen2020scene, zhao2018volcano} used object detectors \cite{liu2020objectdetcsurvey, girshick2015fastrcnn, liu2016ssd, redmon2016yolo, redmon2017yolo9000} to help with identification of essential scene regions and force the model to pay attention to particular objects during classification. Typically, this dependency introduces two-stage training \cite{xie2020scene}, meaning that learning has to be split between the object detector and classifier. 

Several scene classification methods have used multiple layers' outputs to improve scene representation in this context. The high CNN layers are too compact for the dense level features required for scene classification, keeping only large objects, while the low CNN layers still keep the small ones~\cite{wu2015harvesting}. Thus, harnessing multi-layer features reaches overall better representations, but careful feature fusion design is required~\cite{liu2019novel}. Ultimately, this is the most structurally complicated approach, and the fused features generate high-dimensional arrays, making the model overfit easily the more layers are used \cite{yang2015multi}.

To the best of our knowledge, few works have used SSL as means to solve scene classification. In scene semantic segmentation, \cite{mccormac2016scenenetrgbd} showed an improved model on NYUv2 and SUN RGB-D benchmarks using a model pre-trained on a 5M image 3D rendered dataset called SceneNet RGB-D. In order to tackle the PASCAL VOC benchmark (11,530 images containing 27,450 ROI annotated objects and 6,929 segmentations), Ren and Lee \cite{scenenetssl} proposed a multi-task learning approach that pre-trains an AlexNet model to predict edges, surface normals, and depth maps from SceneNet \text{RGB-D} images and a discriminator to distinguish the synthetic images from the Places365 images. Similarly, She and Xu~\cite{she2021contrastive} added a SimCLR contrastive loss as a new pretext task, setting as positive pairs all viewpoints of a single scene. The aforementioned methods reached comparable performance or even outperformed other SSL techniques using the same neural network architecture but underperformed compared to the state-of-the-art method for the tested benchmarks.

\subsection{CSAI Classification}
\label{sec:RelatedWork:CSAM}

Easy access and impunity led to a frenzied growth in the consumption and distribution of CSAI over the past few years, with hundreds of millions of images circulating through the web today \cite{bursztein2019rethinking}. The automation of CSAI recognition can protect adults and children, more easily blocking that sort of content in social media and thus preventing trauma \cite{LEE2020301022}, but this is still a challenging field of research. Such datasets must stay accessible exclusively to law enforcement personnel, dramatically increasing the difficulties in comparing possible models or performing benchmark evaluations \cite{laranjeira2022faact}. NCMEC shows that hash comparison tools are already the most significant source for reports today, but law enforcement personnel are the exclusive mechanism responsible for auditing CSAI when dealing with never-before-seen data, which is not only the source of bias but also the psychological strain on the agents' mental health \cite{macedo2018benchmark}.

In light of these issues, researchers have attempted to design more scalable and reliable methodologies to tackle CSAI recognition. Image hash databases, web crawlers, and file metadata are most commonly used to detect CSAI sources and find criminals \cite{panchenko2012detection, shupo2006toward, guerra2021detecting}. In a more versatile method, NuDetective \cite{de2010nudetective} and iCOP \cite{peersman2014icop} make use of handcrafted nudity image descriptors to build fast CSAI detectors, and \cite{sae2014towards} used textures and facial distances to distinguish adult from child nudity. However, these solutions are not robust against simple modifications, demanding constant updates to stay functional.

With the recent developments in computer vision detectors, machine learning algorithms also started to be used for CSAI recognition. Ground-up training and transfer learning of CNNs were shown to outperform current forensic and commercial tools easily and can be made into portable tools for search and seizure procedures \cite{VITORINO2018303}. Moreover, \cite{gangwar2021attm} demonstrated that pornography and age-group recognition could be leveraged for CSAI with substantial gains in accuracy for binary classification, and \cite{macedo2018benchmark} used a single-model estimation of child presence, age, and gender to improve the performance. \cite{rondeau2019deep} arguments apparent age and nudity detection could also be leveraged for CSAI, while \cite{castrillon2018evaluation} focused on non-adult people detection in images for CSAI age estimation, reaching over 90\% accuracy on their proposed dataset. When regarding video materials (CSA Videos), studies are fewer and rely on extracting features from individual frames (images) using pre-trained networks and detecting snippets with nudity~\cite{peersman2014icop,Borg2022DetectingRankingPornographic} and extracting facial features to perform victim and abuser identification \cite{Westlake2022DevelopingAutomatedMethods, Brewer2023AdvancingChildSexual}; Even thought their method was not tested directly on CSAI, \cite{Borg2022DetectingRankingPornographic} used the former approach and took it one step further, incorporating object detection for body parts to rank videos with nudity by ``severity''/``harmfulness''.

All these studies firmly outline that deep learning can contribute to child abuse inquiries. However, most published works focus on detecting children and nudity/pornography \cite{perez2017video,ishikawa2019combating,avila2019multimodal} and highlight the difficulties of finding large annotated CSAI databases to experiment with, even with the help of law enforcement agents. Several works focus exclusively on age estimation within CSAI, and children's images are collected to build datasets that help accomplish such tasks. For instance, \cite{anda2020deepuage} proposed VisAGe and \cite{castrillon2018evaluation} AgeMega, databases focused on age estimation with thousands of underage and adult facial images, while \cite{gangwar2021attm, castrillon2018evaluation, chaves2020improving} gather images from several age estimation databases. Beyond that, \cite{al2020evaluating} and \cite{tabone2021pornographic} included reproductive organ detection to aid CSAI detection, while~\cite{yiallourou2017detection} used facial expressions to distinguish sensitive from non-sensitive material, which is one of the main challenges in the field. \cite{LEE2020301022} call attention to the fact that ``the best results can be achieved if multiple methods are used in combination'', meaning that CSAI is a complex enough problem that age, nudity, and pornography detectors do not fulfill all the roles for possible feature extractors that can aid on this task. 

Even with these efforts, CSAI classification presents several challenges before accurate predictions can be made automatically. As \cite{french2020csam} explains, the decision of whether an image is explicit enough to be considered indecent is not well defined: the notion of ``sexually explicit'' may sound straightforward but becomes extremely subtle more often than not, given the context of the image or the culture of the viewers. \cite{kloess2019challenges} argue that ``nudity alone is not indicative of indecency'' whereas posing fully clothed can be considered erotic under some circumstances, mainly if other images depicting the same child in more abusive material are found during the investigation. 

Moreover, \cite{laranjeira2022faact} designed an aggregated analysis pipeline to aid CSAI recognition research by showing a general pipeline for statistical analysis of features, such as scenes, skin tone, and nudity, making use of the region-based annotated child pornography dataset (RCPD)~\cite{macedo2018benchmark}, without publicizing information on individual samples. RCPD is a private database used internally by the Brazilian Federal Police and contains 2,138 images among CSAI, pornography, and non-sensitive categories.

In this context, most databases used for CSAI recognition could suffer from biased distributions, not representing what is found in the real world. \cite{laranjeira2022faact} show that CSA data shared online differs from reported cases of physical abuse in Brazil in terms of race: ``most reported victims of child sexual abuse in Brazil are black and brown girls from 8 to 14 years old, but RCPD depicts mostly white children being abused and most of the material apprehended in Brazil has different tendencies than reports of sexual abuse with physical contact''.

Overall, CSAI classification could take advantage of non-direct CSAI recognition, and \cite{bursztein2019rethinking} emphasize that the investigation should optimally consider scene information as means to cluster similar imagery and ``report objects that are present, its environment, and potentially identify landmarks that will help locate the region where an abusive image originated''. Experts use scene information to gather context and understand what is happening in the image or video, especially in low-resolution scenarios and when people are covered or with their backs turned \cite{kloess2019challenges}. CSAI distribution is a massive problem worldwide, and, for the sake of the children under abuse, combating it demands new data-driven scalable solutions.

\section{Methodology}
\label{sec:Methodology}
In this section, we formalize the addressed tasks, the datasets used, and the sub-sets designed; finally, we present the experimental protocol followed throughout this study.

\subsection{Problem Definition}
\label{sec:problem}

Any study with CSAI must consider the ethical concerns surrounding such data. Therefore, it is common to always design methods under a protocol not dissimilar to the Domain Generalization (DG) task. 

In DG, a \textit{domain} is defined as a probability distribution from which data is sampled; for a classification problem, we have then the joint distribution $\mathcal{P}_{XY}$ with samples from $X$ and labels from $Y$. Within the DG task, a collection of $M$ domains with samples $\mathcal{S}^{train} = \{S_i = \{x^{(i)}, y^{(i)}\}\}_{i=1}^{i=M}$ is available for training while another set of $N$ domains with samples $\mathcal{S}^{test} = \{S_j = \{x^{(j)}, y^{(j)}\}\}_{j=1}^{j=N}$ is used only for testing. The $\mathcal{S}^{train}$ set is used to learn a predictive function $h \colon X \mapsto Y$ that should be robust and generalizable enough to work well on the unseen domains $\mathcal{S}^{test}$. It is easy to see how this definition can be directly applied to tasks on CSAI, with the sensitive data automatically composing $\mathcal{S}^{test}$ as a single unseen domain due to constrained access.

The remaining challenge is then to define which $\mathcal{S}^{train}$ domains can be used to train such a robust model, as well as the training procedure itself. As presented previously, our chosen classification task is Indoor Scene Classification. Whereas scene classification may be acknowledged as a homogeneous classification task \cite{zeng2021scenesurvey}, indoor and outdoor environments vary greatly in textures, objects, colors, shadows, framing, and many other features. Indoor environments contain large non-textured regions --- rare in outdoor environments --- making models that work well on outdoor images fail at indoor scenes \cite{IndoorSfMLearner}. Moreover, the distribution and variety of objects are much denser in indoor environments, making semantic segmentation more expensive and prone to error \cite{scenenetssl}.

To solve Indoor Scene Classification with CSAI in mind, we propose to compare the performance of a classic transfer learning approach of fine-tuning a network pre-trained on ImageNet --- our baseline --- against a selection of self-supervised models known for generalizing to a plethora of downstream tasks --- in some instances surpassing their supervised counterparts~\cite{swav}. By comparing the results on our Indoor Scene Classification benchmark, we are able to select the best-performing model for this task, the code, and learned weights, which are then sent to law enforcement partners to do the final assessment on real~CSAI.

\subsection{Datasets}
\label{sec:dataset}

As mentioned in Section~\ref{sec:RelatedWork:SSL}, SSL is trained in two stages; the self-supervised stage trains a model using a pretext task, a pseudo-labeled classification and representation learning task meant to instill general representations into the model. This pre-trained model is then fine-tuned on a downstream task and evaluated. For each of these stages, (i)~pretext, (ii)~downstream, and (iii)~evaluation, we have designed new sub-sets of existing public datasets, leaving (iv)~the final CSAI evaluation and related dataset in the hands of partner law enforcement agents. Creating new sub-sets was necessary because most scene classification datasets with enough samples comprise both outdoor and indoor classes, and we wish to filter out the former. We present the selection for each stage next.

\paragraph*{Indoors.all \& Indoors.real}

We propose two combined datasets for the pretext self-supervised task: Indoors.all and Indoors.real. The former comprises all chosen indoor images from Hypersim~\cite{hypersim2021}, OpenRooms~\cite{li2021openrooms}, InteriorNet~\cite{InteriorNet18}, and Places365-Challenge~\cite{zhou2017places}, while the latter is then simply the Places365-Challenge indoor images. With this in mind, Indoors.all is called so as it contains not only real but also synthetic images and their rendered views (depth, segmentation, etc.). These views are used as pseudo-labels in the SSL pretext task, which we believe is more informative than random augmentations and could potentially produce better models, inspired by \cite{scenenetssl}. Table~\ref{tab:datasets-stats} summarizes the included datasets and their sizes.

\begin{table}[!ht]
\small
\centering
\caption{Datasets used for the self-supervised pretext task. Size stands for the total number of images, and Synth stands for synthetic datasets.  $^\dag$Combinations of the above public~datasets.}
\begin{tabular}{lrc}
\toprule
{Dataset} & {Size} & {Synth} \\
\toprule
Places365 \cite{zhou2017places} & 8,496,949 & \xmark \\
InteriorNet \cite{InteriorNet18} & 4,000,000 & \checkmark \\
Hypersim \cite{hypersim2021}    & 1,432,480 & \checkmark \\
OpenRooms \cite{li2021openrooms}   & 118,233  & \checkmark \\
Indoors.all (ours)$^\dag$  & 7,778,152 & \checkmark \\
Indoors.real (ours)$^\dag$ & 2,490,632 & \xmark \\
\bottomrule
\end{tabular}
\label{tab:datasets-stats}
\end{table}

While most SSL models are readily available pre-trained on the well-known ImageNet dataset, we corroborate the finding from \cite{zhou2017places} that shows that models trained on object-centric datasets do not perform well on scene-centric tasks and vice-versa and even that training on both distributions may be beneficial. This comparison is further discussed with the experimental results in Section~\ref{sec:Results}.

\paragraph*{Places8}

We introduce a new subset of Places --- called Places8 --- where classes are selected to highlight environments most common in CSAI. This is a smaller dataset than the ones used for the pretext task; it represents our downstream task and is used for fine-tuning the model post self-supervised learning.

Places365-Challenge indoor classes were initially grouped from 159 to 62 new categories following WordNet synonyms and sometimes direct hyponyms or related words. For example, \textit{bedroom} and \textit{bedchamber} were joined, while \textit{child room} was kept in a separate category given its importance in CSAI investigation. Next, we filtered the remapped dataset into 8~final classes from 23 different scenes of Places365-Challenge. The selection of such scenes followed conversations with the partner Brazilian Federal Police agents and CSAI investigation and labeling experts. Places365-Challenge already provides training and validation splits mapped accordingly. The test split was then generated from a stratified 10\% split from the training set, given that the remapping and filtering made for a highly imbalanced dataset. The complete remapping can be seen in Table~\ref{tab:places8} under ``Original Categories'' and further details for the novel~sub-set. 

\begin{table}[!ht]
\centering
\caption{Description of the Places8 dataset. The class represents the final label used, while the original categories stand for the original Places365 labels. Places365 already provides training and validation splits mapped accordingly. The test set comes from a stratified 10\% split from the training~set.}
\small
\begin{tabular}{>{\raggedright\arraybackslash}p{0.11\linewidth}rrrr>{\raggedright\arraybackslash}p{0.255\linewidth}}
\toprule
\multirow{2}{*}{Class} & \multirow{2}{*}{Test} & \multirow{2}{*}{Train} & \multirow{2}{*}{Val} & \multirow{2}{*}{\%} & Original Categories               \\ \toprule
bathroom      & 5,740  & 51,655  & 200        & 13.4  & bathroom, shower                  \\
bedroom     & 11,112 & 100,012 & 600 & 25.9 & bedchamber, bedroom, hotel room, berth, dorm room, youth hostel           \\
child's room  & 4,650  & 41,849  & 300        & 10.8  & child's room, nursery, playroom    \\
classroom     & 3,751  & 33,763  & 200        & 8.7   & classroom, kindergarden classroom \\
dressing room & 2,432  & 21,889  & 200        & 5.7   & closet, dressing room             \\
living room & 9,940  & 89,458  & 500 & 28.7 & home theater, living room, recreation room, television room, waiting room \\
studio        & 1,404  & 12,633  & 100        & 3.3   & television studio                 \\
swimming pool & 1,505  & 13,547  & 200        & 3.5   & jacuzzi, swimming pool            \\ \midrule
Total         & 40,534 & 364,806 & 2300       & 100 & \\
\bottomrule
\end{tabular}
\label{tab:places8}
\end{table}

We emphasize, however, that one class mentioned by the practitioners was not found within Places365-Challenge: photographic studio. Thus, we selected the ``television studio'' class of Places365, as it depicts people posing among photo and video cameras.

\paragraph*{Out-of-Distribution (OOD) Scenes}
\label{sec:Custom-test-dataset}

While the introduced Places8 already comprises a test set, we sought to create an additional evaluation set to understand better our approach's limitations when exposed to a domain gap. This is especially necessary when we consider that CSAI is known to come from diverse demographics and social backgrounds \cite{laranjeira2022faact}.

Thus, we designed a small ``custom dataset'' from online images to check if the model performance is outside of the controlled nature of Places8.  The dataset comprises 80~images, 10~images per class from the 8 Places8 classes: bathroom, bedroom, child's room, classroom, dressing room, living room, studio, and swimming pool.

The OOD Scenes set is a sample of images taken~from Google images, Bing images, and the Dollar Street data\-set~\cite{rojasdollar} in a 4:3:3 ratio. All images are free to share, modify, and use, including Dollar Street, licensed under CC-BY 4.0 \text{Commercial}. 

Dollar Street is an annotated image dataset of 289 everyday household items photographed from 404 homes in 63~countries worldwide. It contains 38,479 pictures, split among abstractions (image answers for abstract questions), objects, and places within a home. This dataset explicitly depicts underrepresented populations and is grouped by country and income. Not all countries are present, but there is a balanced amount of pictures per region, and most images come from families who live with less than USD \$1000 per month \cite{rojasdollar}. 

These sources were chosen to contrast the Places dataset data from the web with underrepresented data. For reproducibility, Places8 and OOD Scenes are available at Zenodo \url{https://doi.org/10.5281/zenodo.13910526}, upon request.

\section{Model Selection}
\label{sec:selection}
In this section, we go through the model selection process. We recall that experiments that directly use CSAI are limited by law enforcement agents' availability and local hardware requirements and are generally discouraged. For this reason, we perform model selection with data that approximates the problem, hence our creation of Places8 in partnership with agents (Section~\ref{sec:dataset}). The following experiments then sought to best answer the three questions posed in Section~\ref{sec:Methodology} by using \text{5-fold} cross-validation and $3\times$ experiment repetitions, comparing model variants on Places8 while avoiding spurious results from training randomization.

\subsection{Experimental Setup}
\label{sec:experimental_setup}

We conducted all experiments using 6 NVIDIA RTX GPUs A6000 with 48 GB each. We used PyTorch 1.10\footnote{\url{https://pytorch.org}} as the deep learning framework with the Vision library for state-of-the-art self-supervised learning (VISSL)\footnote{\url{https://vissl.ai}}. ResNet-50 \cite{resnet} was chosen as the backbone for all experiments. ImageNet pre-trained models are used directly from public ``model zoos'' while scene-centric SSL pre-trained models are trained from scratch or ImageNet-based weights; all models are later fine-tuned on Places8. The training times were optimized through mixed precision and activation checkpointing --- allowing training to progress at 1.5 hours/epoch. The batch size was set to 1024 samples per GPU ($1024\times 6$). The LARS optimizer was used with learning rate scheduler of cosine half wave 3$\times$ restarts given the high batch size we chose, which follows the same training hyperparameters and guidelines from VISSL~\cite{goyal2022vision}. SSL pre-training was performed in 50 epochs, given our training budget, while fine-tuning was done on 100 epochs for Places8 with early stopping supported by the validation~loss.



\subsection{SSL on Object-centric Data}

To answer our first question (\textbf{Q1}), we investigate the influence of starting SSL with the object-centric dataset ImageNet. Previous studies, including most SSL methods \cite{swav, simclr, zbontar2021barlow, scl}, have shown ImageNet data to be an excellent starting point for any downstream task, and the creators of the Places dataset \cite{zhou2017places} complemented that combining scene and object-centric data for pre-training may be beneficial for both tasks. 

Because training within an SSL protocol is expensive (Section~\ref{sec:experimental_setup}), we sought to limit our exploration of this first question to a single SSL method and observe only the effect of adding object-centric learning to the mix; SwAV was the chosen method for this section for it is the best performing on the ImageNet classification downstream task of the four SSL methods (75.3\% vs.~73.2\% from second place Barlow Twins); we, therefore, study the 5 model variants yielded by questions \textbf{Q1} and \textbf{Q2} to isolate the impact of training first on ImageNet (\textbf{Q1}). The results can be found in Table~\ref{tab:swav-experiments}. This evaluation shows that \textit{significant improvements can be made using the widely available object-centric data to complement training, even when the downstream task is scene-centric}. Comparing all experiments on all folds, we could measure with 97\%~confidence an average improvement of 3.1~percentage points in balanced accuracy on Places8 validation when starting with ImageNet SSL \textit{vs}.~models trained with scene-centric SSL from~scratch.

\begin{table}[h]
\centering
\caption{Comparing the impact of training first on the object-centric ImageNet dataset to answer our first research question (\textbf{Q1}). Results are reported for the SwAV SSL protocol; we report accuracy on the Places8 downstream task of Indoor Scene Classification.}
\begin{tabular}{llc}
\toprule
Object-centric SSL & Scene-centric SSL & Accuracy \\ 
\toprule
ImageNet & w/o & $0.64$ \scriptsize{$\pm$ $0.12$} \\
ImageNet & indoors.real & $0.67$ \scriptsize{$\pm$ $0.12$}\\
ImageNet & indoors.all & $0.66$ \scriptsize{$\pm$ $0.12$} \\ \cmidrule(lr){1-3}
w/o & indoors.real & $0.45$ \scriptsize{$\pm$ $0.25$} \\
w/o & indoors.all & $0.55$ \scriptsize{$\pm$ $0.21$} \\ 
\bottomrule
\end{tabular}
    \label{tab:swav-experiments}
\end{table}

\subsection{SSL on Scene-centric Data}

Our second question (\textbf{Q2}) then considers not only whether it is beneficial to perform further SSL with scene-centric data or not but also if the inclusion of a large set of synthetic scenes together with real samples improves downstream performance.  With the established answer to \textbf{Q1}, all of our models for the following experiments are trained first with SSL on ImageNet and each variant is run with all four chosen self-supervision protocols (therefore also covering \textbf{Q3}). The result of our full evaluation can be found in Table~\ref{tab:Q2-Q3-experiments}. 

To compare the benefit of including a Scene-centric SSL step in addition to SSL on ImageNet, we use BEST, contrasting the distributions of experiments without Scene-centric SSL and with training on indoors.real; we found that including the extra task yields an average improvement of 6 percentage points in balanced accuracy (with 96.2\% confidence).

When considering the inclusion of synthetic data, we compare SSL training with indoors.real versus indoors.all and found a 3\% advantage to not using additional data (with 89.8\% confidence). We hypothesize that the available synthetic data --- while abundant in the number of samples --- is limited in its diversity, with a large number of views being derived from the same 3D environment.  

Our conclusion for (\textbf{Q2}) is then that \textit{including a scene-centric pretext task had a positive impact} over only fine-tuning from an ImageNet trained object-centric model and that \textit{synthetic data, when lacking diversity, does not contribute to downstream performance on real indoor scene images}.

\begin{table}
\centering
\caption{Comparing the impact of the inclusion scene-centric --- real and synthetic --- datasets for pre-training and the SSL training protocol. Results are reported as accuracy on the downstream Indoor Scene Classification task on the Places8 dataset.}
\begin{tabular}{llc}
\toprule
Protocol & Scene-centric SSL & Accuracy \\ 
\toprule
Supervised & w/o & $0.71$ \scriptsize{$\pm$ $0.04$}  \\ \cmidrule(lr){1-3}
\multirow{3}{*}{Barlow Twins} & w/o & $0.76$ \scriptsize{$\pm$ $0.13$} \\
 & indoors.real & $0.83$ \scriptsize{$\pm$ $0.07$} \\
 & indoors.all & $0.85$ \scriptsize{$\pm$ $0.04$} \\ \cmidrule(lr){1-3}
\multirow{3}{*}{SimCLR} & w/o & $0.54$ \scriptsize{$\pm$ $0.15$} \\
 & indoors.real & $0.62$ \scriptsize{$\pm$ $0.24$} \\
 & indoors.all & $0.49$ \scriptsize{$\pm$ $0.25$} \\ \cmidrule(lr){1-3}
\multirow{3}{*}{SupCon} & w/o & $0.82$ \scriptsize{$\pm$ $0.06$} \\
 & indoors.real & $0.58$ \scriptsize{$\pm$ $0.21$} \\
 & indoors.all & $0.53$ \scriptsize{$\pm$ $0.24$} \\ \cmidrule(lr){1-3}
\multirow{3}{*}{SwAV} & w/o & $0.64$ \scriptsize{$\pm$ $0.12$} \\
 & indoors.real & $0.67$ \scriptsize{$\pm$ $0.12$} \\
 & indoors.all & $0.66$ \scriptsize{$\pm$ $0.12$} \\ 
\bottomrule
\end{tabular}
    \label{tab:Q2-Q3-experiments}
\end{table}

\subsection{The Self-Supervised Learning Protocol}

\begin{figure}[h]
    \centering
    \includegraphics[width=0.4\textwidth]{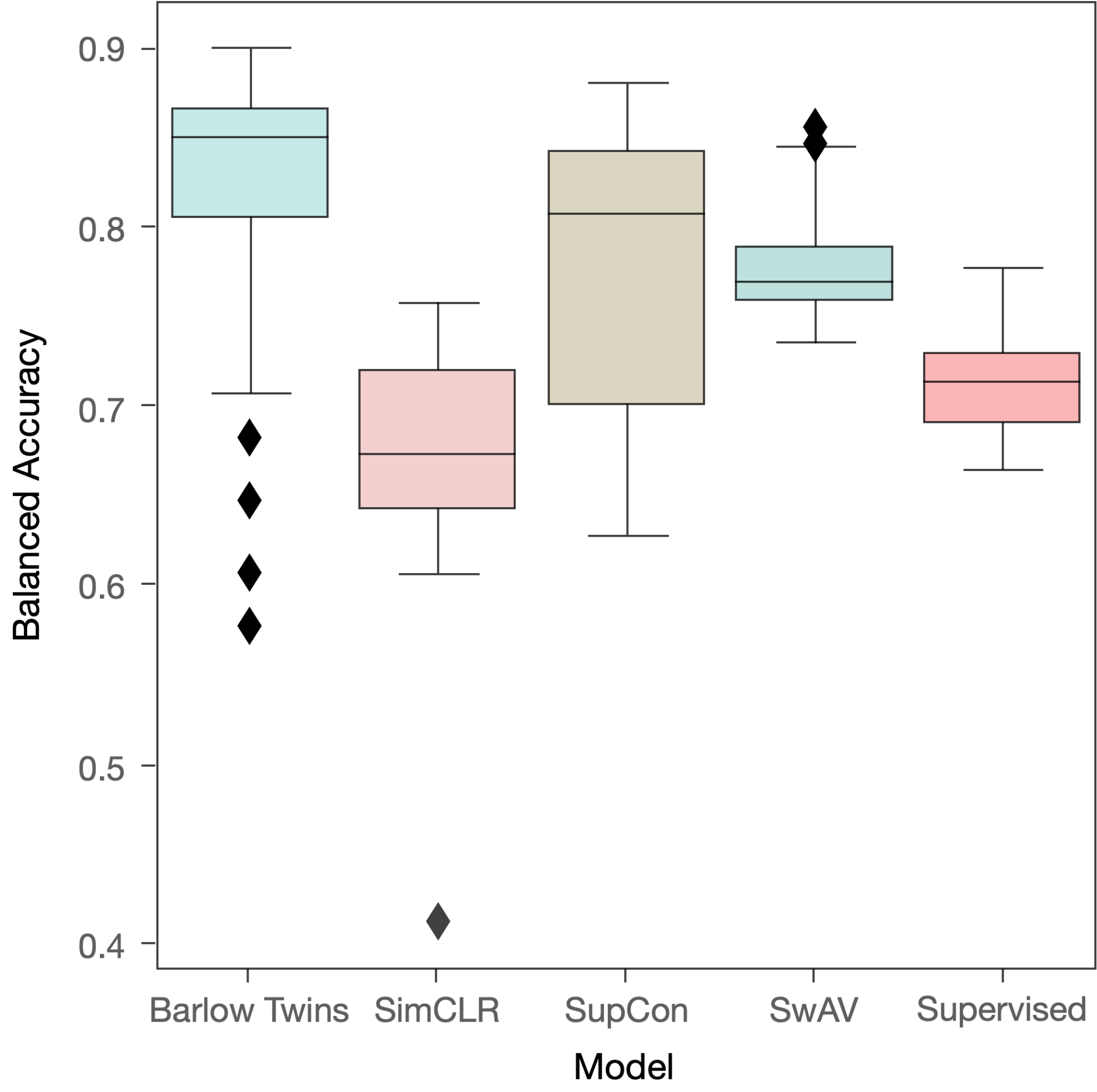}
    \caption{Box-plots of balanced accuracy versus the different techniques (SSL and supervised) fine-tuned on Places8.}
    \label{fig:general-tech-comparison}
\end{figure}

With all decisions made regarding training phases and the data used, we finally consider which SSL protocol performs best on our Indoor Scene Classification downstream task. The results were already presented in Table~\ref{tab:Q2-Q3-experiments} but we now perform the comparison across techniques (instead of training sets). From the table and the box-plot in Fig.~\ref{fig:general-tech-comparison}, it is possible to see how both Barlow Twins and SwAV outperform the common Supervised pre-training and fine-tuning protocol, with the former having a stronger advantage. By comparing with BEST~\cite{pymc2022best}, we have found that Barlow Twins outperforms the Supervised method with an average improvement of 2.2\% (98\% confidence) in balanced accuracy. Our answer to (\textbf{Q3}) is then that \textit{Barlow Twins performs best for our downstream task, and SSL improves on the purely supervised result}.

Our selected model is then comprised of the answers, following the Barlow Twins SSL protocol with the pipeline of object-centric SSL with ImageNet, scene-centric SSL with indoors.real, and fine-tuning on the Places8 downstream task. The final model was trained end-to-end with the full training set of Places8 (no folds) and evaluated on its test set. The results showed then 71.6\% balanced accuracy. This is higher than what is commonly shown for the full Places dataset \cite{zeng2021scenesurvey}, but with fewer classes, it is impossible to do a fair comparison. We hope future studies consider using Places8, given its relationship to the CSAI final task.

    
    

\begin{figure*}[h]
\captionsetup[subfigure]{labelformat=empty,skip=0pt}
\captionsetup[subfloat]{captionskip=0pt}
    \centering
    {\centering \rotatebox[origin=lb]{90}{\scriptsize{~~~bathroom}}}
    \subfloat[\scriptsize{\checkmark}]{\includegraphics[width=1.6cm,height=1.6cm]{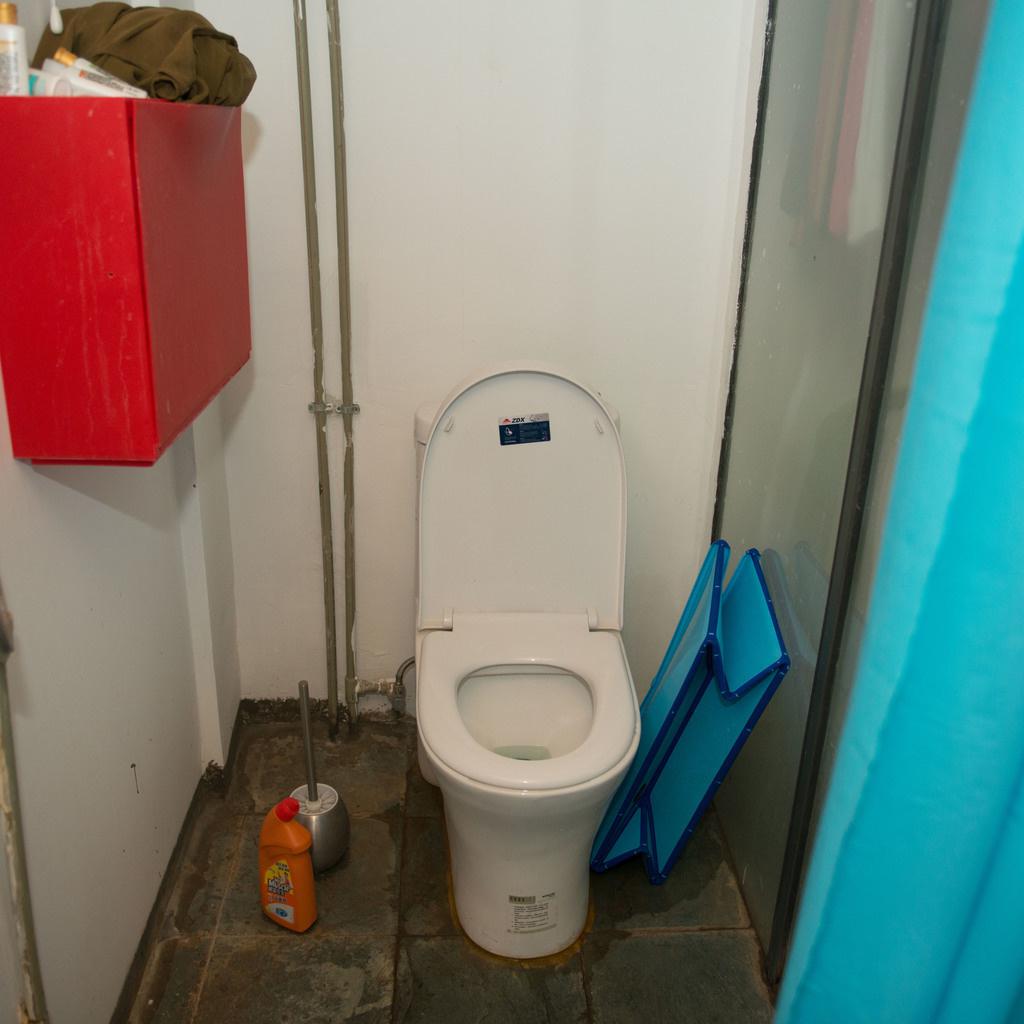}}\hspace{0.01cm}
    \subfloat[\scriptsize{\checkmark}]{\includegraphics[width=1.6cm,height=1.6cm]{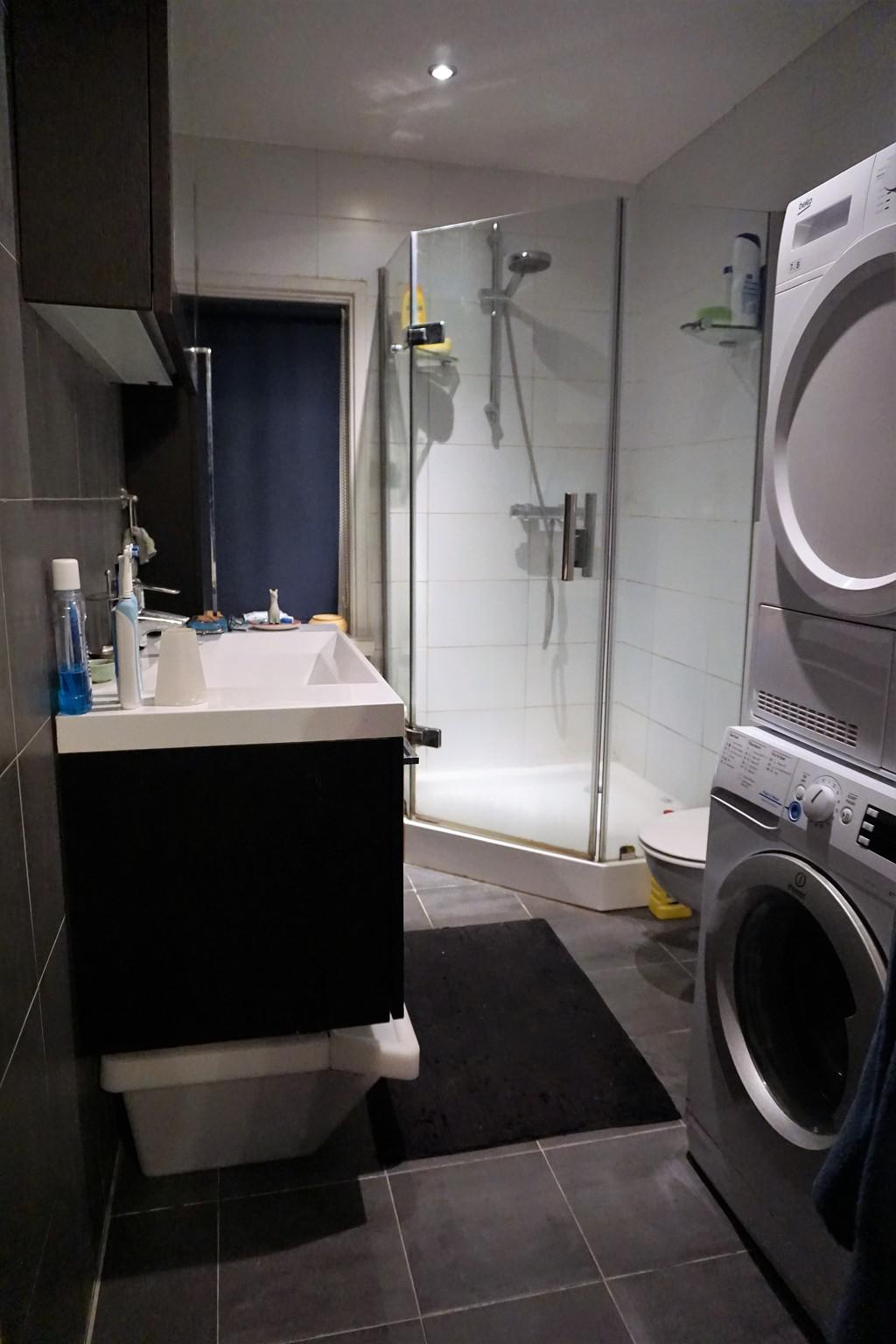}}\hspace{0.01cm}
    \subfloat[\scriptsize{\checkmark}]{\includegraphics[width=1.6cm,height=1.6cm]{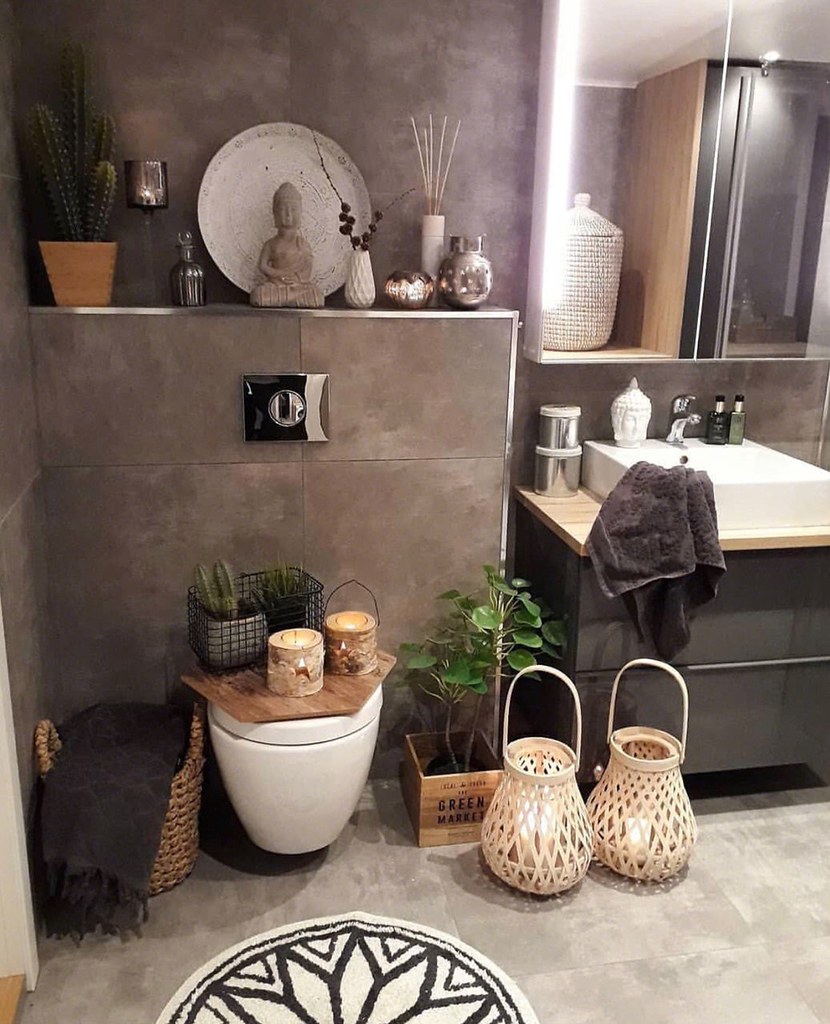}}\hspace{0.01cm}
    \subfloat[\scriptsize{\checkmark}]{\includegraphics[width=1.6cm,height=1.6cm]{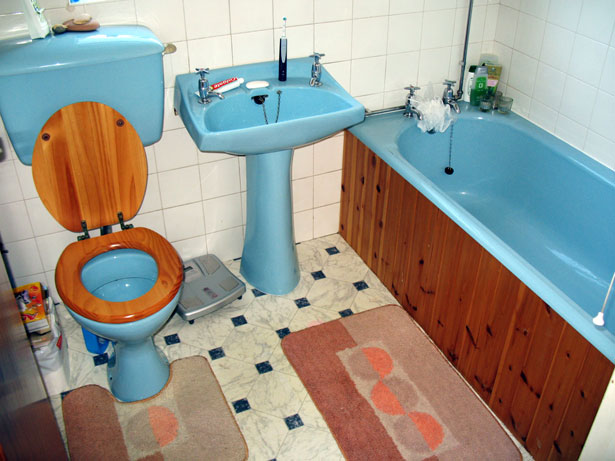}}\hspace{0.01cm}
    \subfloat[\scriptsize{\checkmark}]{\includegraphics[width=1.6cm,height=1.6cm]{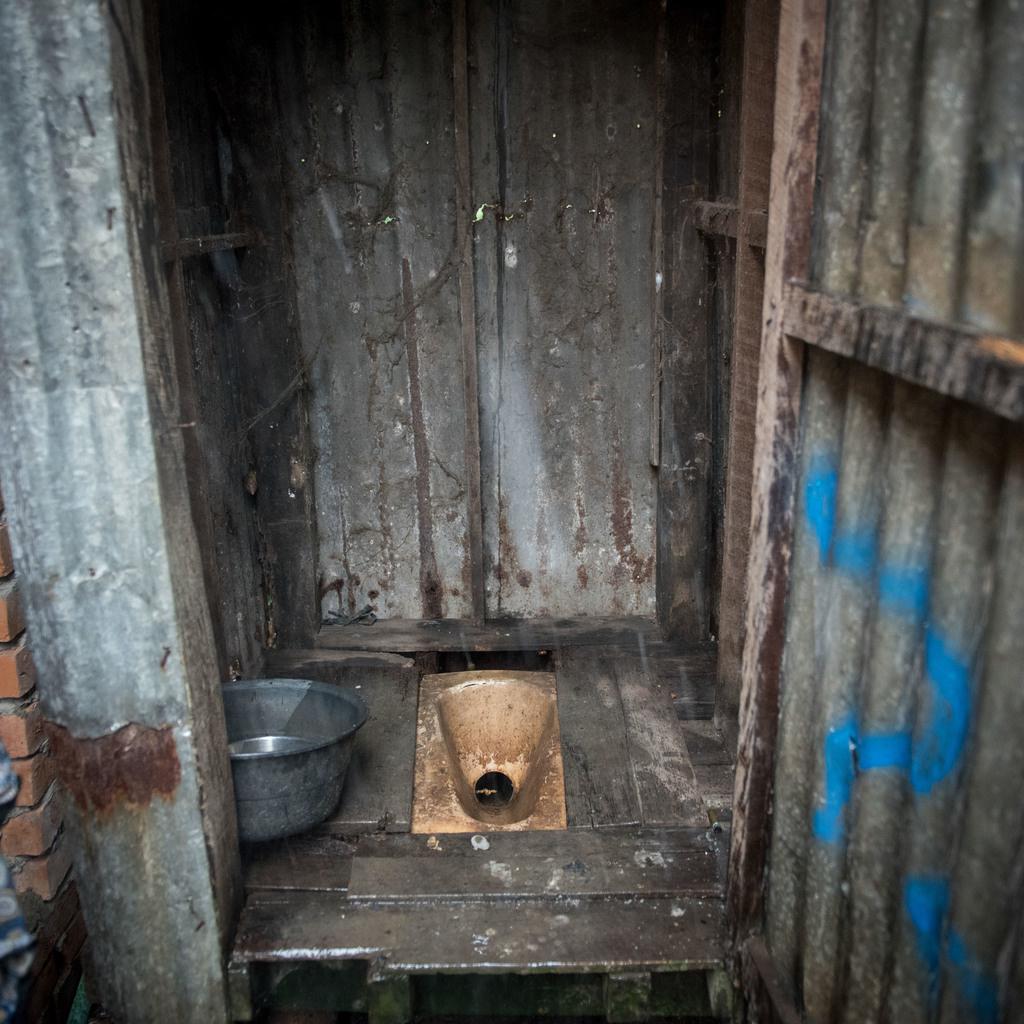}}\hspace{0.01cm}
    \subfloat[\scriptsize{\checkmark}]{\includegraphics[width=1.6cm,height=1.6cm]{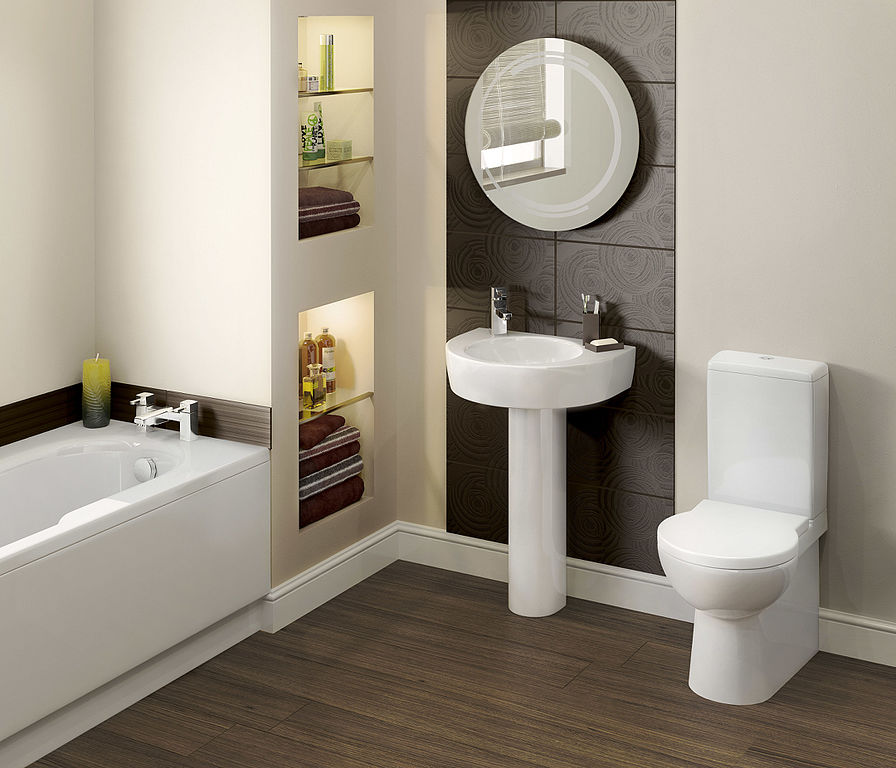}}\hspace{0.01cm}
    \subfloat[\scriptsize{\checkmark}]{\includegraphics[width=1.6cm,height=1.6cm]{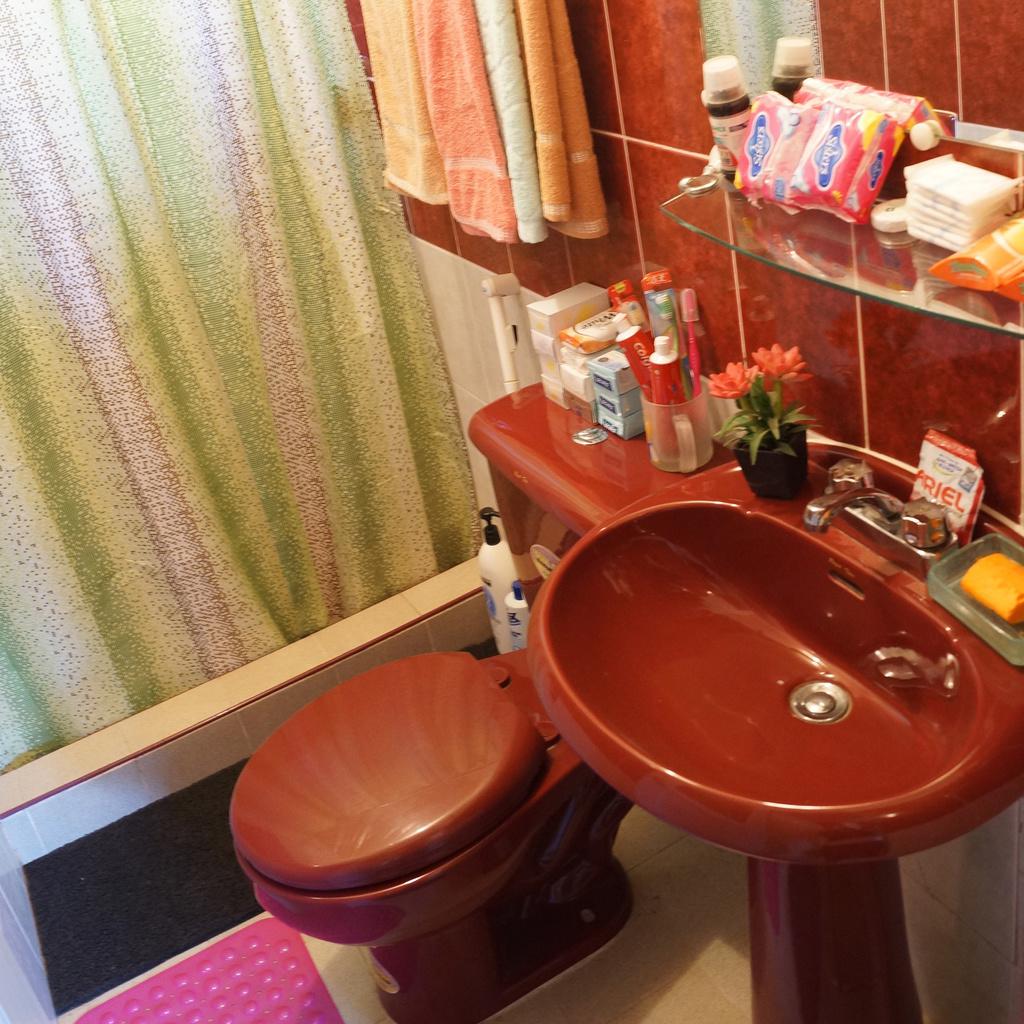}}\hspace{0.01cm}
    \subfloat[\scriptsize{\checkmark}]{\includegraphics[width=1.6cm,height=1.6cm]{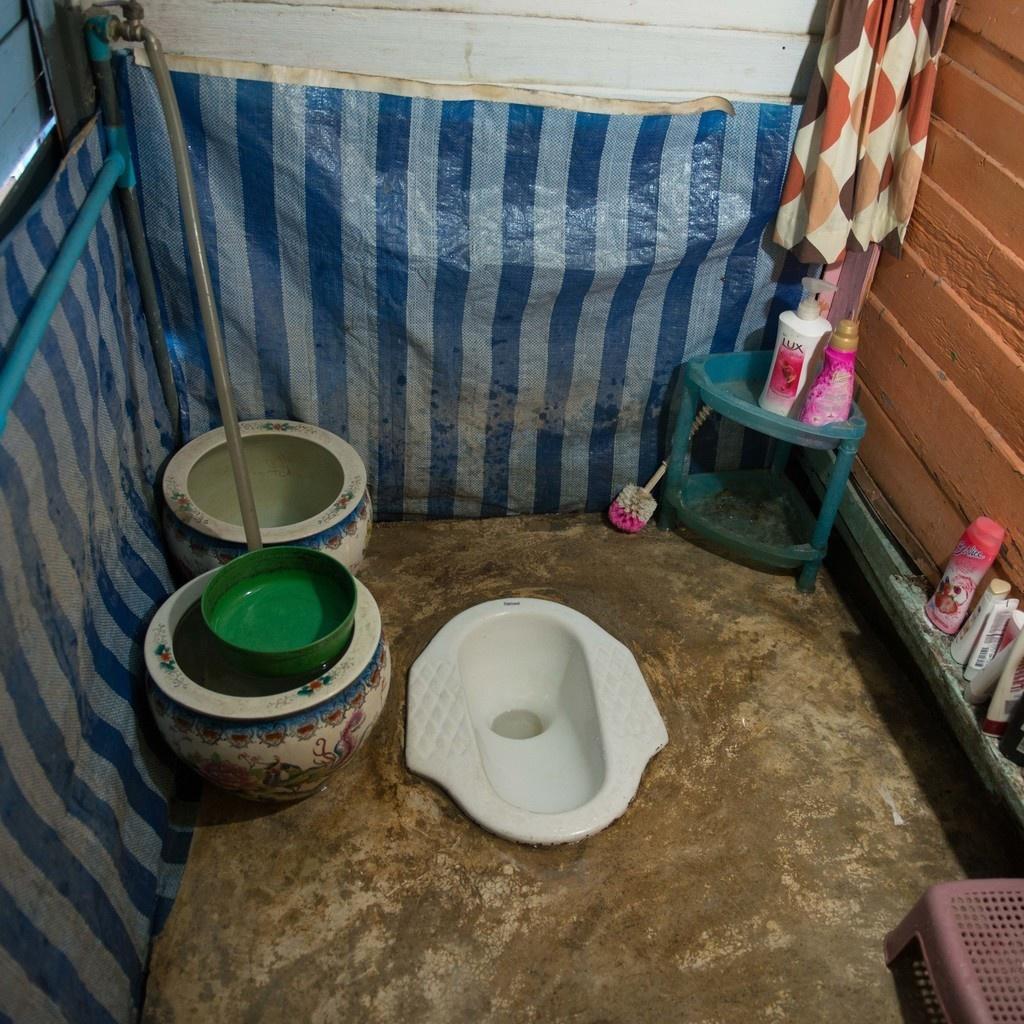}}\hspace{0.01cm}
    \subfloat[\scriptsize{\checkmark}]{\includegraphics[width=1.6cm,height=1.6cm]{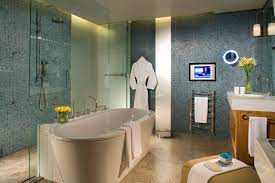}}\hspace{0.01cm}
    \subfloat[\scriptsize{\checkmark}]{\includegraphics[width=1.6cm,height=1.6cm]{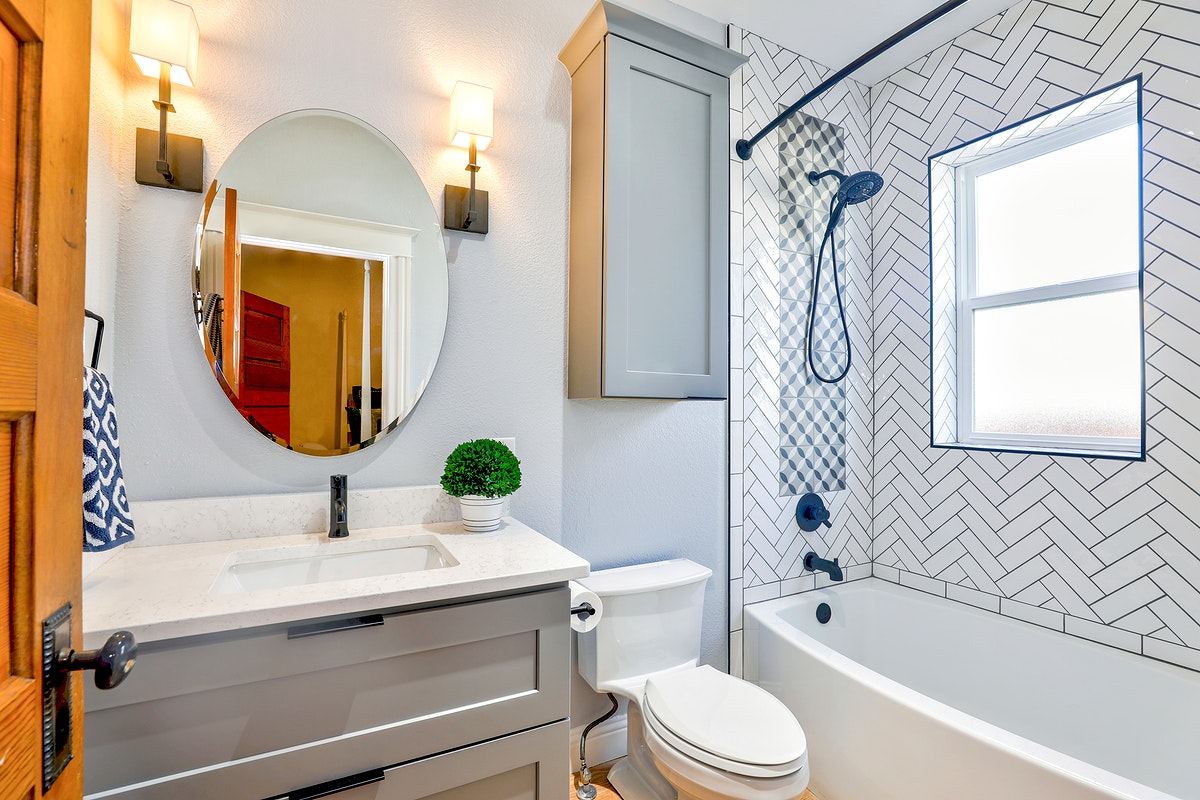}}\\ \vspace{-0.2cm}

    {\centering \rotatebox[origin=lb]{90}{\scriptsize{~~~~bedroom}}}
    \subfloat[\scriptsize{\checkmark}]{\includegraphics[width=1.6cm,height=1.6cm]{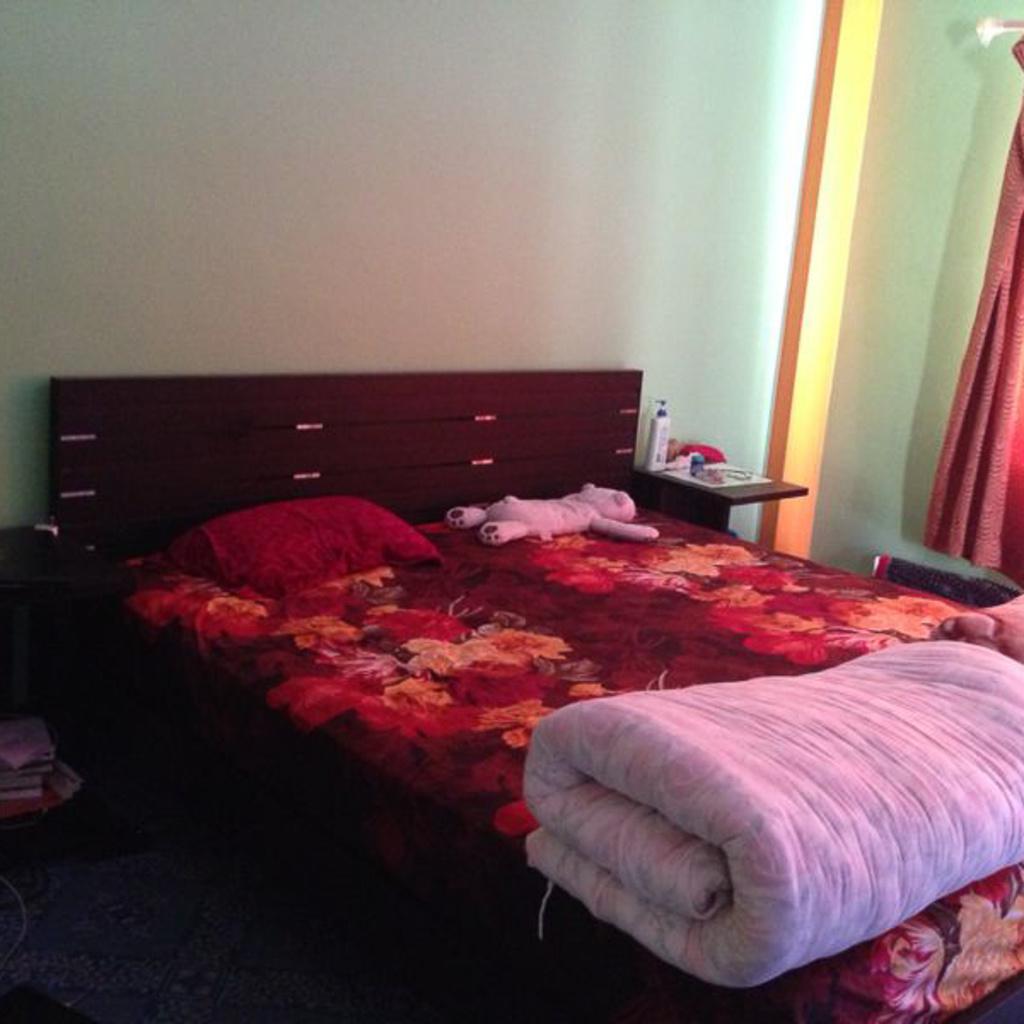}}\hspace{0.01cm}
    \subfloat[\scriptsize{\checkmark}]{\includegraphics[width=1.6cm,height=1.6cm]{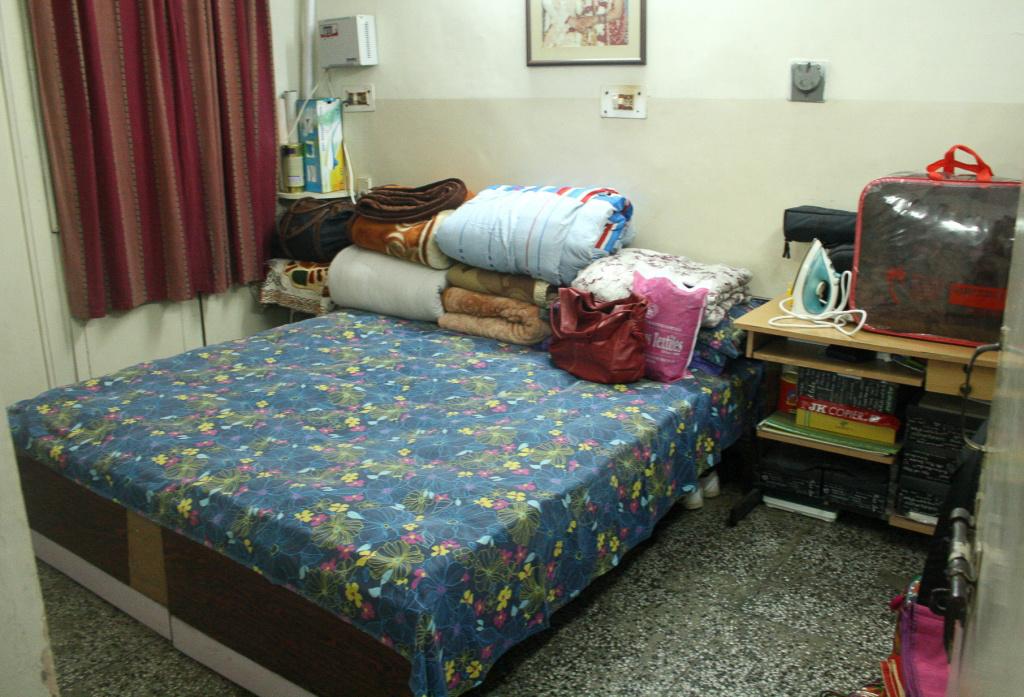}}\hspace{0.01cm}
    \subfloat[\scriptsize{\checkmark}]{\includegraphics[width=1.6cm,height=1.6cm]{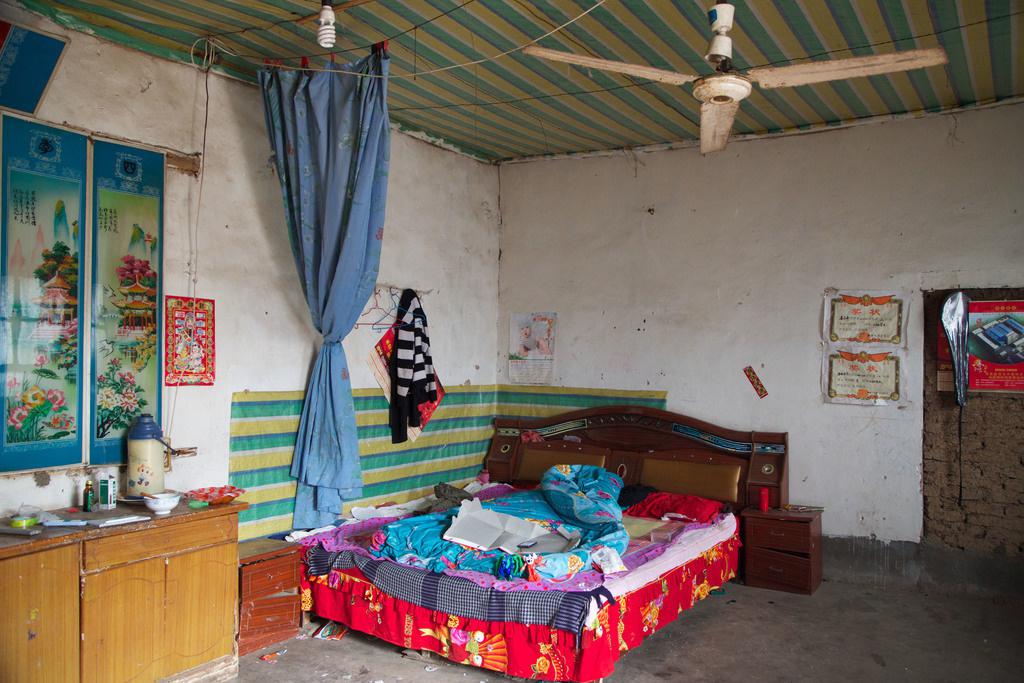}}\hspace{0.01cm}
    \subfloat[\scriptsize{\checkmark}]{\includegraphics[width=1.6cm,height=1.6cm]{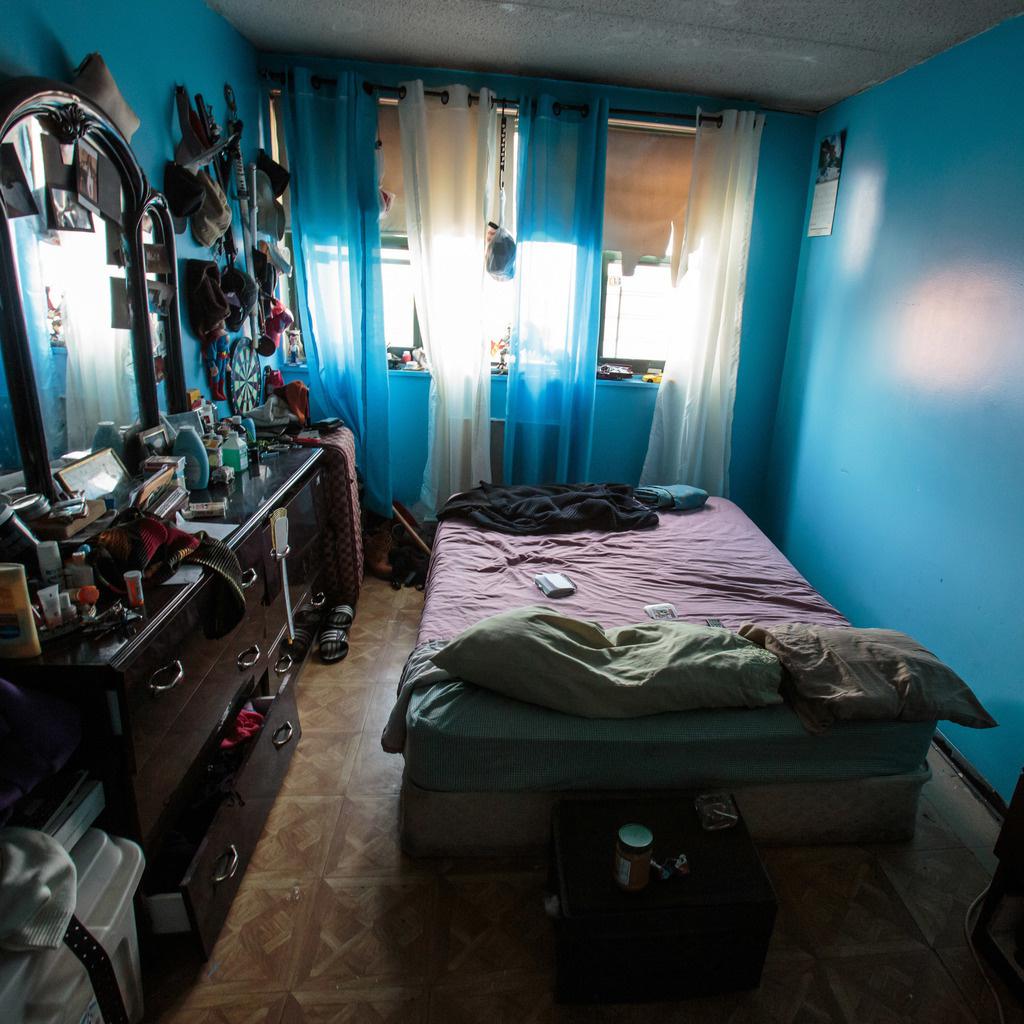}}\hspace{0.01cm}
    \subfloat[\scriptsize{dressing room}]{\includegraphics[width=1.6cm,height=1.6cm]{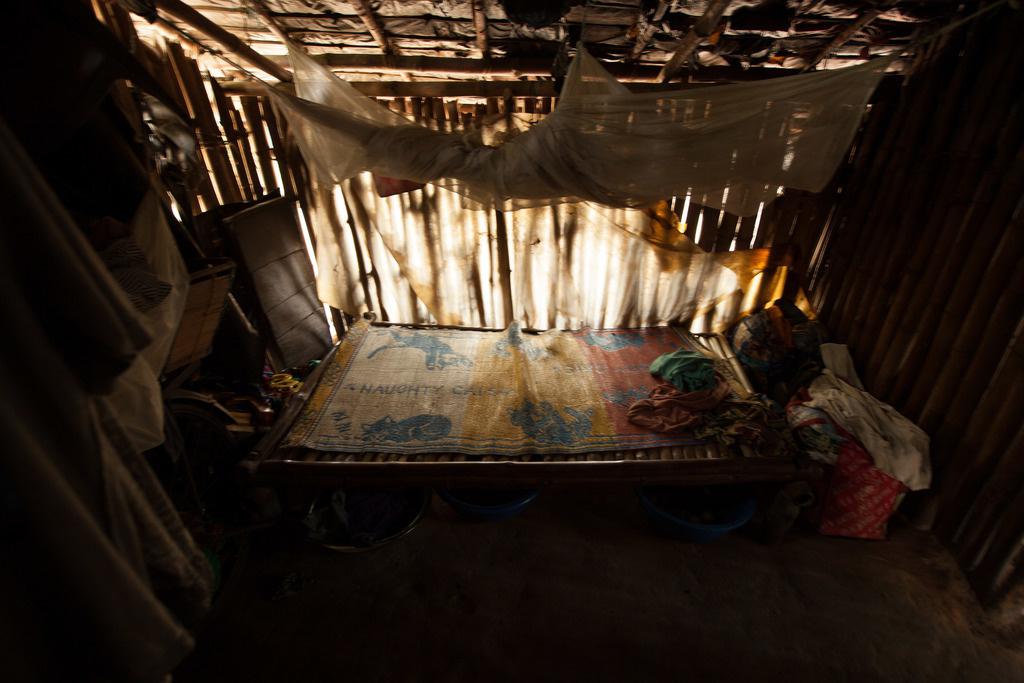}}\hspace{0.01cm}
    \subfloat[\scriptsize{\checkmark}]{\includegraphics[width=1.6cm,height=1.6cm]{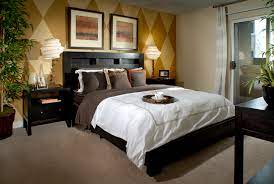}}\hspace{0.01cm}
    \subfloat[\scriptsize{\checkmark}]{\includegraphics[width=1.6cm,height=1.6cm]{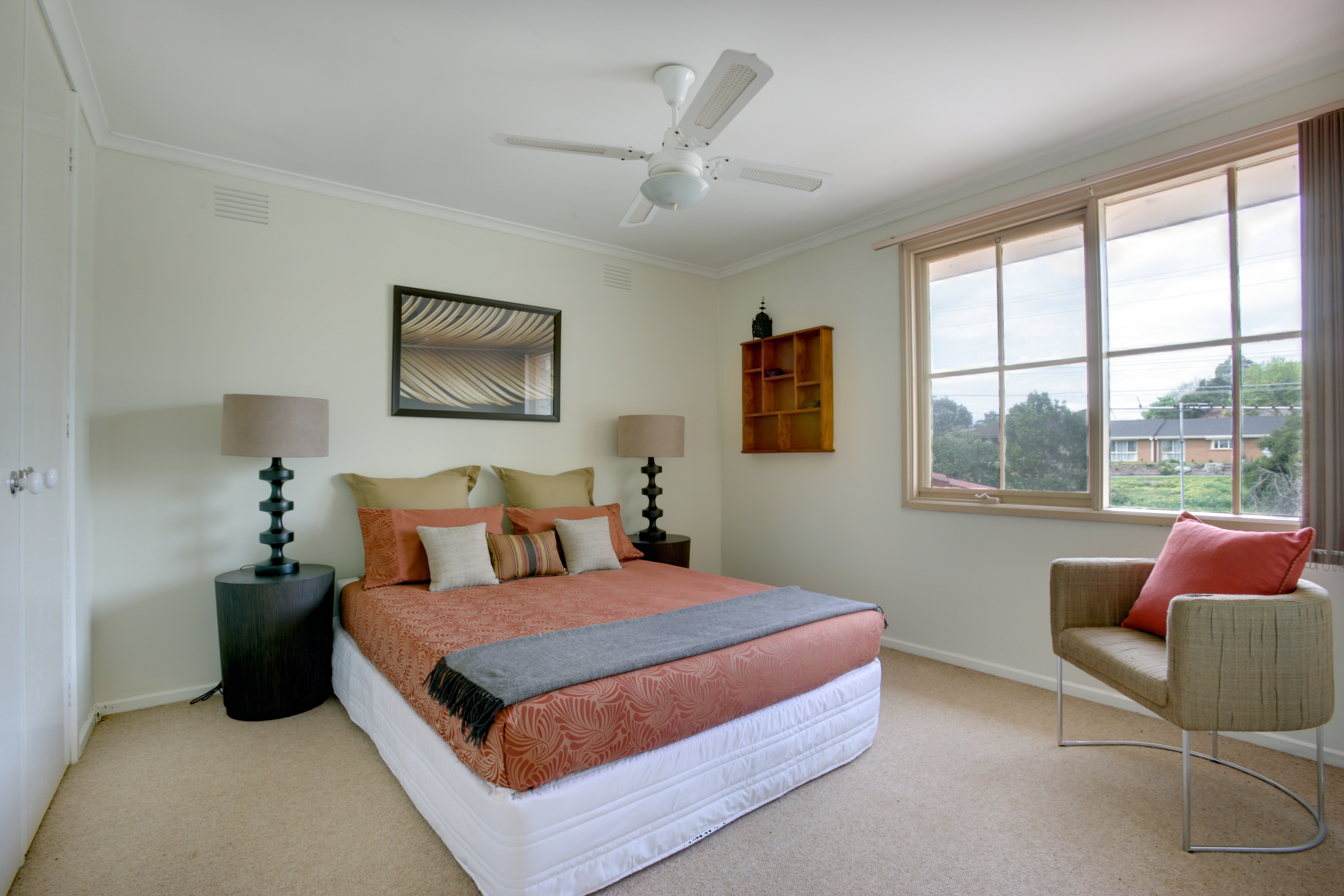}}\hspace{0.01cm}
    \subfloat[\scriptsize{\checkmark}]{\includegraphics[width=1.6cm,height=1.6cm]{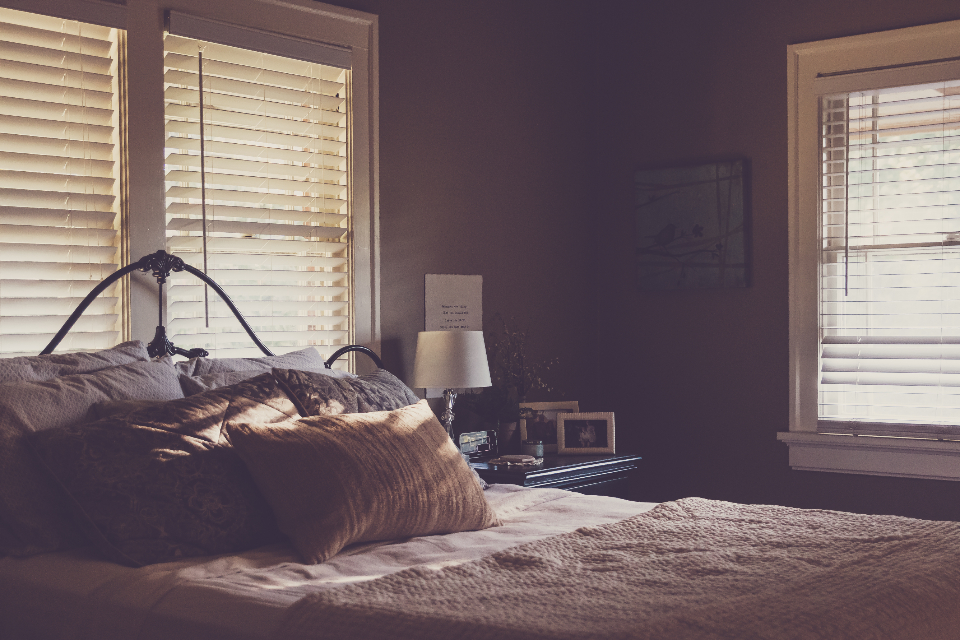}}\hspace{0.01cm}
    \subfloat[\scriptsize{\checkmark}]{\includegraphics[width=1.6cm,height=1.6cm]{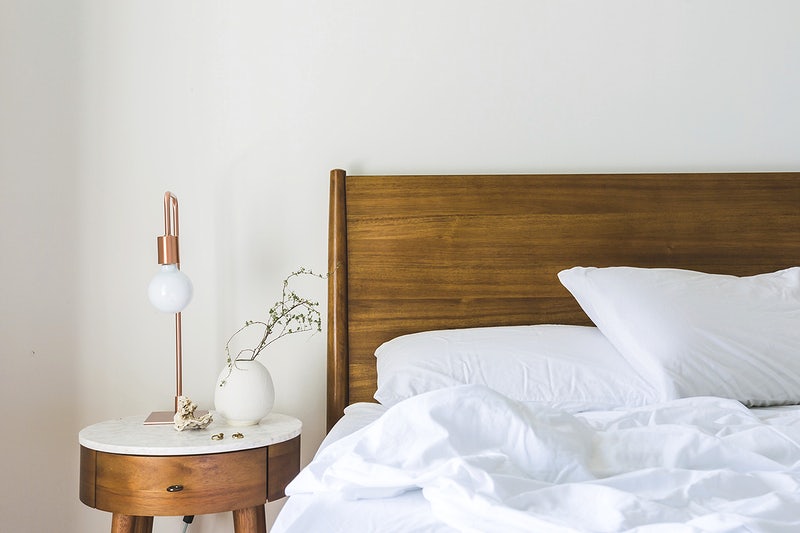}}\hspace{0.01cm}
    \subfloat[\scriptsize{\checkmark}]{\includegraphics[width=1.6cm,height=1.6cm]{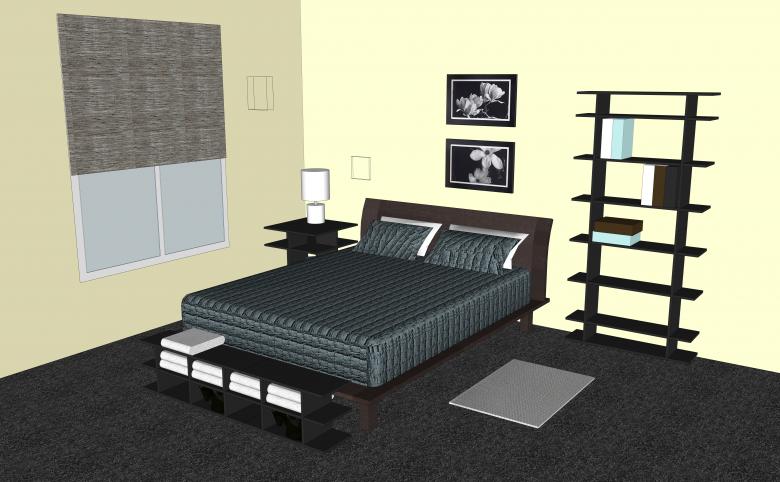}}\\ \vspace{-0.2cm}

    {\centering \rotatebox[origin=lb]{90}{\scriptsize{~~child's room}}}
    \subfloat[\scriptsize{dressing room}]{\includegraphics[width=1.6cm,height=1.6cm]{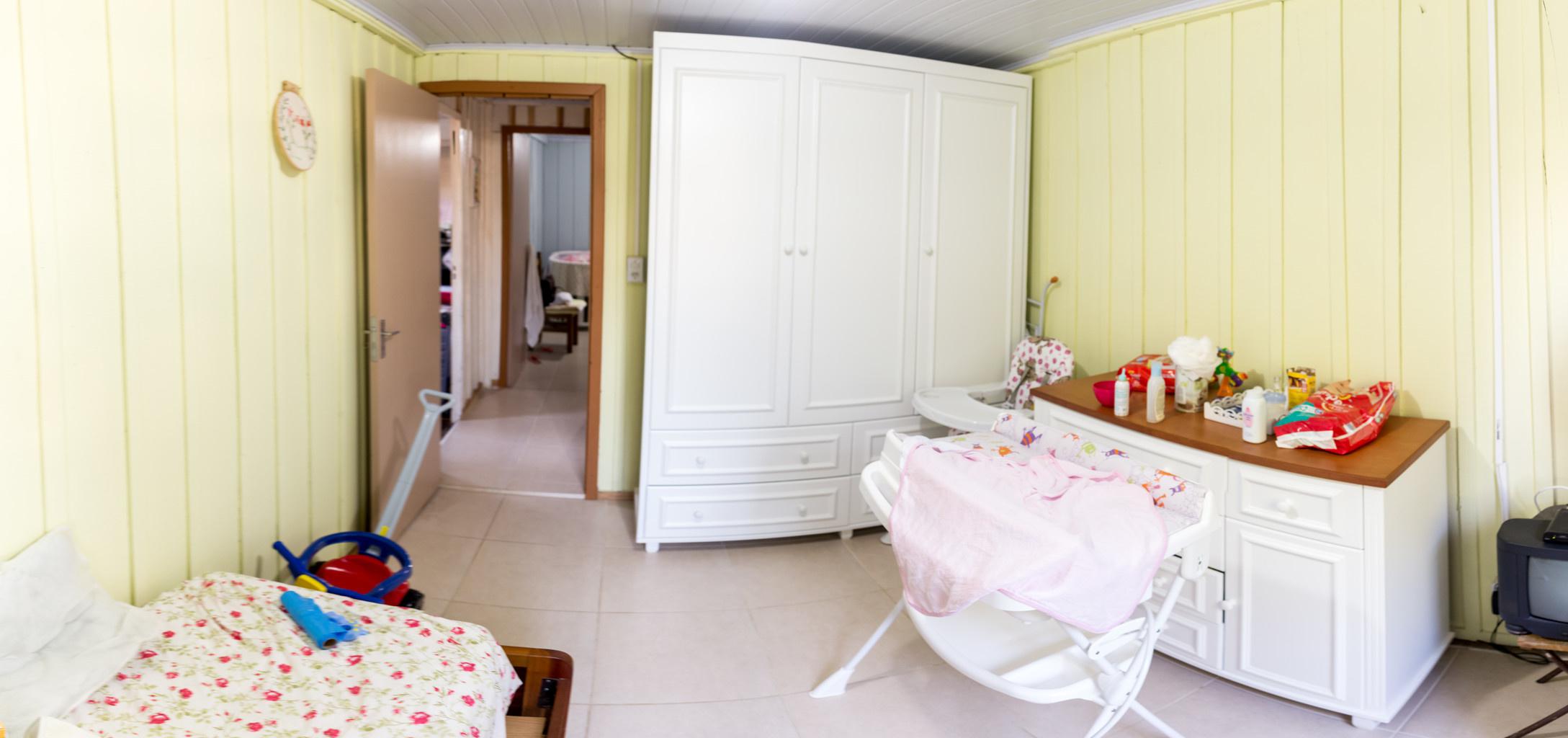}}\hspace{0.01cm}
    \subfloat[\scriptsize{bedroom}]{\includegraphics[width=1.6cm,height=1.6cm]{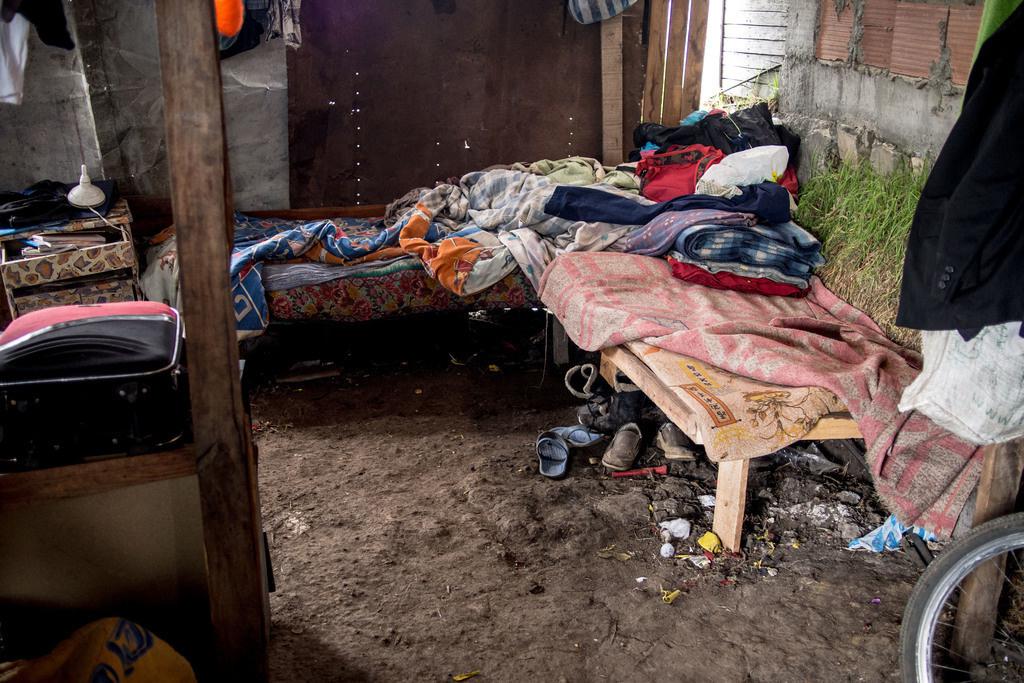}}\hspace{0.01cm}
    \subfloat[\scriptsize{bedroom}]{\includegraphics[width=1.6cm,height=1.6cm]{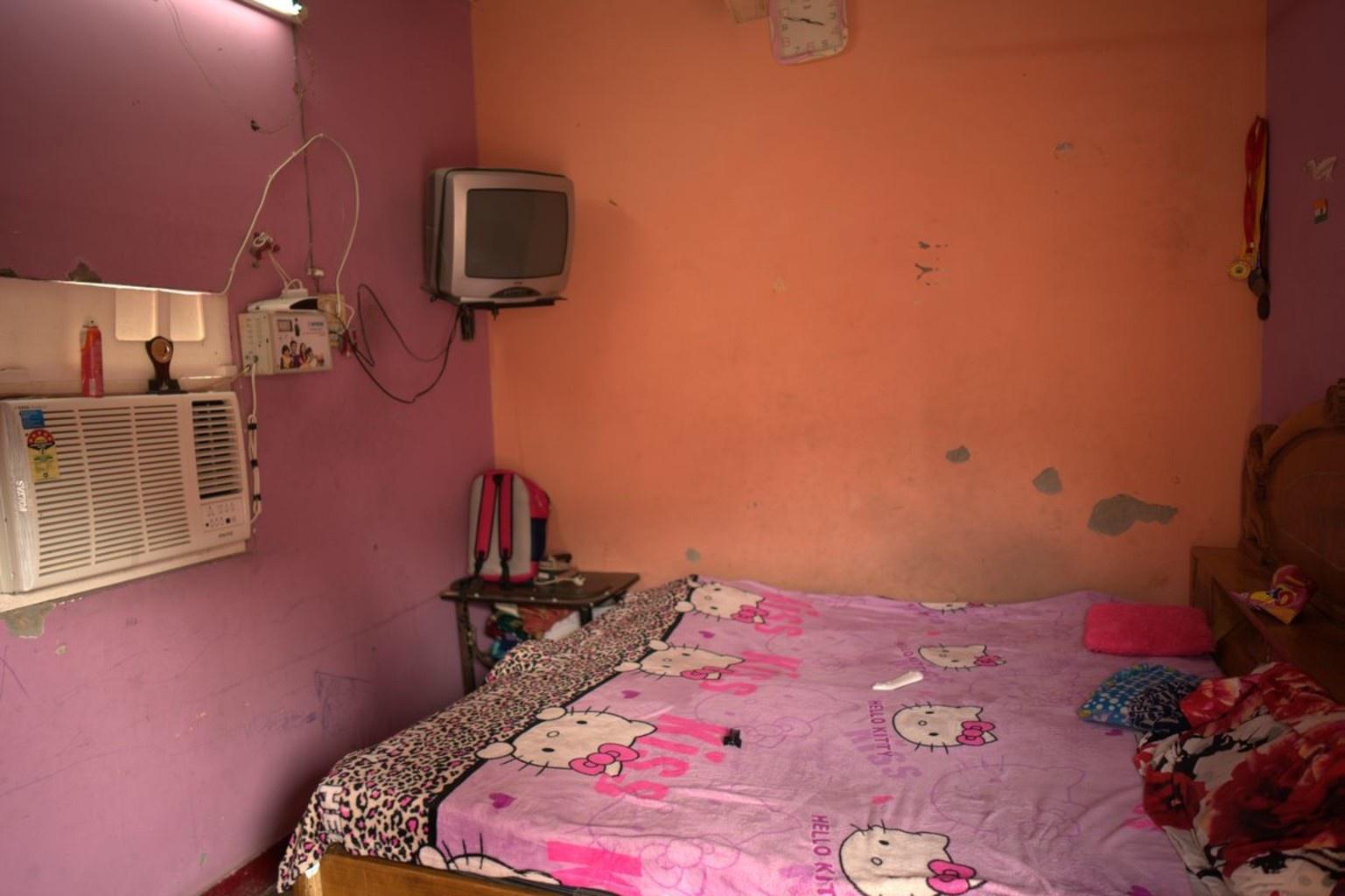}}\hspace{0.01cm}
    \subfloat[\scriptsize{\checkmark}]{\includegraphics[width=1.6cm,height=1.6cm]{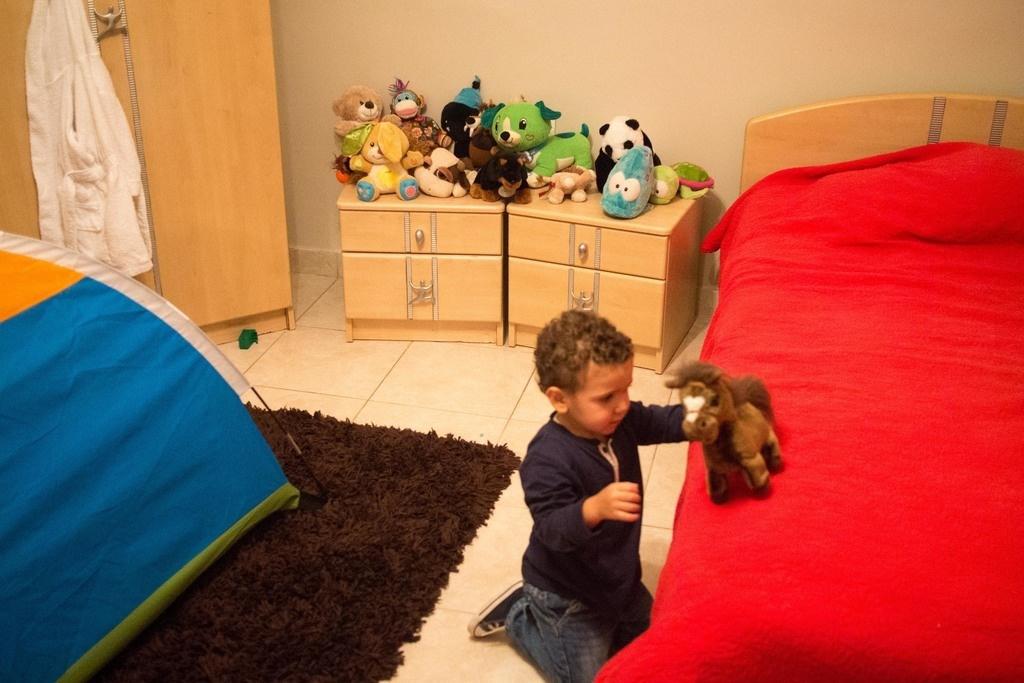}}\hspace{0.01cm}
    \subfloat[\scriptsize{\checkmark}]{\includegraphics[width=1.6cm,height=1.6cm]{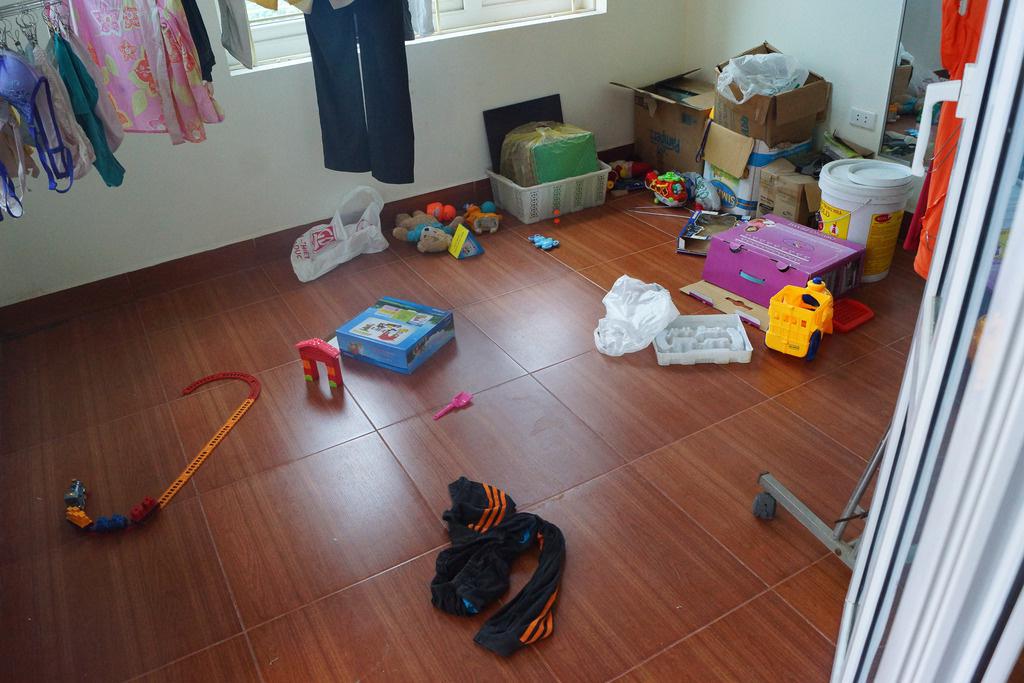}}\hspace{0.01cm}
    \subfloat[\scriptsize{\checkmark}]{\includegraphics[width=1.6cm,height=1.6cm]{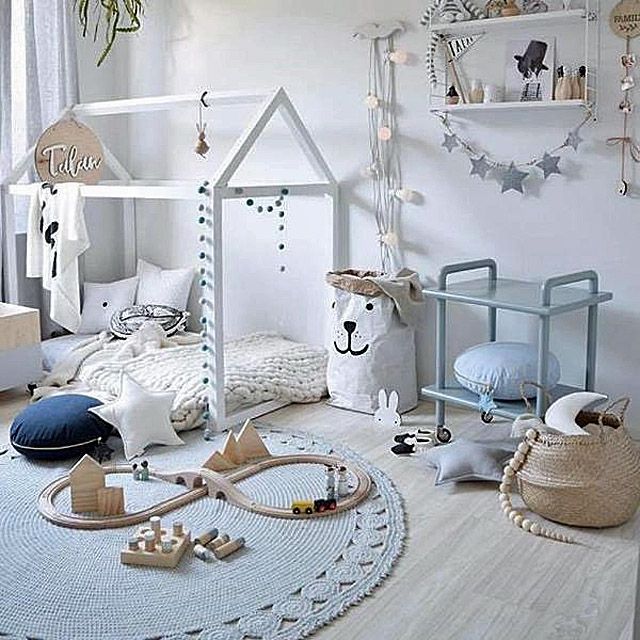}}\hspace{0.01cm}
    \subfloat[\scriptsize{bedroom}]{\includegraphics[width=1.6cm,height=1.6cm]{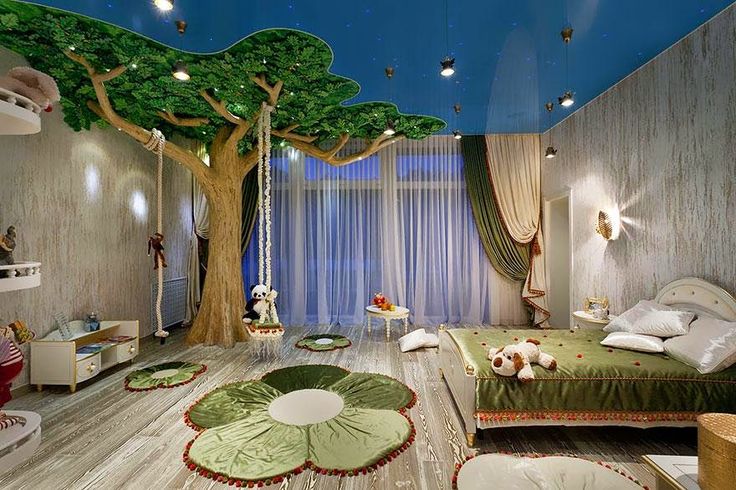}}\hspace{0.01cm}
    \subfloat[\scriptsize{bedroom}]{\includegraphics[width=1.6cm,height=1.6cm]{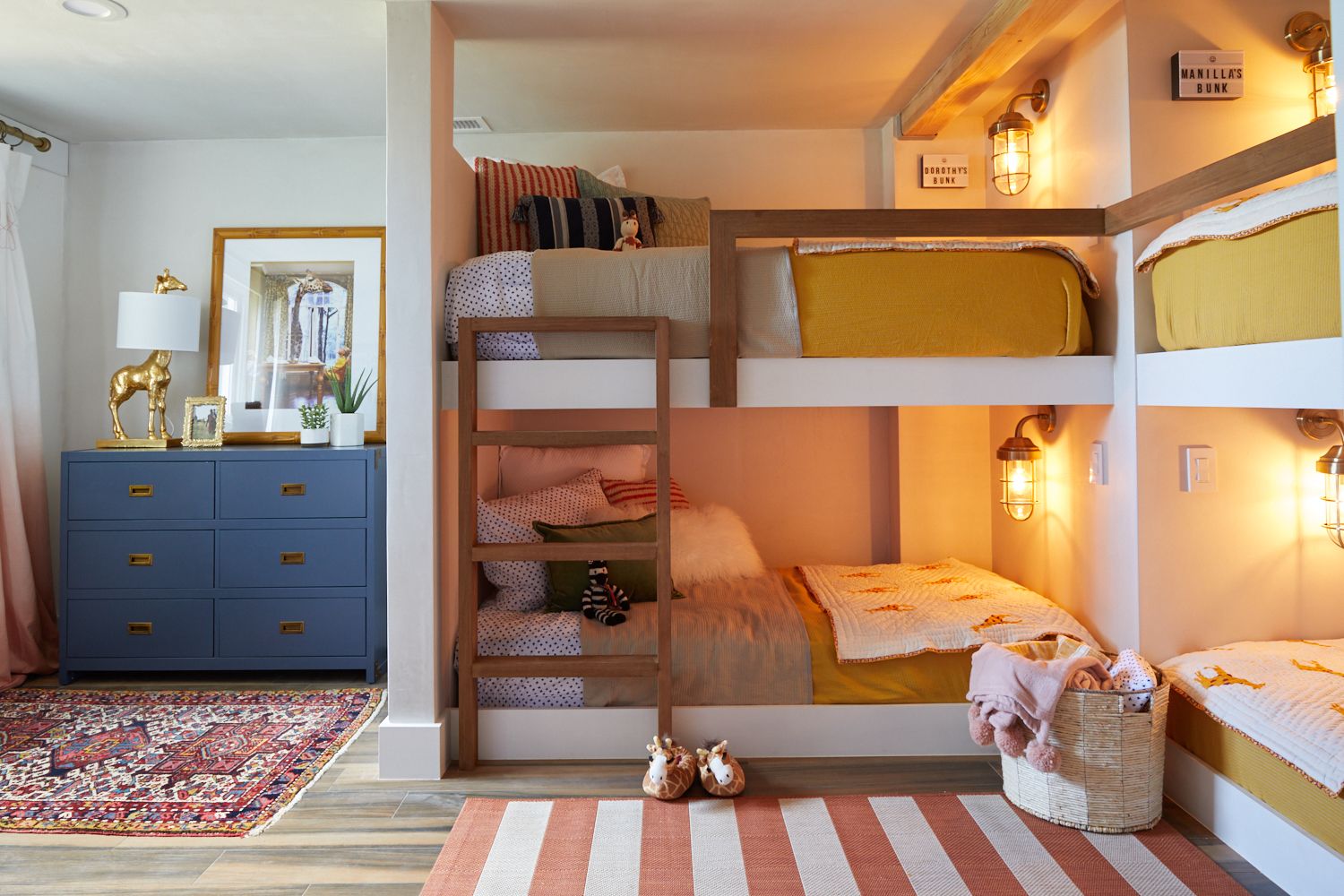}}\hspace{0.01cm}
    \subfloat[\scriptsize{\checkmark}]{\includegraphics[width=1.6cm,height=1.6cm]{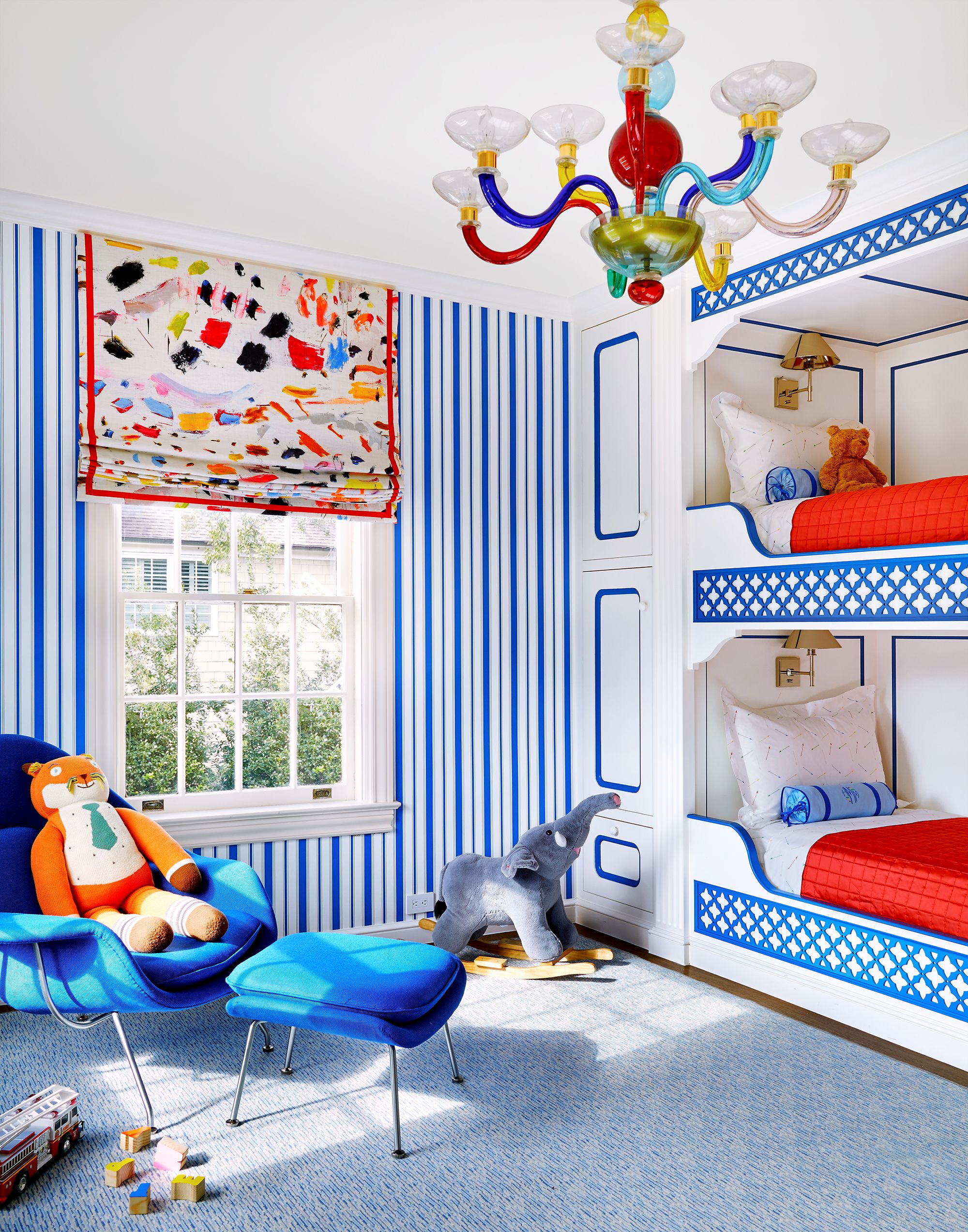}}\hspace{0.01cm}
    \subfloat[\scriptsize{bedroom}]{\includegraphics[width=1.6cm,height=1.6cm]{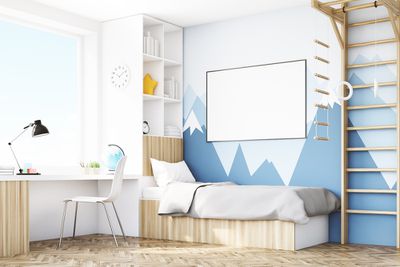}}\\ \vspace{-0.2cm}

    {\centering \rotatebox[origin=lb]{90}{\scriptsize{~~~classroom}}}
    \subfloat[\scriptsize{\checkmark}]{\includegraphics[width=1.6cm,height=1.6cm]{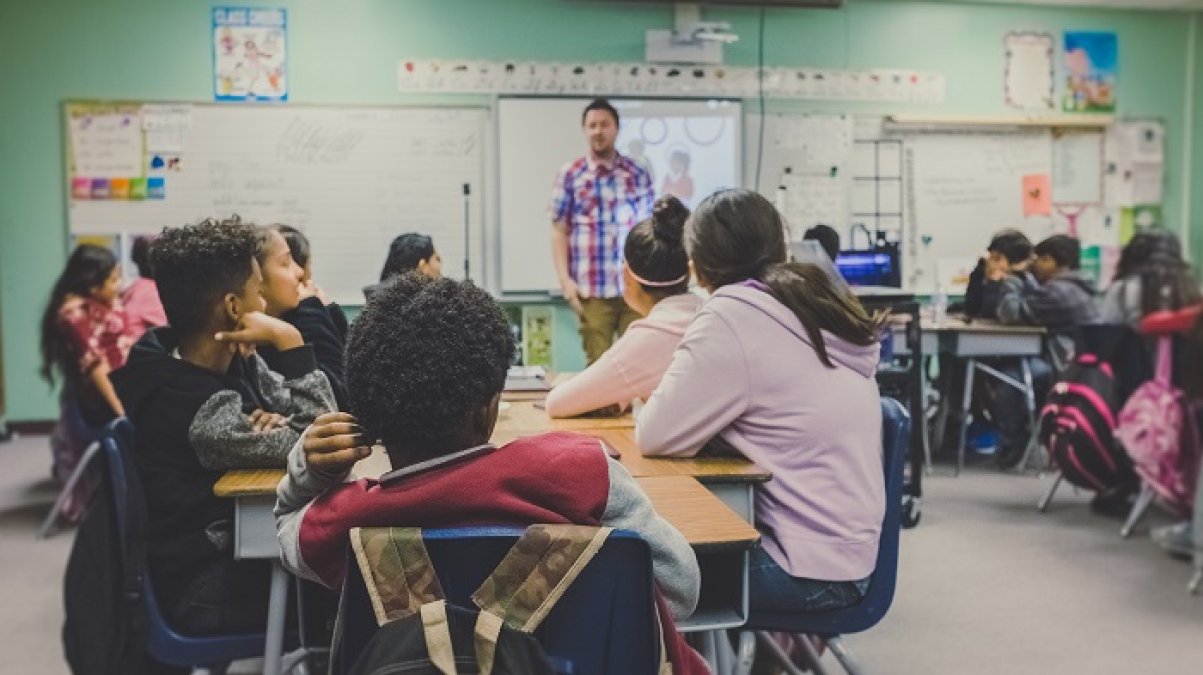}}\hspace{0.01cm}
    \subfloat[\scriptsize{\checkmark}]{\includegraphics[width=1.6cm,height=1.6cm]{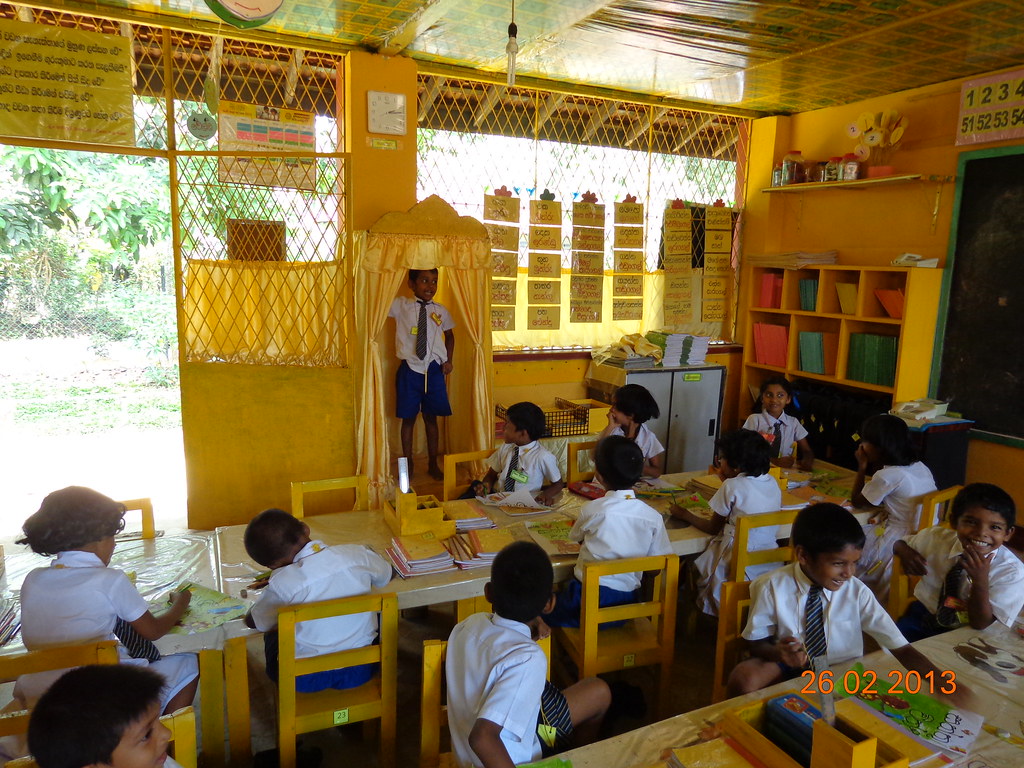}}\hspace{0.01cm}
    \subfloat[\scriptsize{\checkmark}]{\includegraphics[width=1.6cm,height=1.6cm]{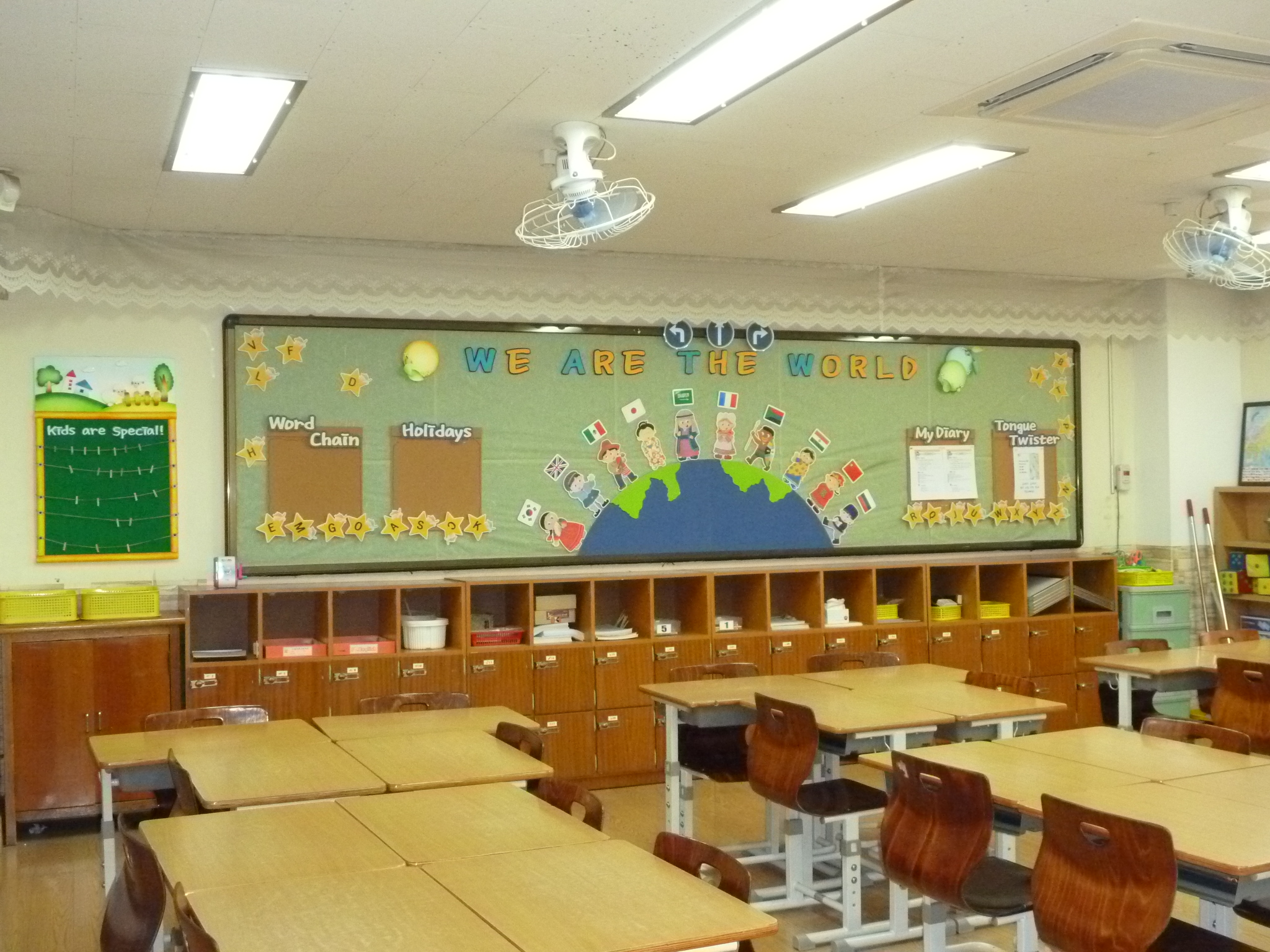}}\hspace{0.01cm}
    \subfloat[\scriptsize{\checkmark}]{\includegraphics[width=1.6cm,height=1.6cm]{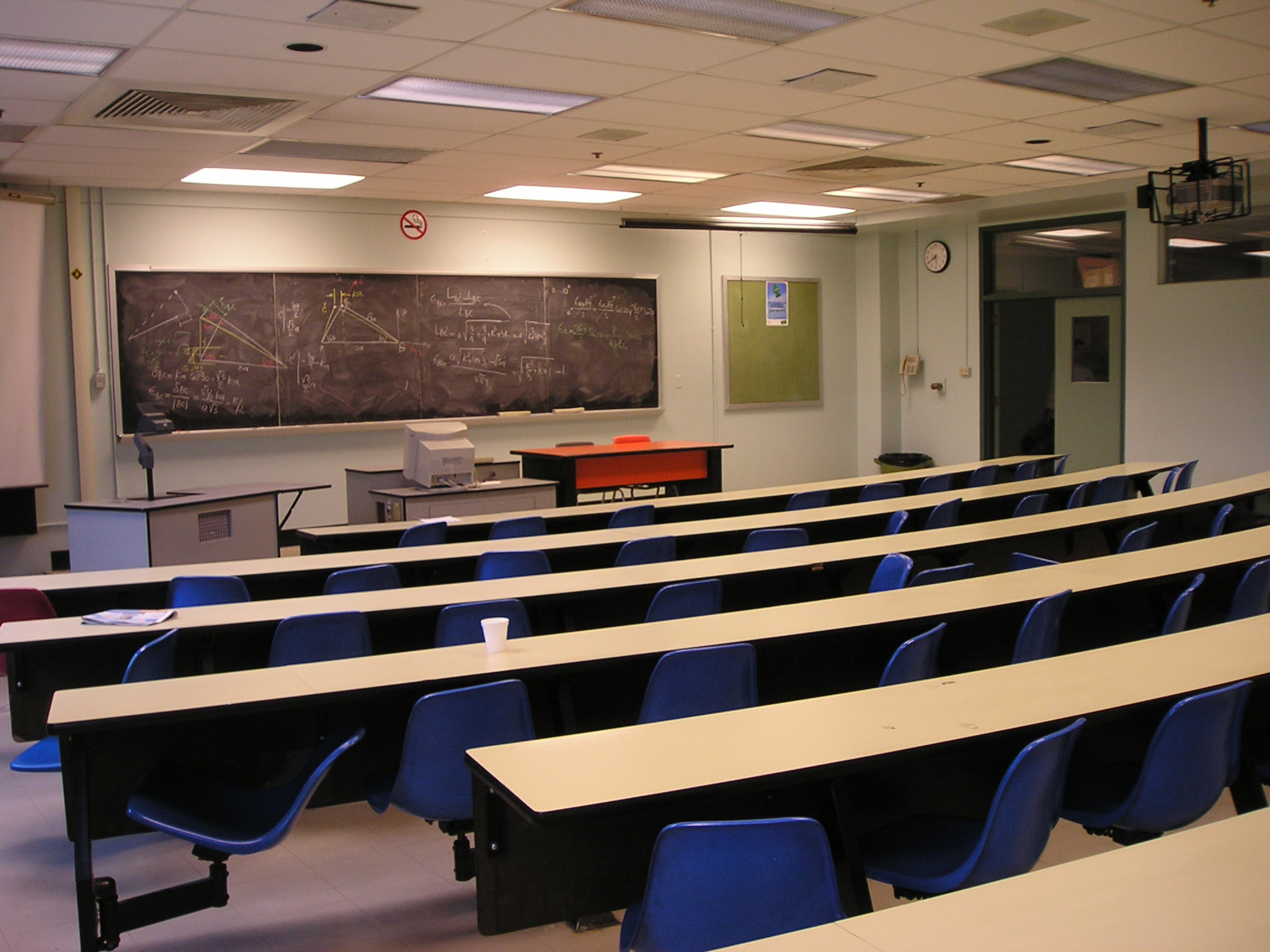}}\hspace{0.01cm}
    \subfloat[\scriptsize{\checkmark}]{\includegraphics[width=1.6cm,height=1.6cm]{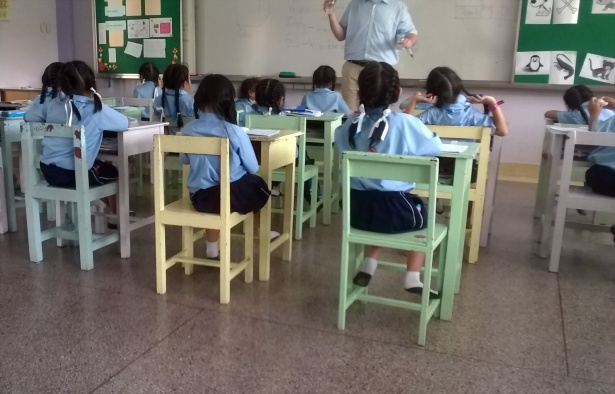}}\hspace{0.01cm}
    \subfloat[\scriptsize{\checkmark}]{\includegraphics[width=1.6cm,height=1.6cm]{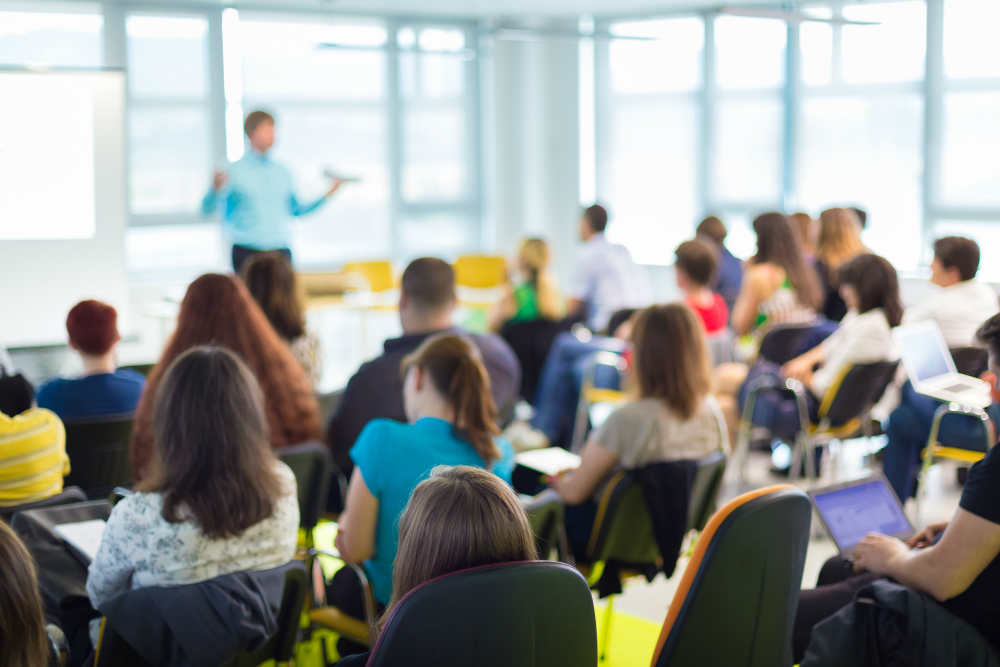}}\hspace{0.01cm}
    \subfloat[\scriptsize{\checkmark}]{\includegraphics[width=1.6cm,height=1.6cm]{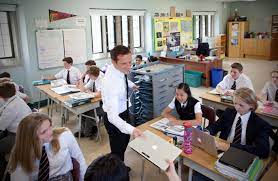}}\hspace{0.01cm}
    \subfloat[\scriptsize{\checkmark}]{\includegraphics[width=1.6cm,height=1.6cm]{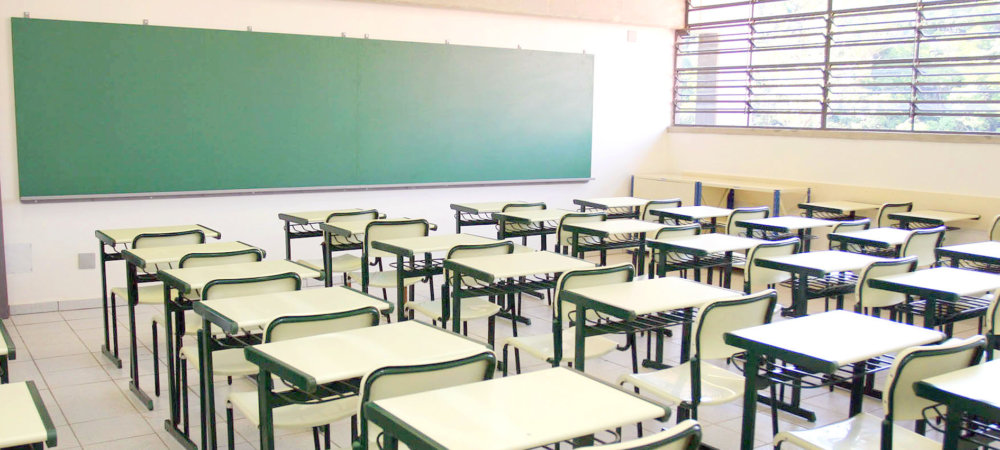}}\hspace{0.01cm}
    \subfloat[\scriptsize{\checkmark}]{\includegraphics[width=1.6cm,height=1.6cm]{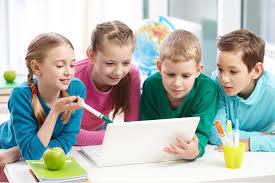}}\hspace{0.01cm}
    \subfloat[\scriptsize{\checkmark}]{\includegraphics[width=1.6cm,height=1.6cm]{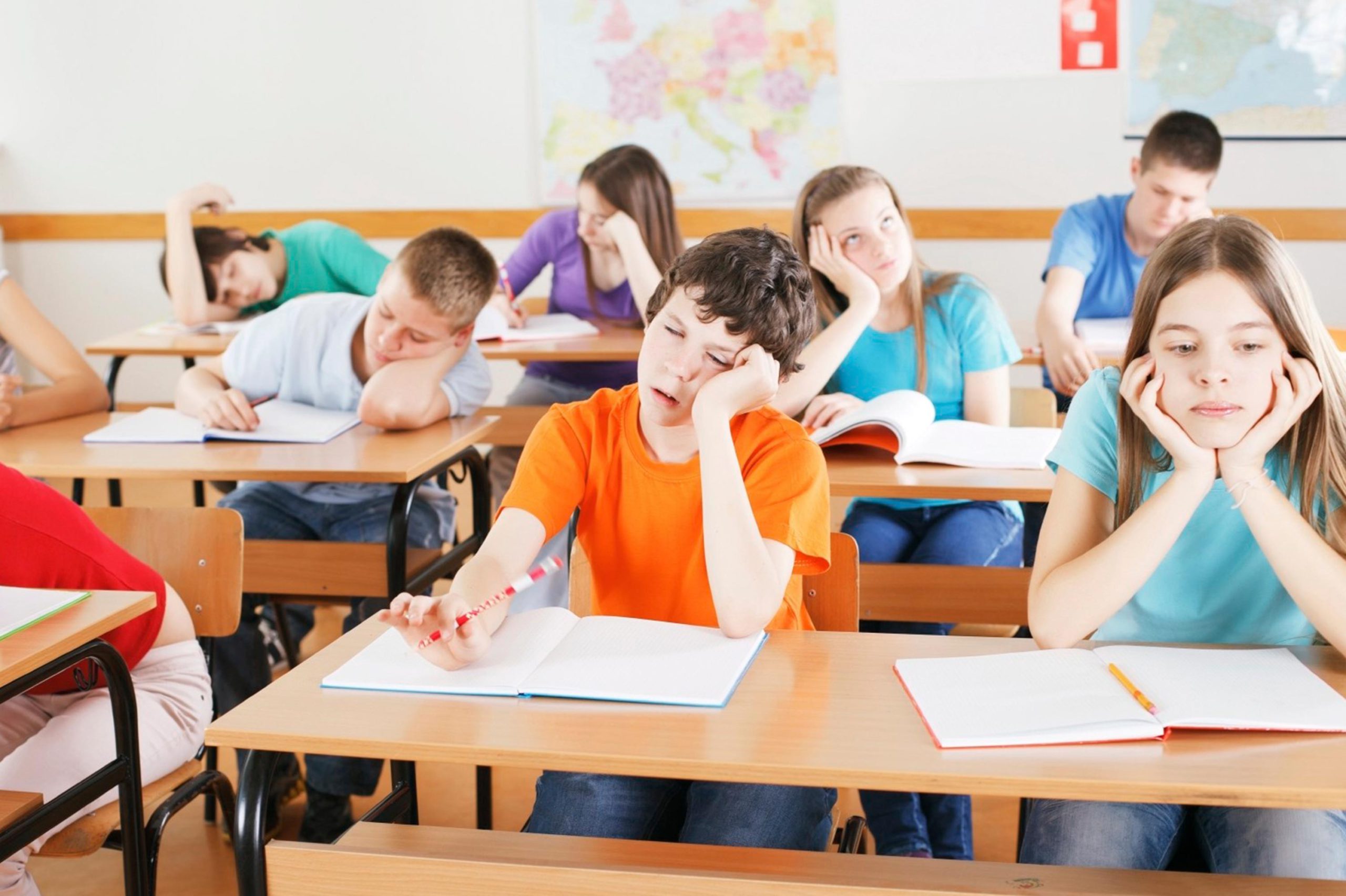}}\\ \vspace{-0.2cm}

    {\centering \rotatebox[origin=lb]{90}{\scriptsize{~~dressing room}}}
    \subfloat[\scriptsize{bathroom}]{\includegraphics[width=1.6cm,height=1.6cm]{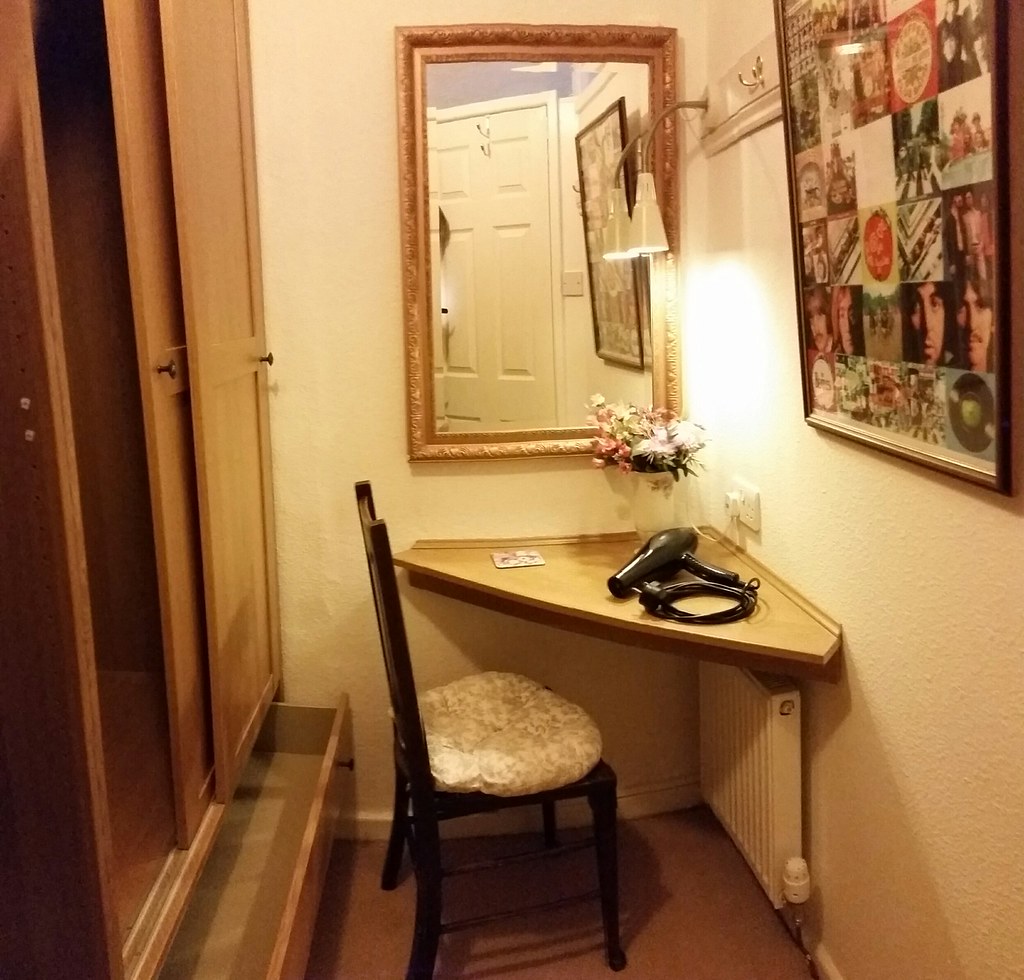}}\hspace{0.01cm}
    \subfloat[\scriptsize{\checkmark}]{\includegraphics[width=1.6cm,height=1.6cm]{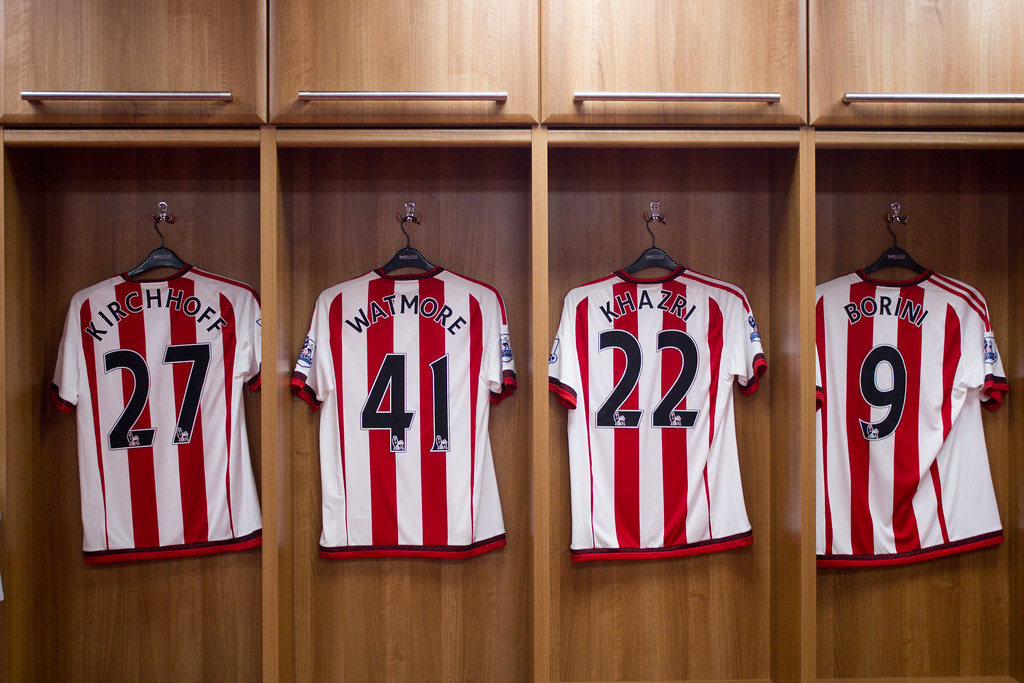}}\hspace{0.01cm}
    \subfloat[\scriptsize{living room}]{\includegraphics[width=1.6cm,height=1.6cm]{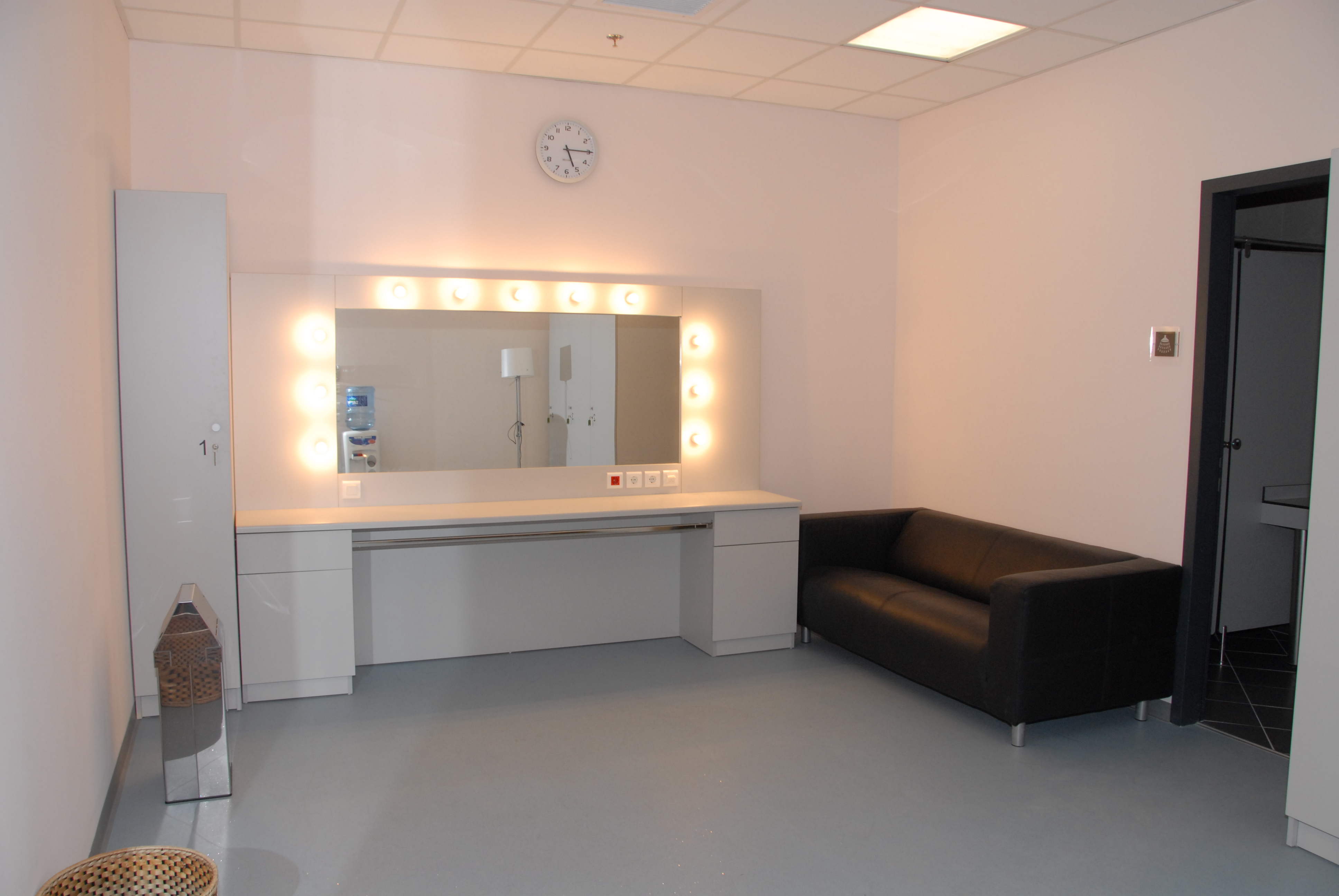}}\hspace{0.01cm}
    \subfloat[\scriptsize{\checkmark}]{\includegraphics[width=1.6cm,height=1.6cm]{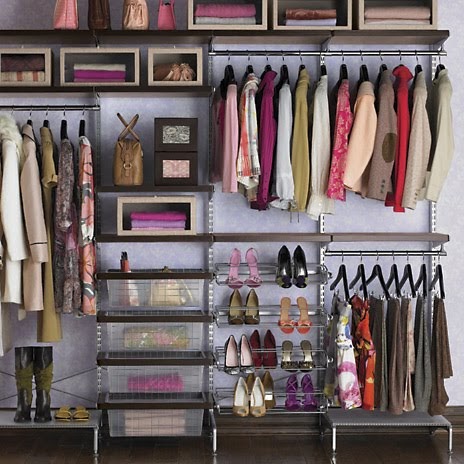}}\hspace{0.01cm}
    \subfloat[\scriptsize{\checkmark}]{\includegraphics[width=1.6cm,height=1.6cm]{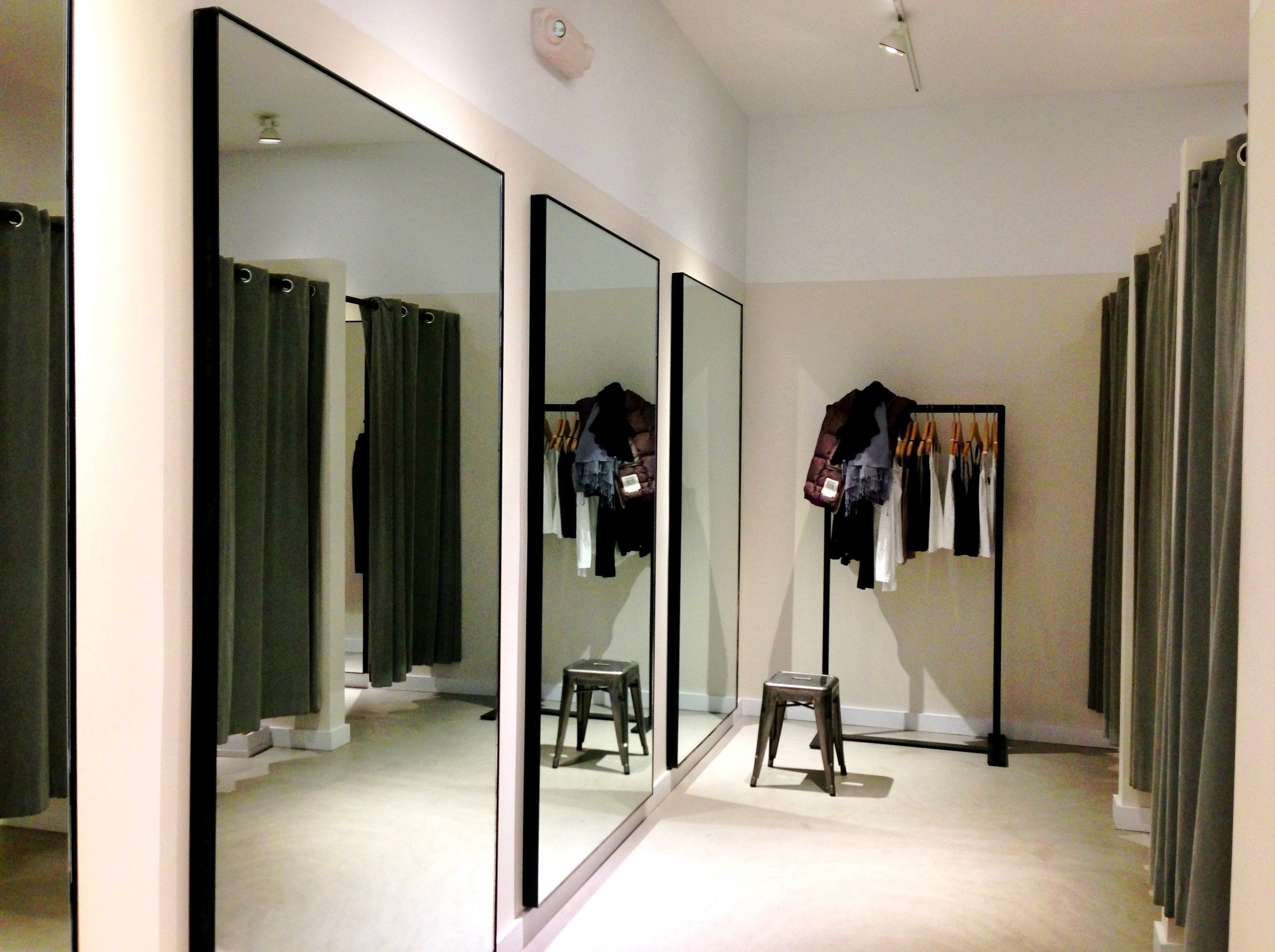}}\hspace{0.01cm}
    \subfloat[\scriptsize{\checkmark}]{\includegraphics[width=1.6cm,height=1.6cm]{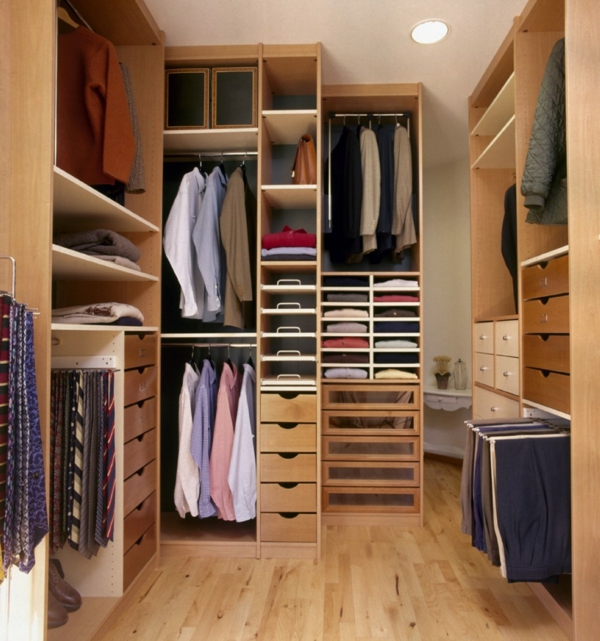}}\hspace{0.01cm}
    \subfloat[\scriptsize{\checkmark}]{\includegraphics[width=1.6cm,height=1.6cm]{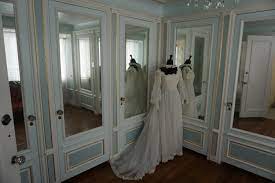}}\hspace{0.01cm}
    \subfloat[\scriptsize{\checkmark}]{\includegraphics[width=1.6cm,height=1.6cm]{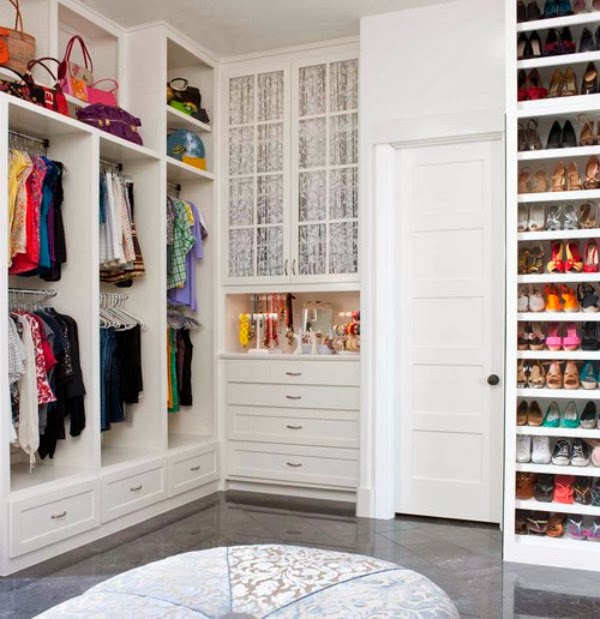}}\hspace{0.01cm}
    \subfloat[\scriptsize{\checkmark}]{\includegraphics[width=1.6cm,height=1.6cm]{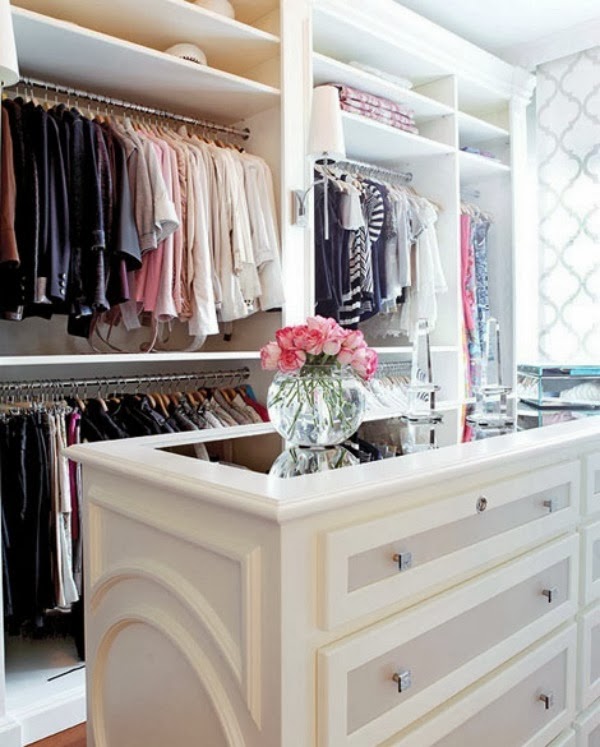}}\hspace{0.01cm}
    \subfloat[\scriptsize{\checkmark}]{\includegraphics[width=1.6cm,height=1.6cm]{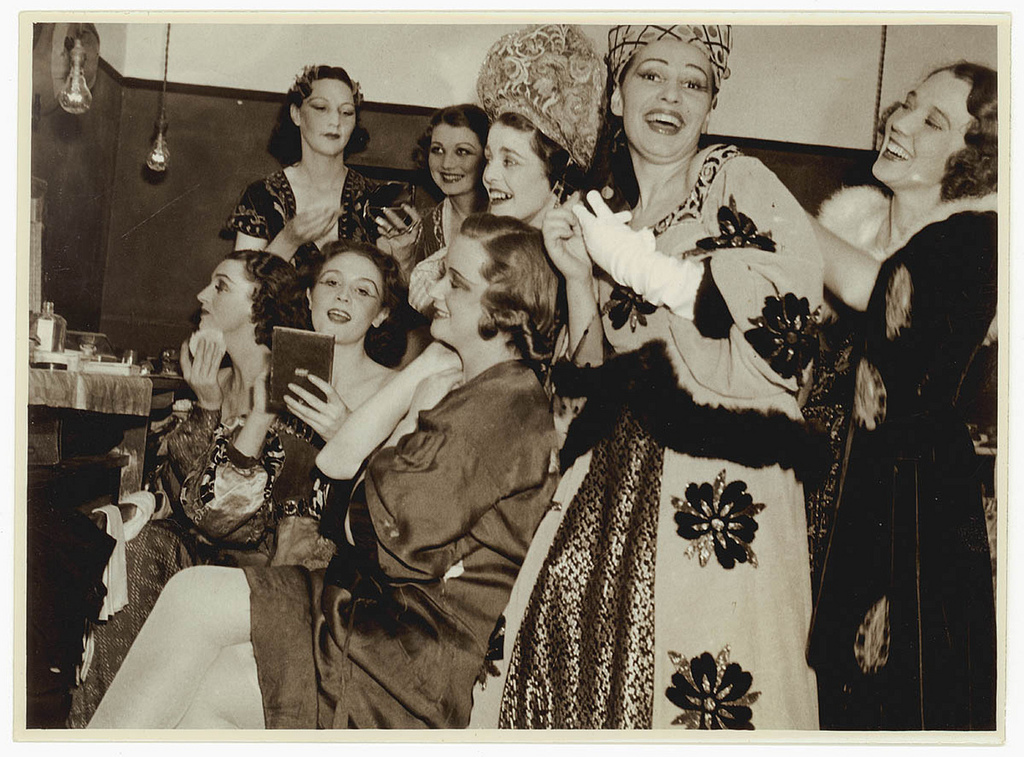}}\\ \vspace{-0.2cm}

    {\centering \rotatebox[origin=lb]{90}{\scriptsize{~~~living room}}}
    \subfloat[\scriptsize{bathroom}]{\includegraphics[width=1.6cm,height=1.6cm]{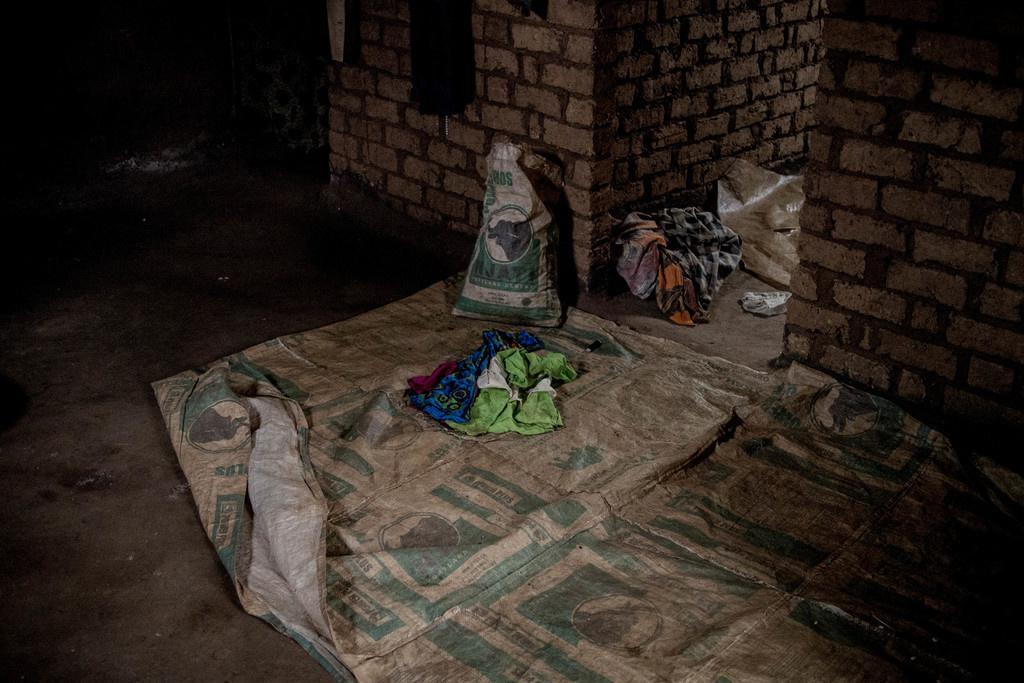}}\hspace{0.01cm}
    \subfloat[\scriptsize{\checkmark}]{\includegraphics[width=1.6cm,height=1.6cm]{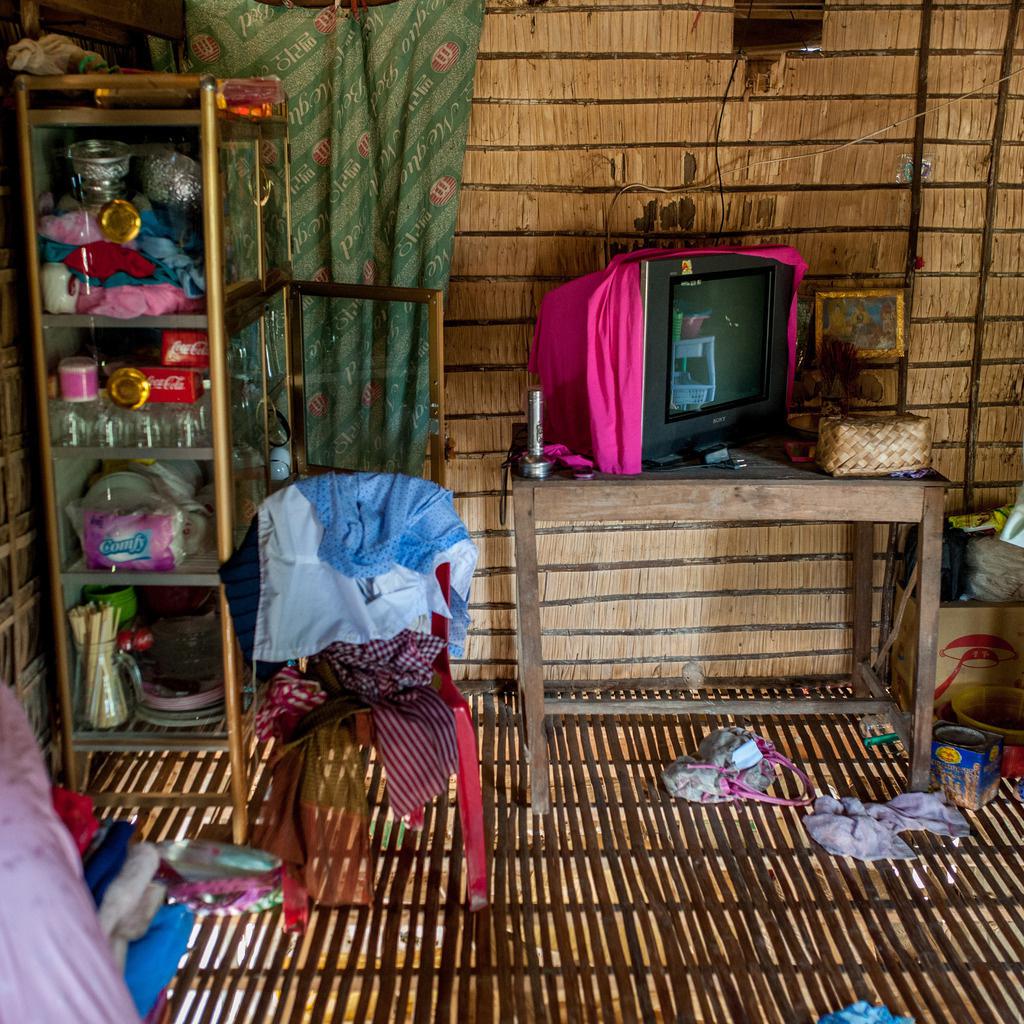}}\hspace{0.01cm}
    \subfloat[\scriptsize{child's room}]{\includegraphics[width=1.6cm,height=1.6cm]{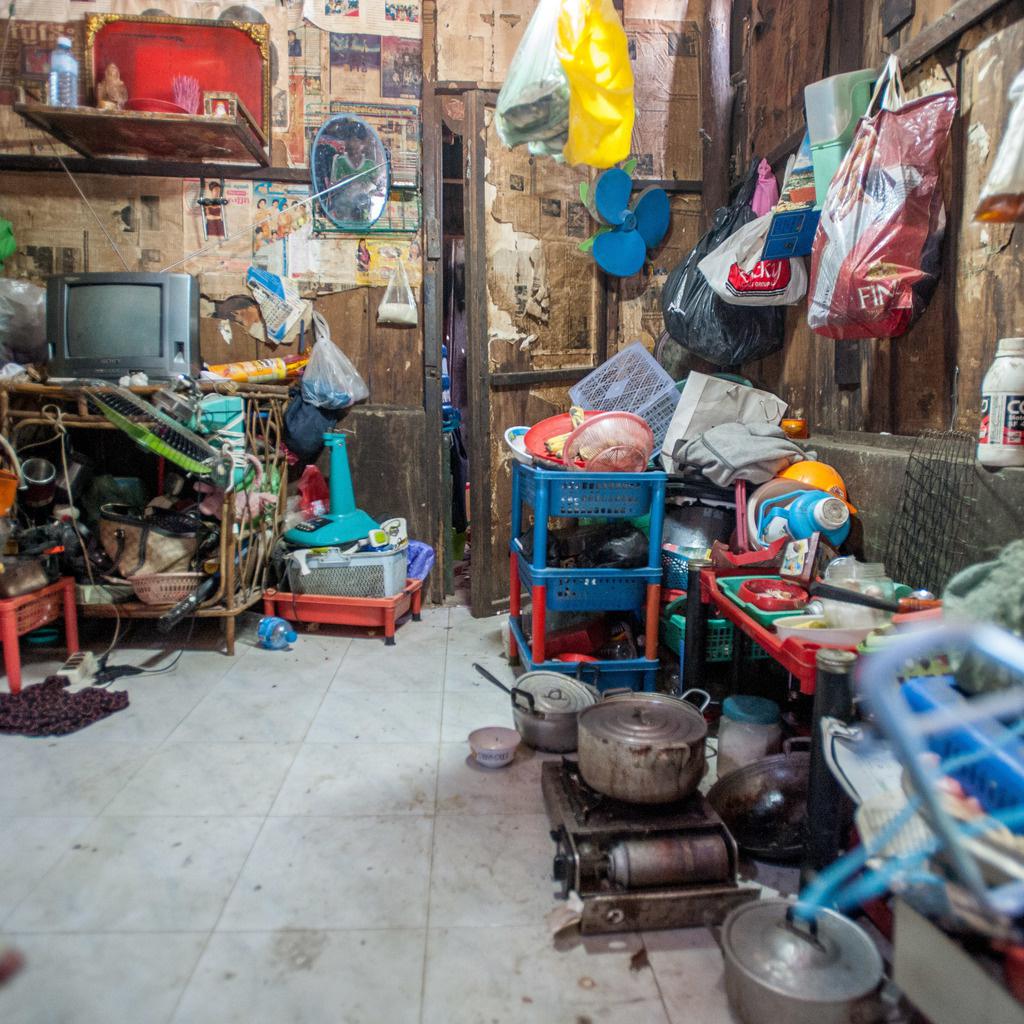}}\hspace{0.01cm}
    \subfloat[\scriptsize{child's room}]{\includegraphics[width=1.6cm,height=1.6cm]{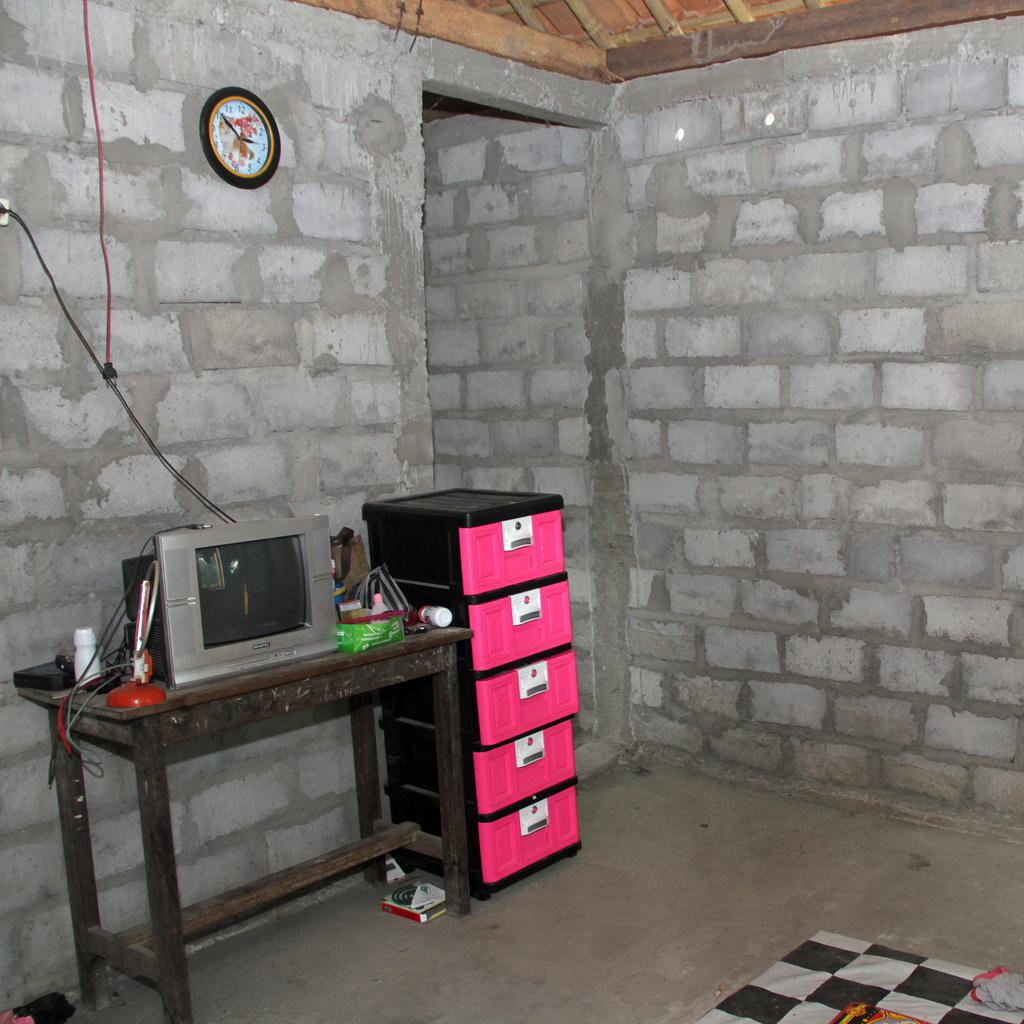}}\hspace{0.01cm}
    \subfloat[\scriptsize{bedroom}]{\includegraphics[width=1.6cm,height=1.6cm]{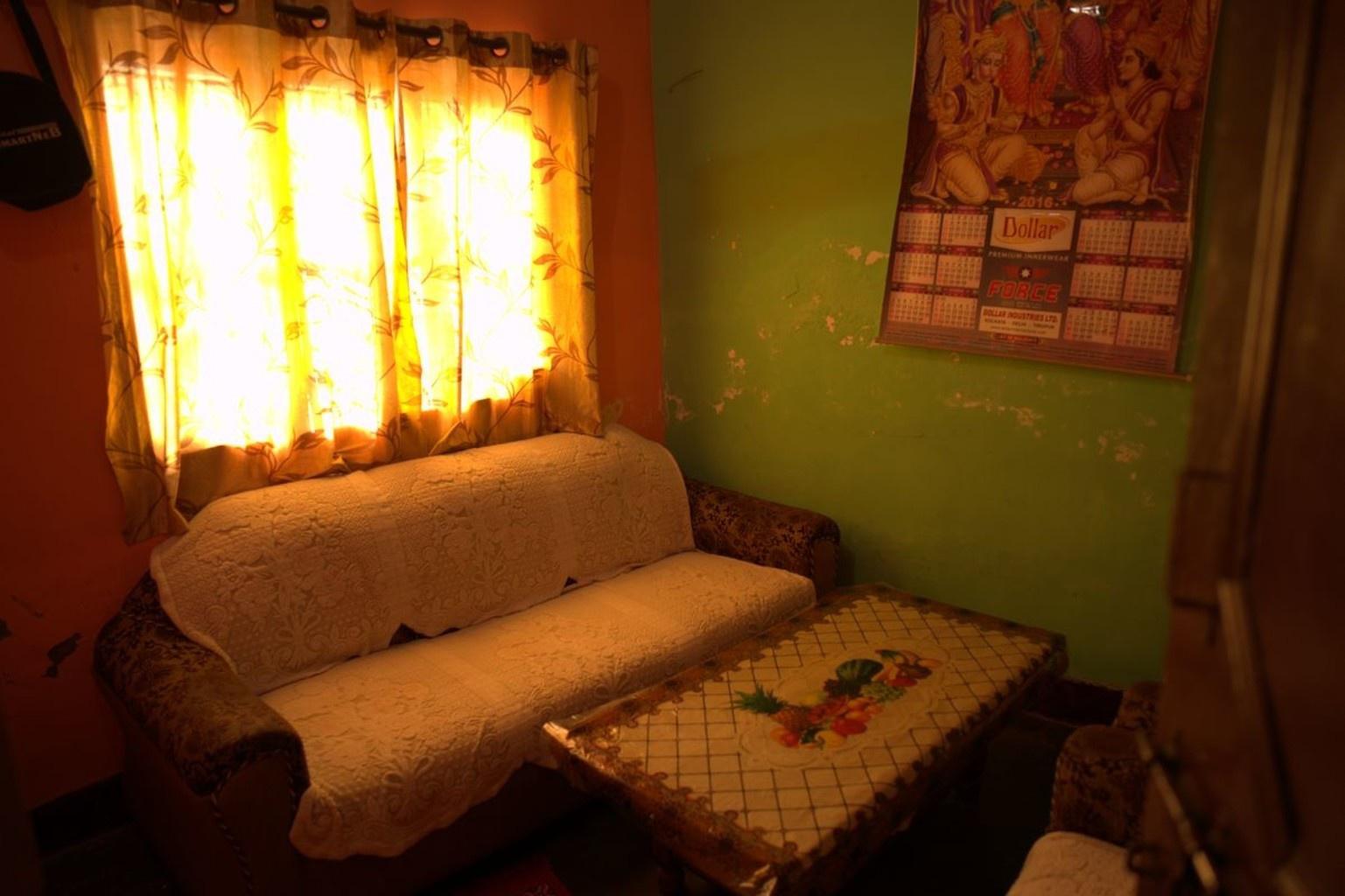}}\hspace{0.01cm}
    \subfloat[\scriptsize{\checkmark}]{\includegraphics[width=1.6cm,height=1.6cm]{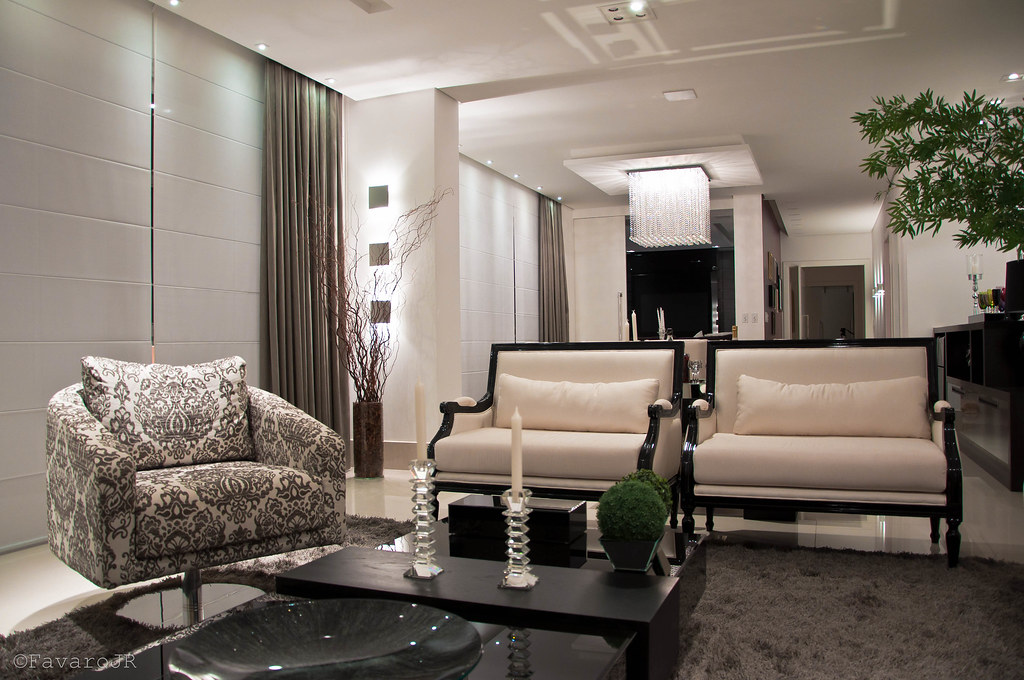}}\hspace{0.01cm}
    \subfloat[\scriptsize{\checkmark}]{\includegraphics[width=1.6cm,height=1.6cm]{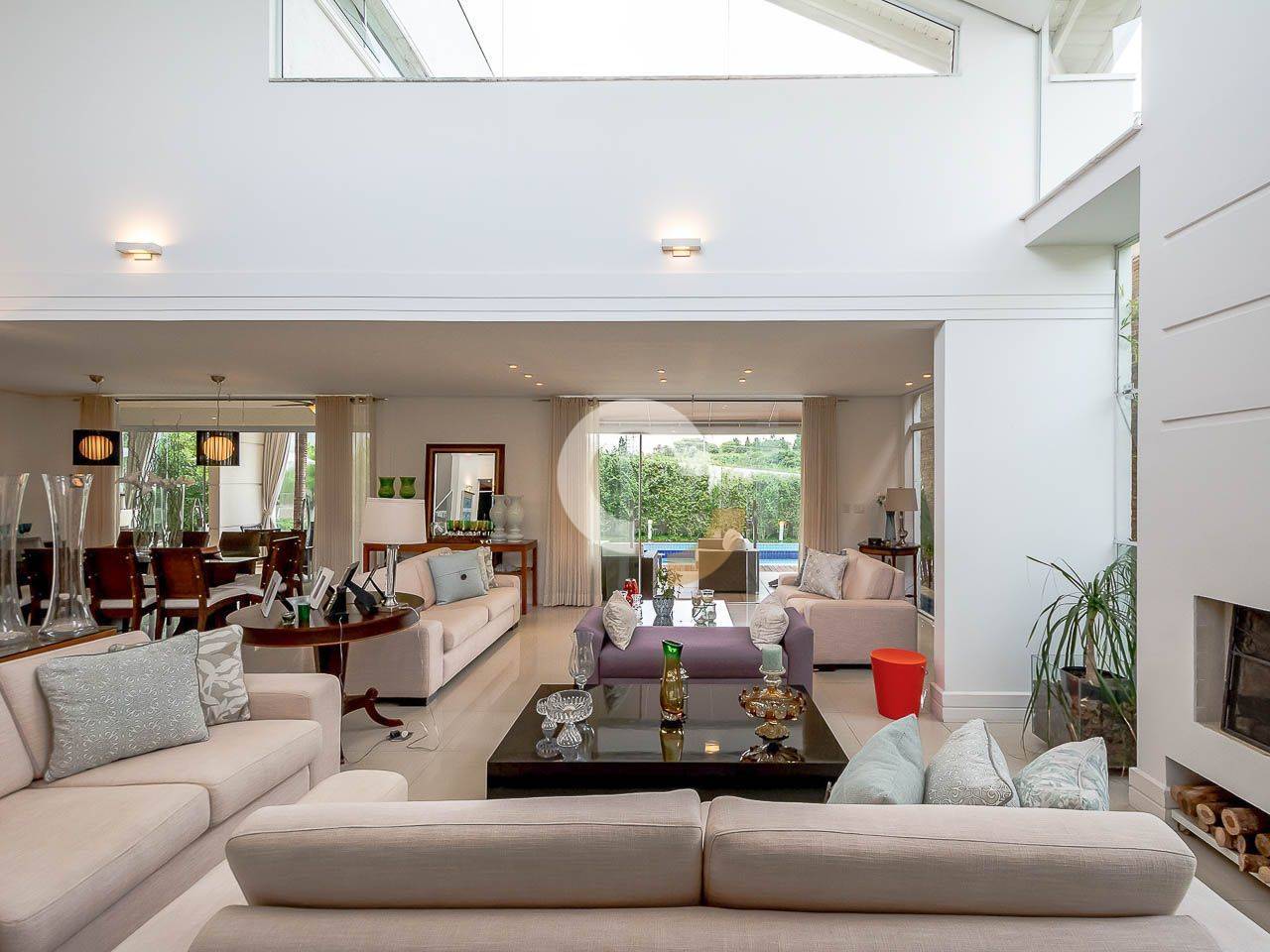}}\hspace{0.01cm}
    \subfloat[\scriptsize{\checkmark}]{\includegraphics[width=1.6cm,height=1.6cm]{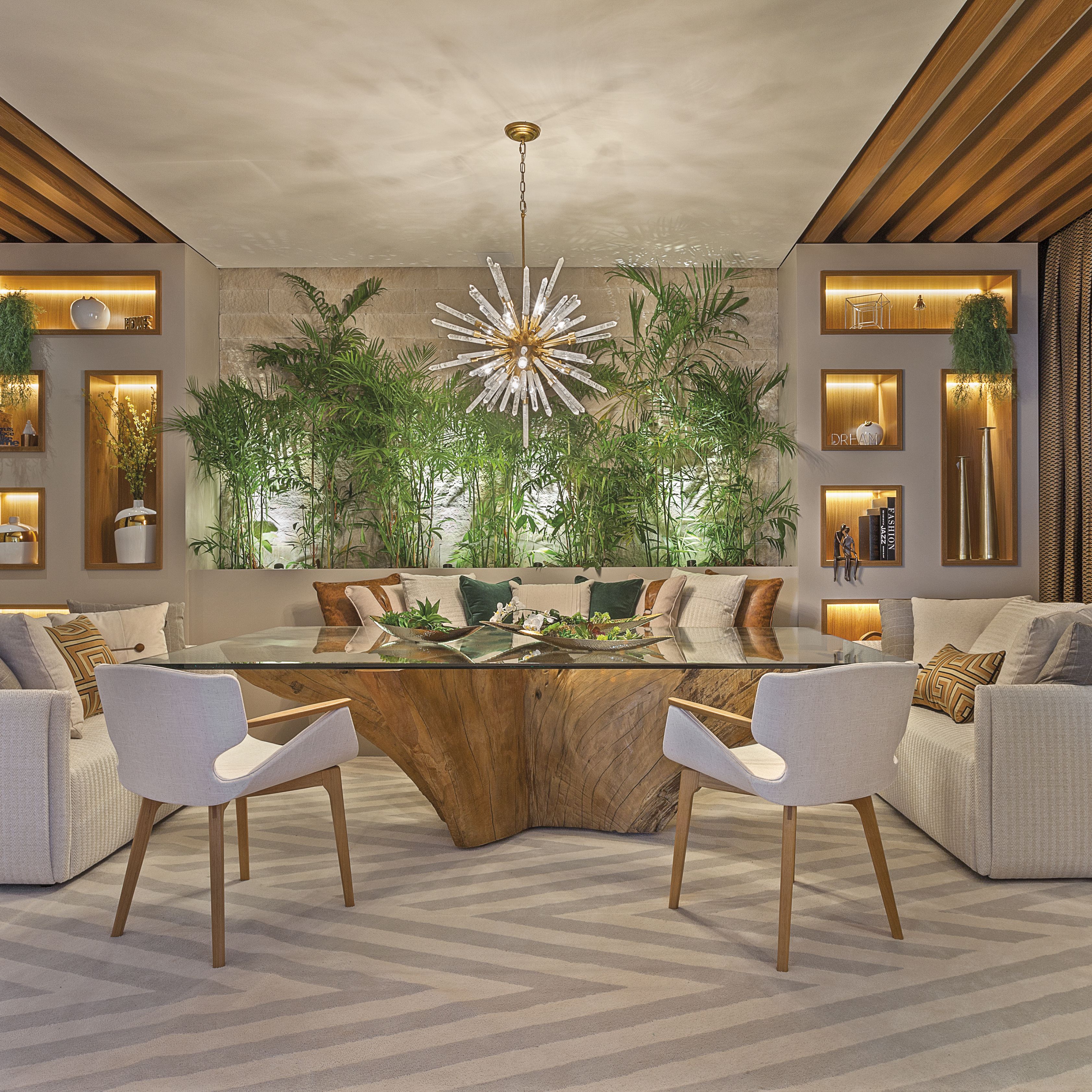}}\hspace{0.01cm}
    \subfloat[\scriptsize{\checkmark}]{\includegraphics[width=1.6cm,height=1.6cm]{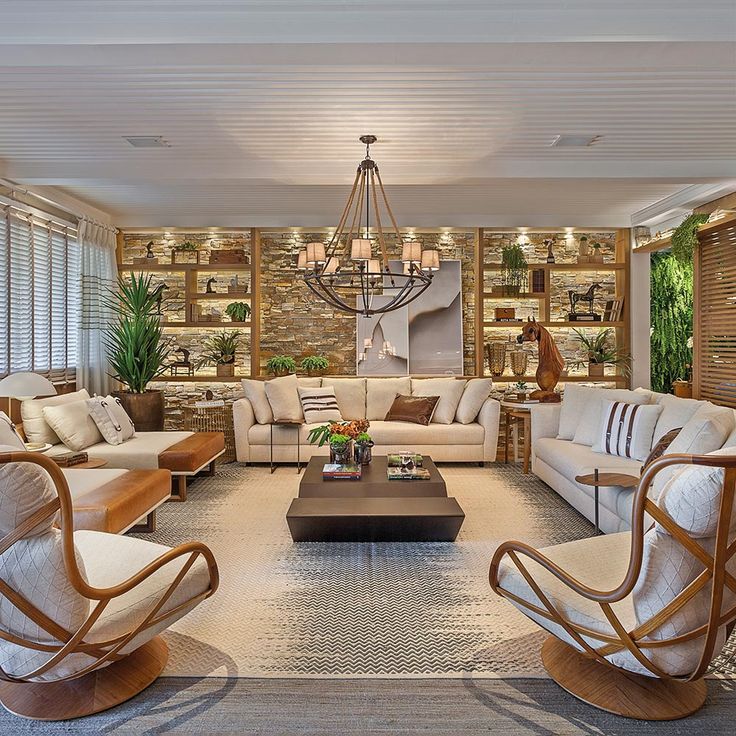}}\hspace{0.01cm}
    \subfloat[\scriptsize{\checkmark}]{\includegraphics[width=1.6cm,height=1.6cm]{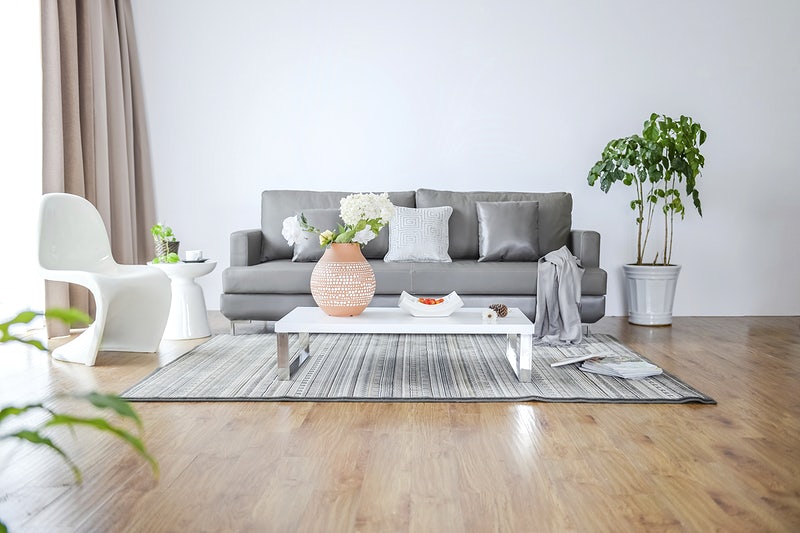}}\\ \vspace{-0.2cm}

    {\centering \rotatebox[origin=lb]{90}{\scriptsize{~~~~~~studio}}}
    \subfloat[\scriptsize{dressing room}]{\includegraphics[width=1.6cm,height=1.6cm]{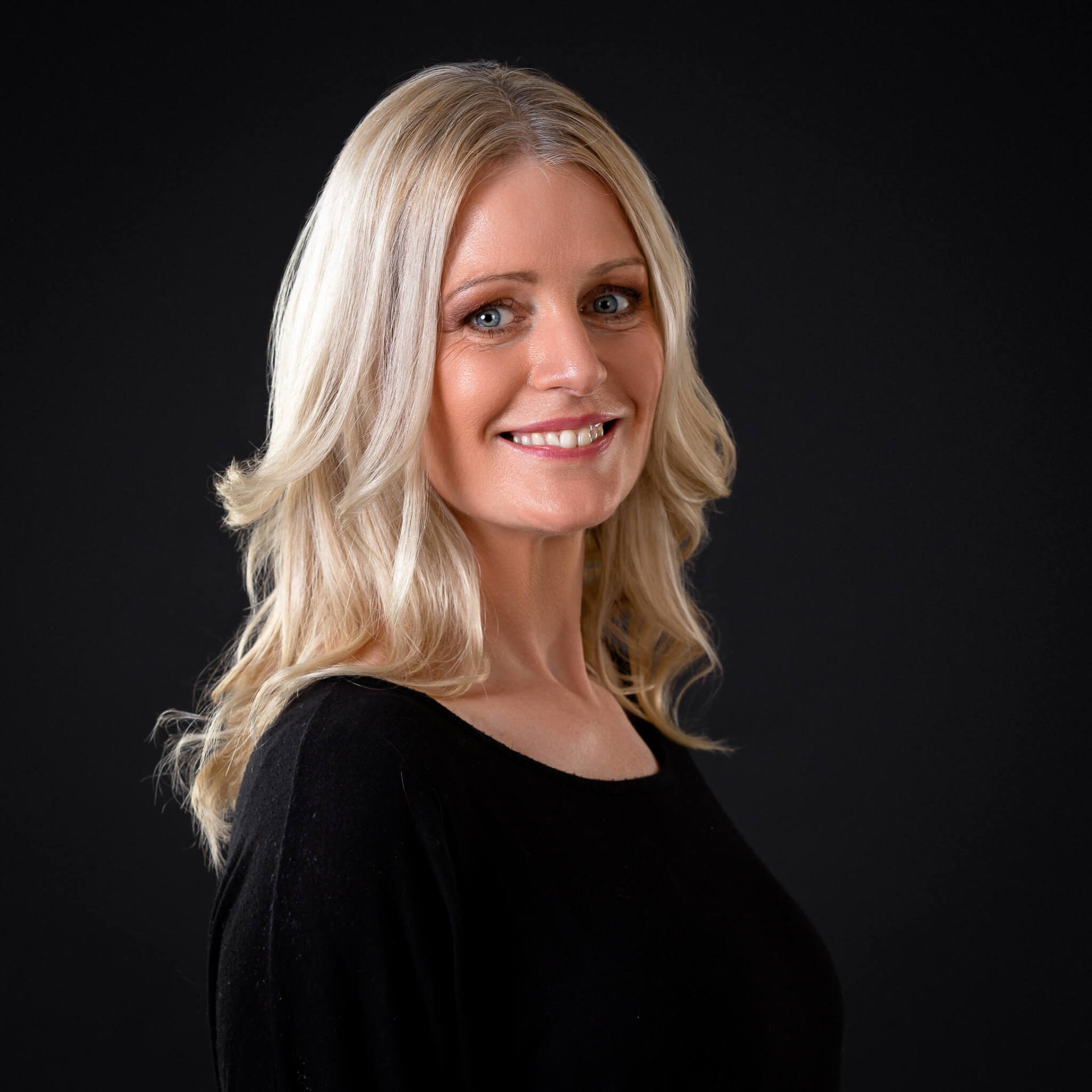}}\hspace{0.01cm}
    \subfloat[\scriptsize{\checkmark}]{\includegraphics[width=1.6cm,height=1.6cm]{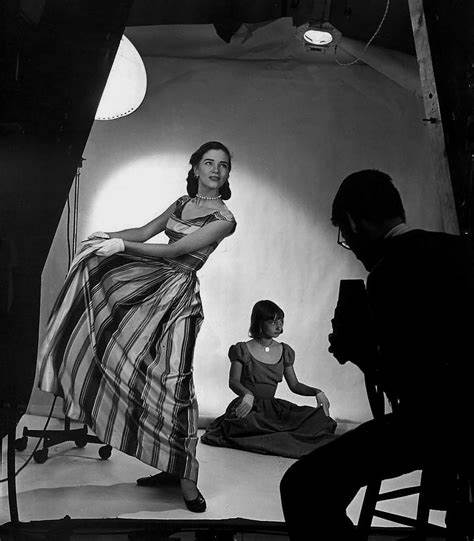}}\hspace{0.01cm}
    \subfloat[\scriptsize{\checkmark}]{\includegraphics[width=1.6cm,height=1.6cm]{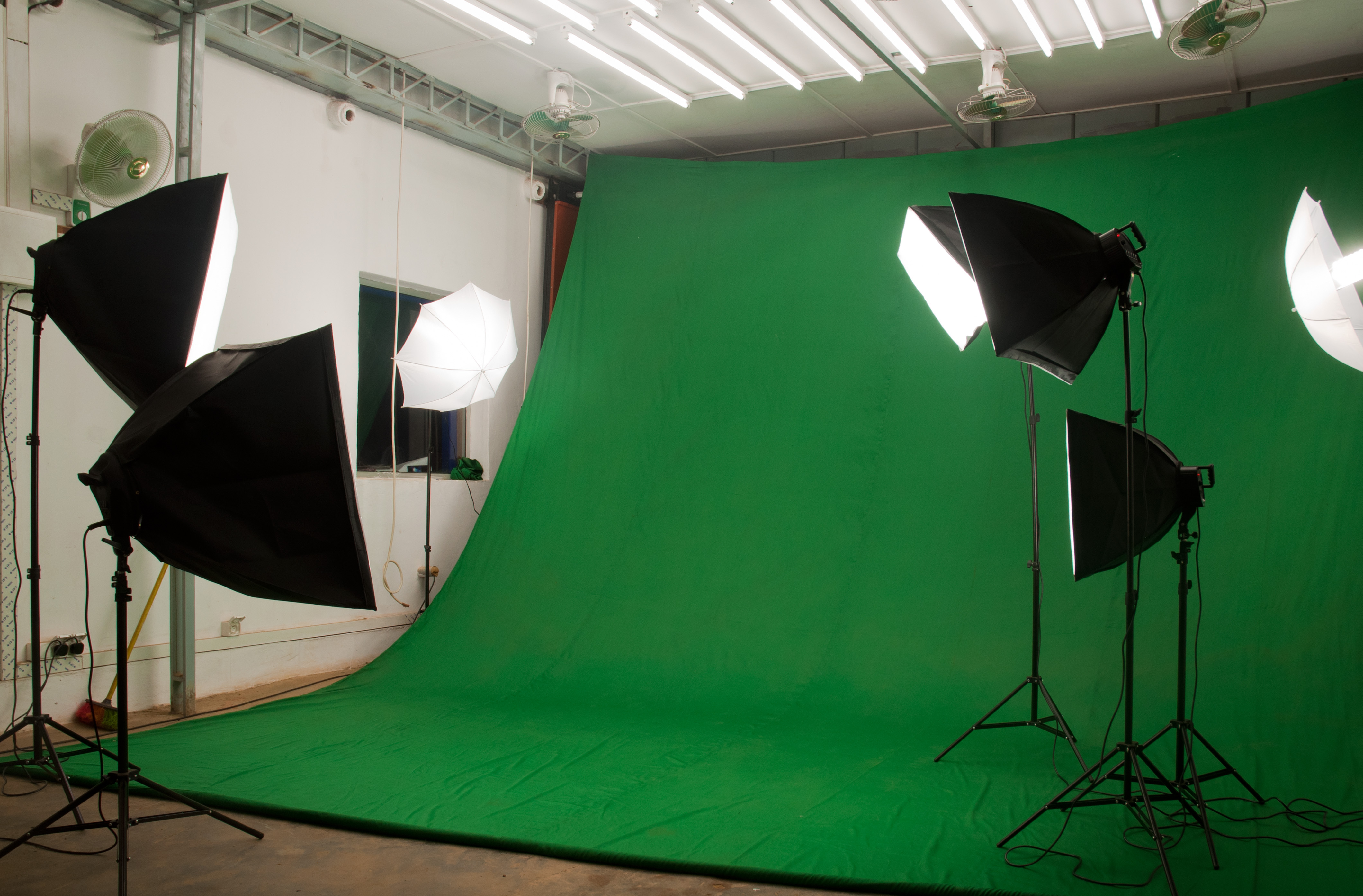}}\hspace{0.01cm}
    \subfloat[\scriptsize{child's room}]{\includegraphics[width=1.6cm,height=1.6cm]{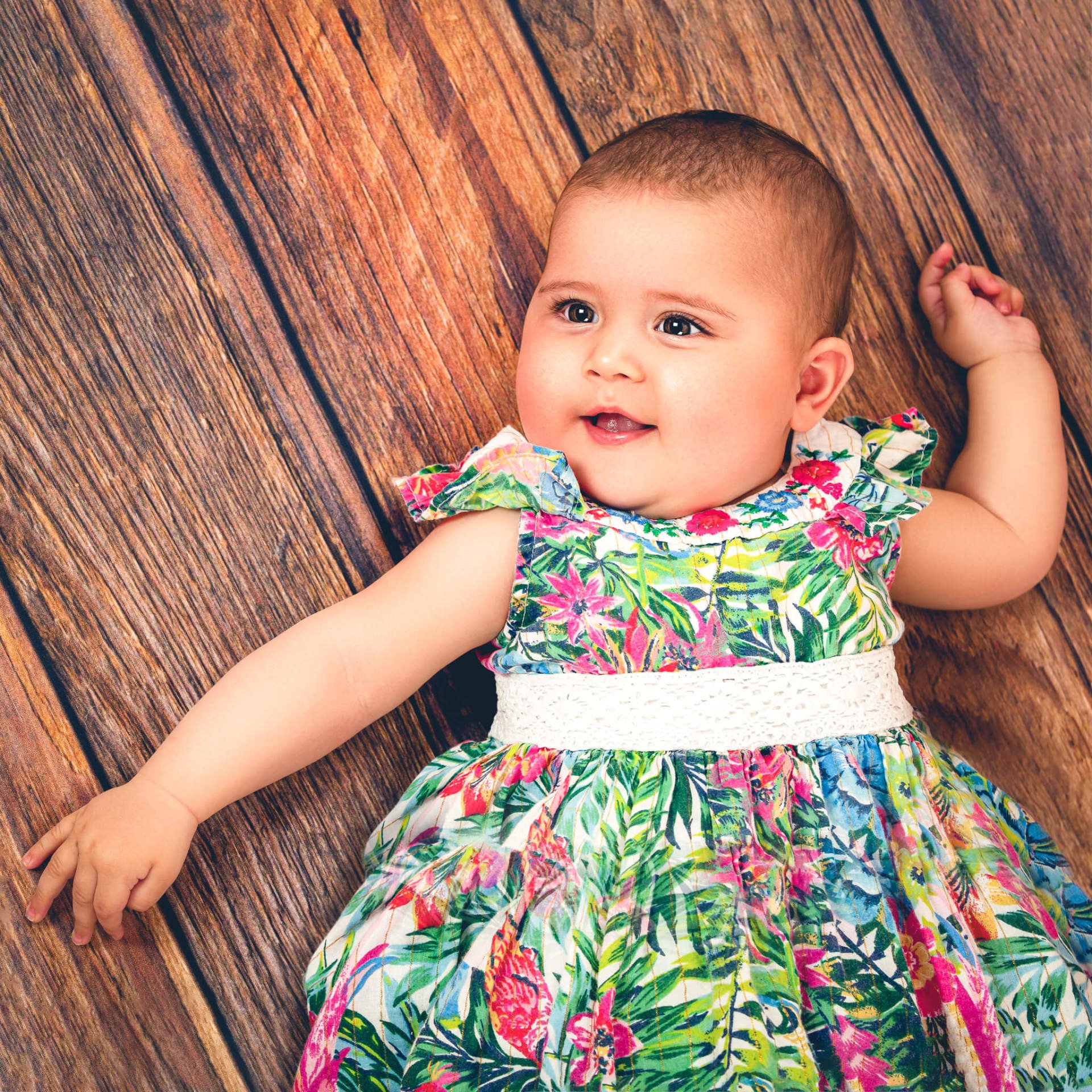}}\hspace{0.01cm}
    \subfloat[\scriptsize{\checkmark}]{\includegraphics[width=1.6cm,height=1.6cm]{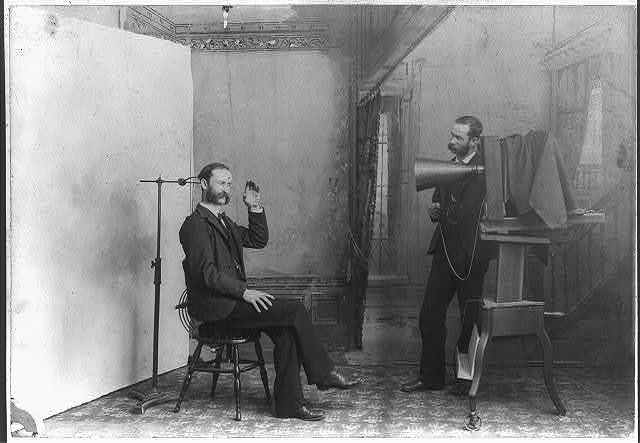}}\hspace{0.01cm}
    \subfloat[\scriptsize{\checkmark}]{\includegraphics[width=1.6cm,height=1.6cm]{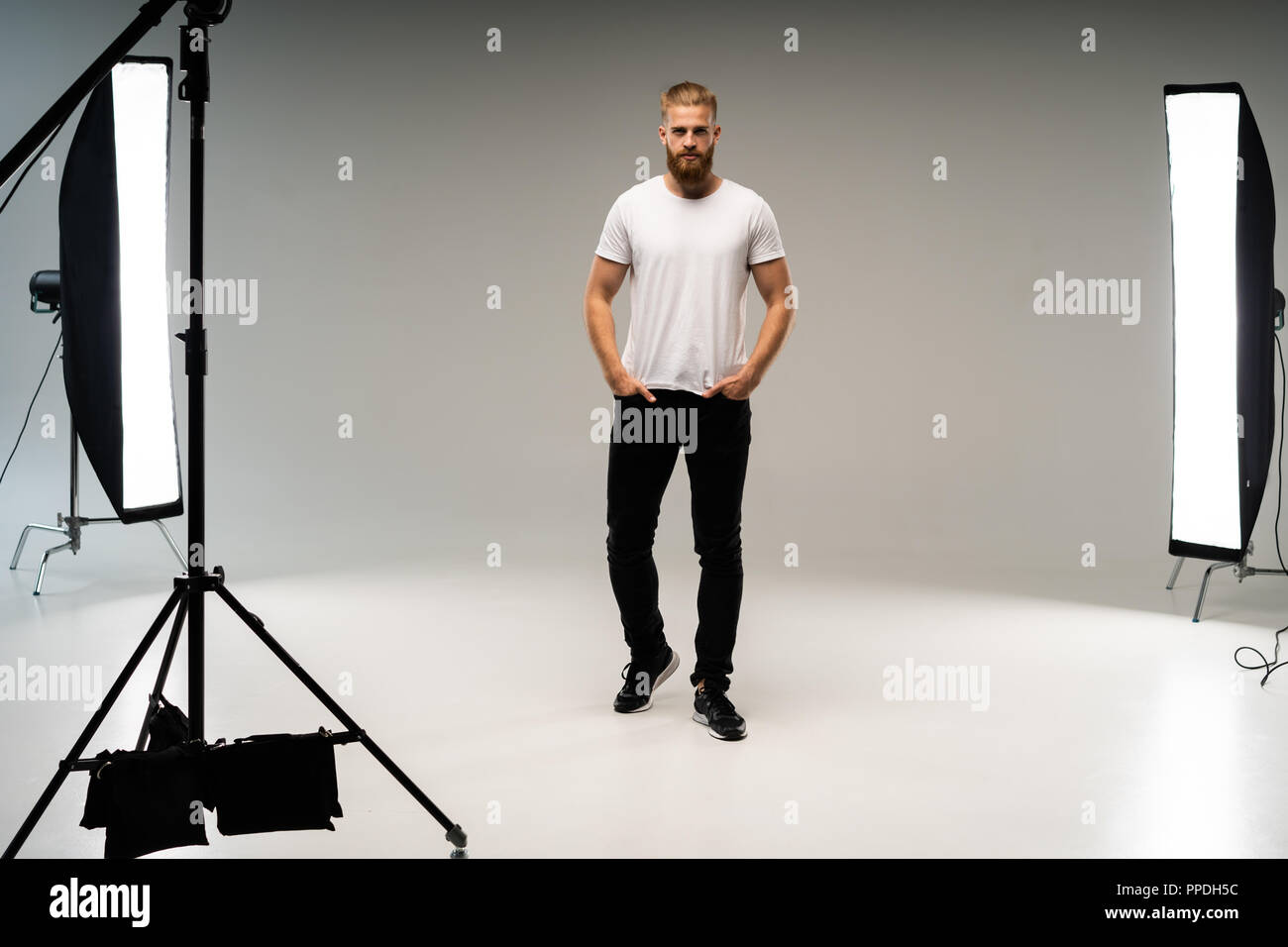}}\hspace{0.01cm}
    \subfloat[\scriptsize{dressing room}]{\includegraphics[width=1.6cm,height=1.6cm]{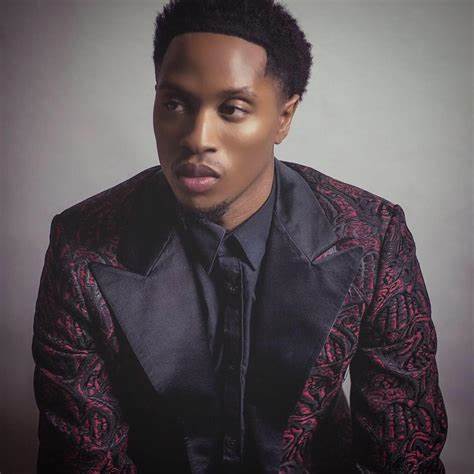}}\hspace{0.01cm}
    \subfloat[\scriptsize{bathroom}]{\includegraphics[width=1.6cm,height=1.6cm]{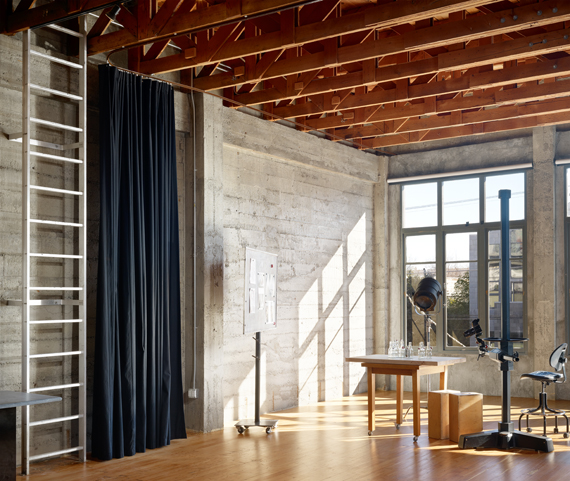}}\hspace{0.01cm}
    \subfloat[\scriptsize{\checkmark}]{\includegraphics[width=1.6cm,height=1.6cm]{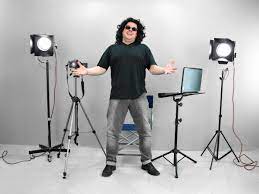}}\hspace{0.01cm}
    \subfloat[\scriptsize{\checkmark}]{\includegraphics[width=1.6cm,height=1.6cm]{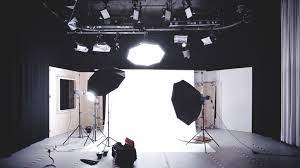}}\\ \vspace{-0.2cm}

    {\centering \rotatebox[origin=lb]{90}{\scriptsize{~swimming pool}}}
    \subfloat[\scriptsize{\checkmark}]{\includegraphics[width=1.6cm,height=1.6cm]{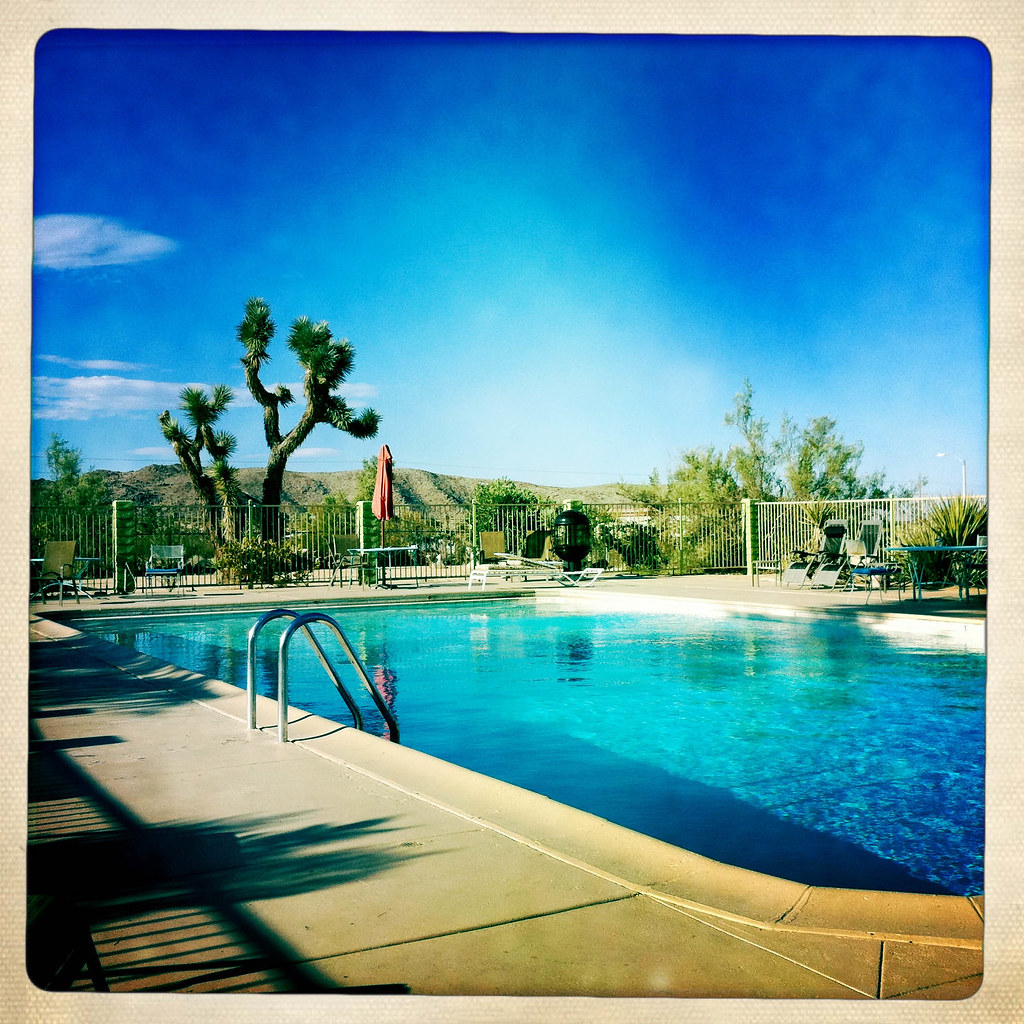}}\hspace{0.01cm}
    \subfloat[\scriptsize{bathroom}]{\includegraphics[width=1.6cm,height=1.6cm]{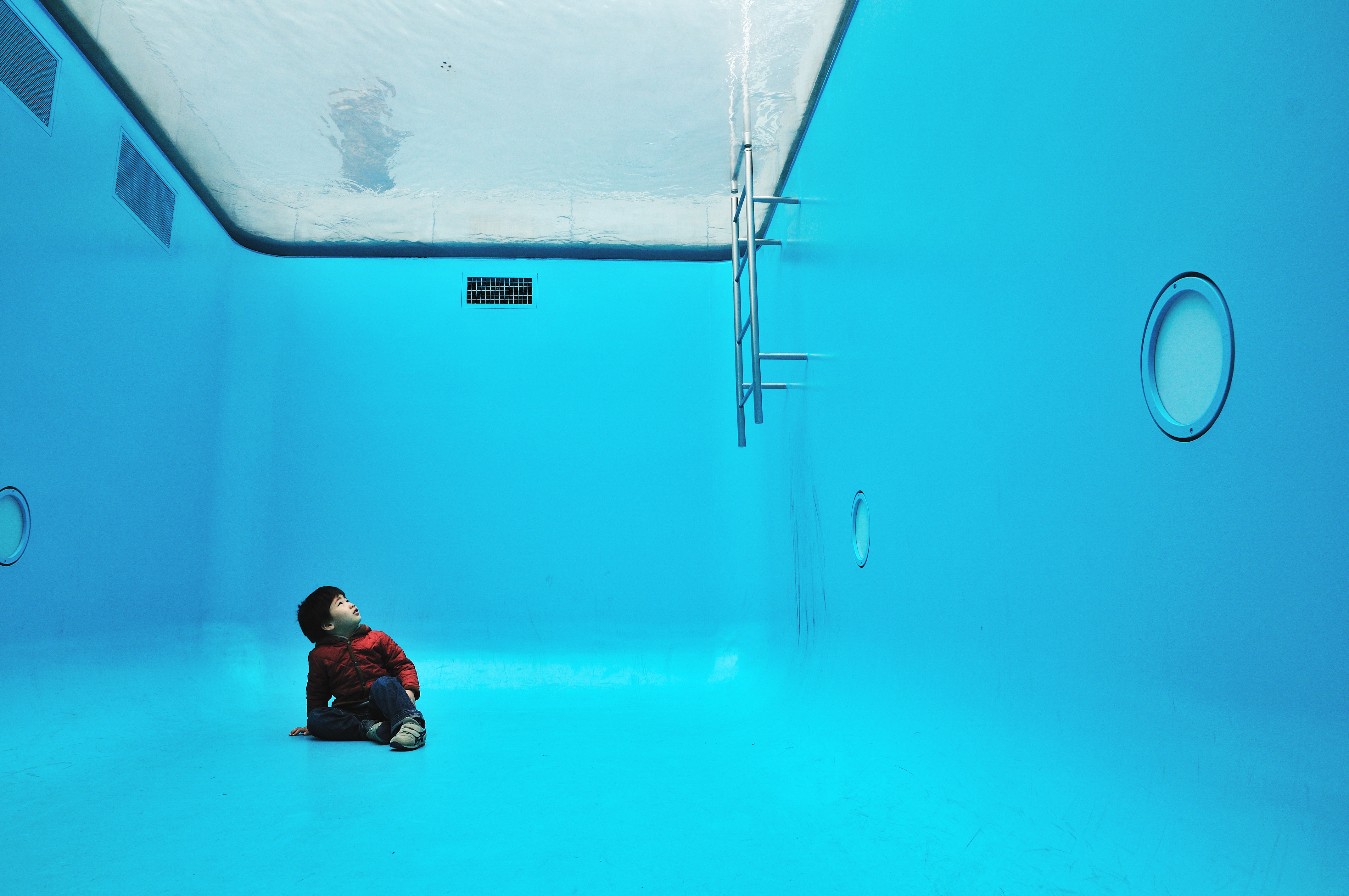}}\hspace{0.01cm}
    \subfloat[\scriptsize{\checkmark}]{\includegraphics[width=1.6cm,height=1.6cm]{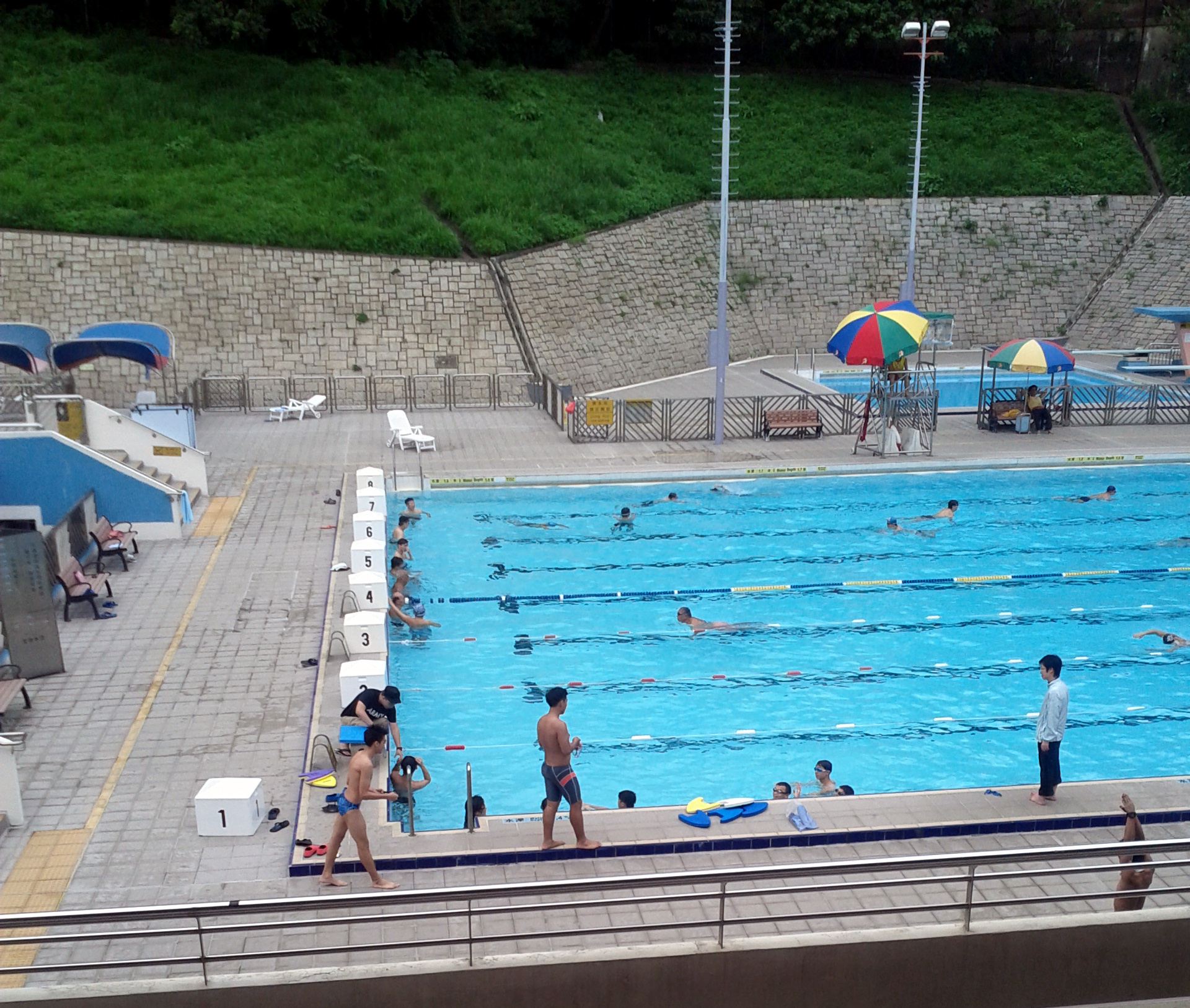}}\hspace{0.01cm}
    \subfloat[\scriptsize{\checkmark}]{\includegraphics[width=1.6cm,height=1.6cm]{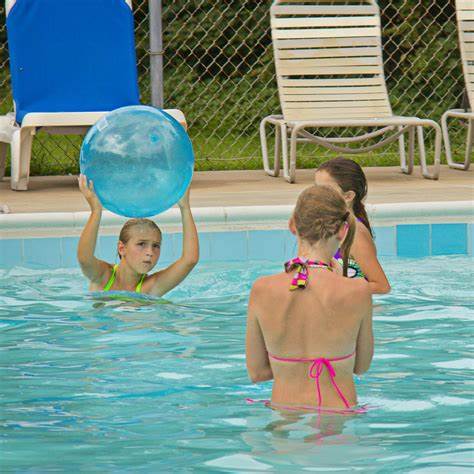}}\hspace{0.01cm}
    \subfloat[\scriptsize{\checkmark}]{\includegraphics[width=1.6cm,height=1.6cm]{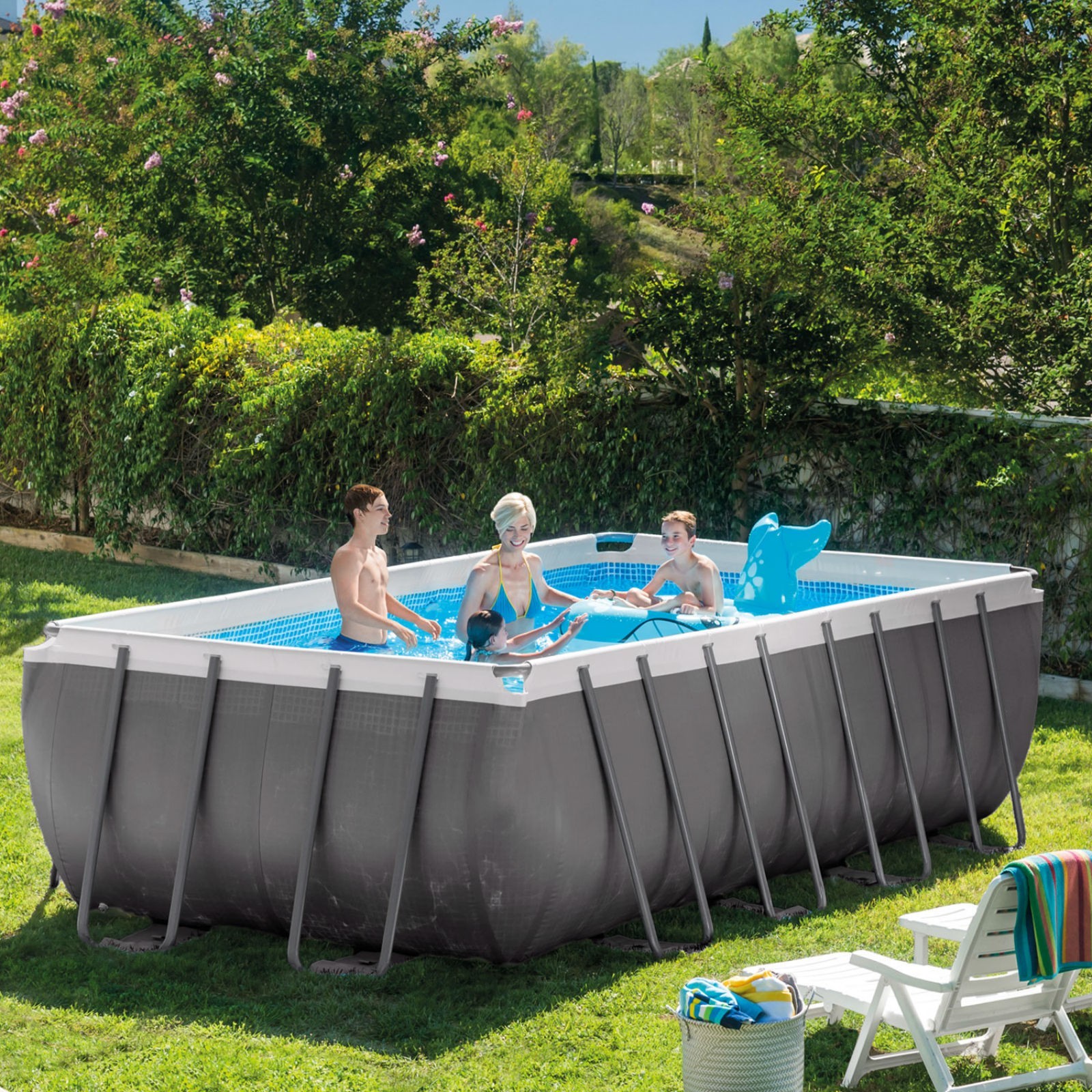}}\hspace{0.01cm}
    \subfloat[\scriptsize{\checkmark}]{\includegraphics[width=1.6cm,height=1.6cm]{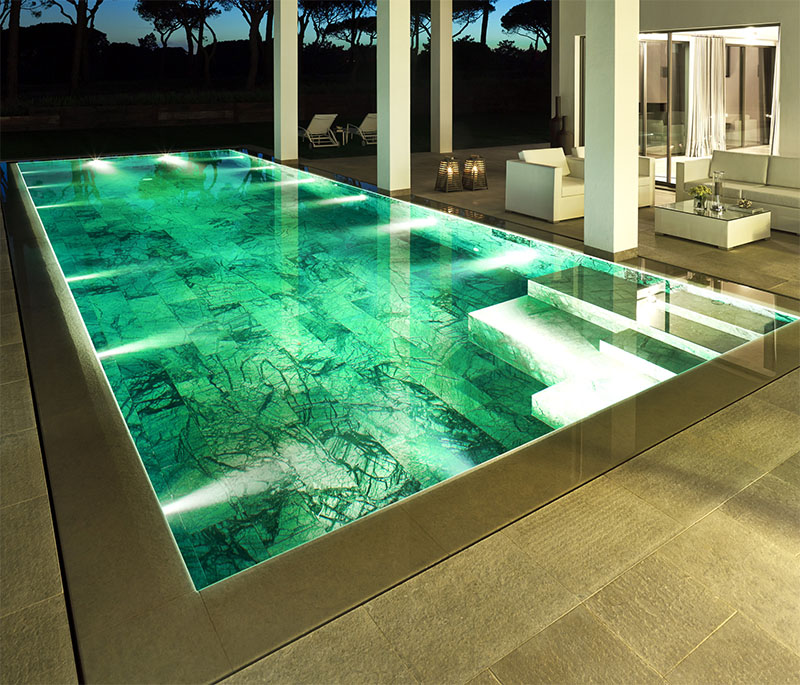}}\hspace{0.01cm}
    \subfloat[\scriptsize{\checkmark}]{\includegraphics[width=1.6cm,height=1.6cm]{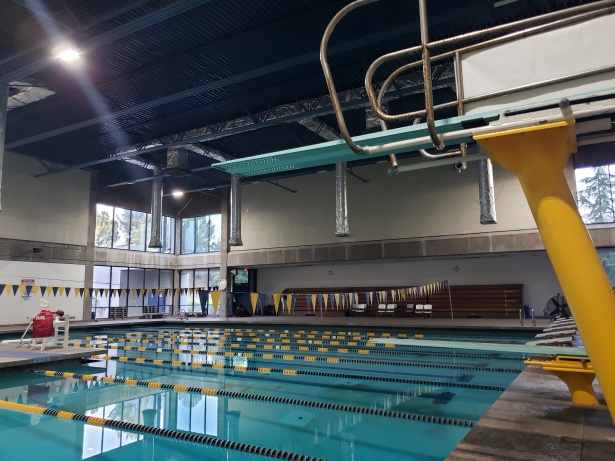}}\hspace{0.01cm}
    \subfloat[\scriptsize{\checkmark}]{\includegraphics[width=1.6cm,height=1.6cm]{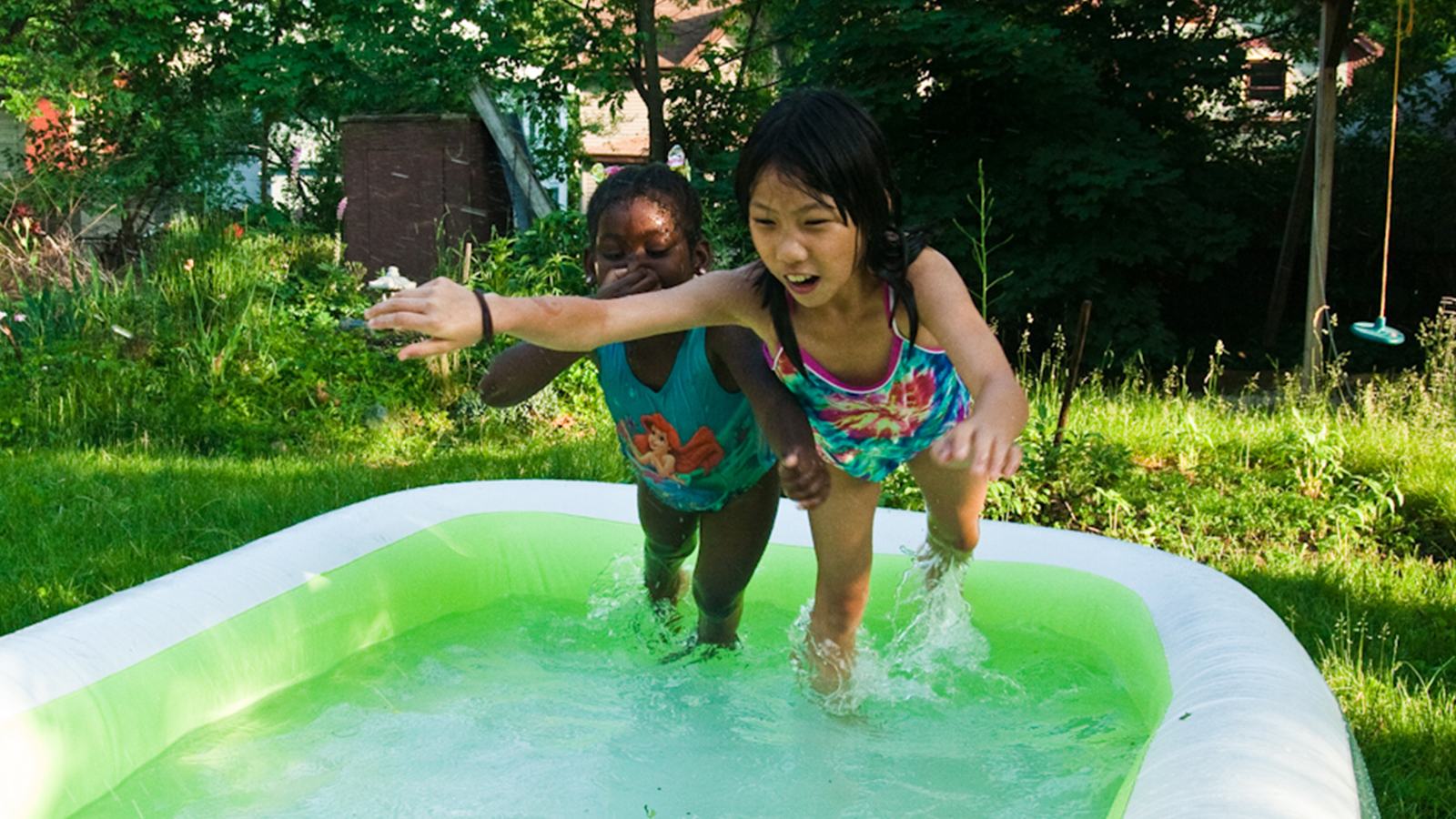}}\hspace{0.01cm}
    \subfloat[\scriptsize{\checkmark}]{\includegraphics[width=1.6cm,height=1.6cm]{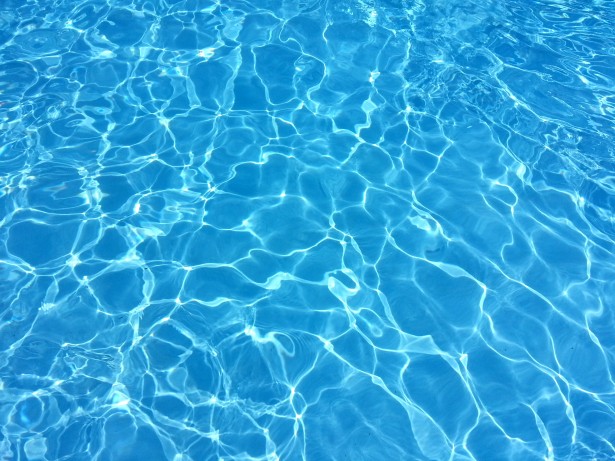}}\hspace{0.01cm}
    \subfloat[\scriptsize{\checkmark}]{\includegraphics[width=1.6cm,height=1.6cm]{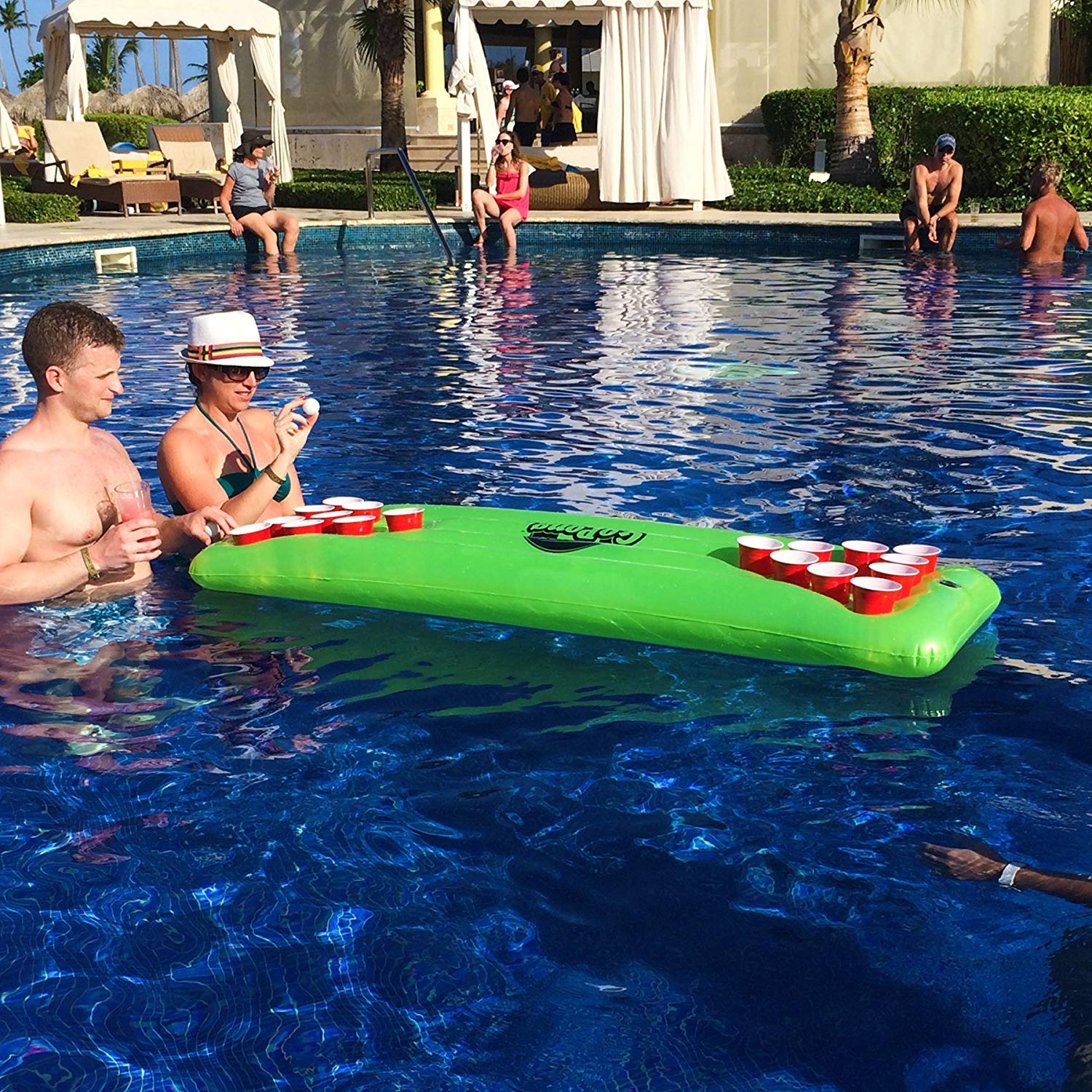}}
    \caption{Inference results on the OOD Scenes. Each row represents one label, and the name on the left is the true label for each image. A check mark (\checkmark) highlights if the prediction matches the true label; otherwise, the predicted label is placed below the image.}
    \label{fig:litmus-results}
\end{figure*}

\section{Results on OOD Data and CSAI}
\label{sec:Results}
Finally, in this section, we test our validated model --- fine-tuned with the full training set of Places8 --- on our two evaluation datasets. First, we consider the newly introduced OOD Scenes (Section~\ref{sec:dataset}), where images were selected to represent a sample distribution distinct from the Places dataset. 

Through our partnership with law enforcement agents, we test how our Indoor Scene Classification model performs on real CSAI data. This experiment on CSAI was executed in a GPU workstation owned by and located on the partner agency. No training is performed with either set.

\subsection{OOD Scenes}
\label{sec:custom-dataset-results}

The model inferred on 80 images, with 62 correct and 18~incorrect predictions, yielding 77.5\% accuracy. However, our most significant interest lies in how this result exposes the biases and successes of our model, so we take an individual look through the response for each class.

Given the small size of the dataset, the entire classification result is included in Fig.~\ref{fig:litmus-results}. On initial analysis, ``classroom'' and ``bathroom'' have the best overall performance, with all instances correctly classified (with ``classroom'' in particular showing no false positives). We hypothesize that this is due to these environments being very well defined through their object dispositions, such as the rows of student desks or the presence of mirrors and sinks in a constrained space.

On the other hand, the model struggles with classifying ``child's room'' correctly, often misjudging the scenes as ``bedroom''. While the natural overlap between both classes can explain these misclassifications, we hypothesize that the ``child's room'' false positives with ``living room'' and ``studio'' are due to contrasting colors between objects and the presence of children in the images. These particular biases will show up prominently when we consider the test on~CSAI.

Considering now the ``studio'' class, mapped from the ``television studio'' Places365 category on the training set, the model seems to adapt to other kinds of studios, with the presence of lights or homogeneous backgrounds likely influencing the result. It does not, however, always identify ``studio'' when the sample depicts instead a photo taken \textit{at a studio}, without cameras or lights being shown. This latter kind of studio picture is common in CSAI, per conversations with partner agents, and will present a problem in the later~evaluation.

Beyond that, ``swimming pool'' is still correctly classified when depicted outdoors. The water texture is a tell-tale sign of a pool that is easily captured by deep representations. It is still classified correctly even when combined with unseen traits such as the sky. The only exception is when the picture is taken from inside the pool, as shown in the second image in the ``swimming pool'' row in Fig.~\ref{fig:litmus-results}.

These results are then a ``preview'' of what we expect when working with CSAI, which is also out-of-distribution. We dive into those results then in the next section.

\subsection{CSAI}

These experiments were conducted through our partnership with expert Brazilian Federal Police agents; our validated model weights and inference code were sent to the agents, who proceeded to run and collect metrics.

Because Indoor Scene Classification is a new task for CSAI data, the agents first selected 615 random samples from their allotted-for-research dataset. These samples were then annotated with scene labels. The randomness of the process showed that while a significant amount of the images were taken indoors (392), some were outdoors (118) or did not depict enough of the environment to allow for a label (105). The random selection showing most photos taken indoors corroborates the need for research into indoor scenes for CSAI. We follow with the 392 indoor scenes for our experiments. 

\begin{figure}[h]
    \centering
    \includegraphics[width=0.45\textwidth]{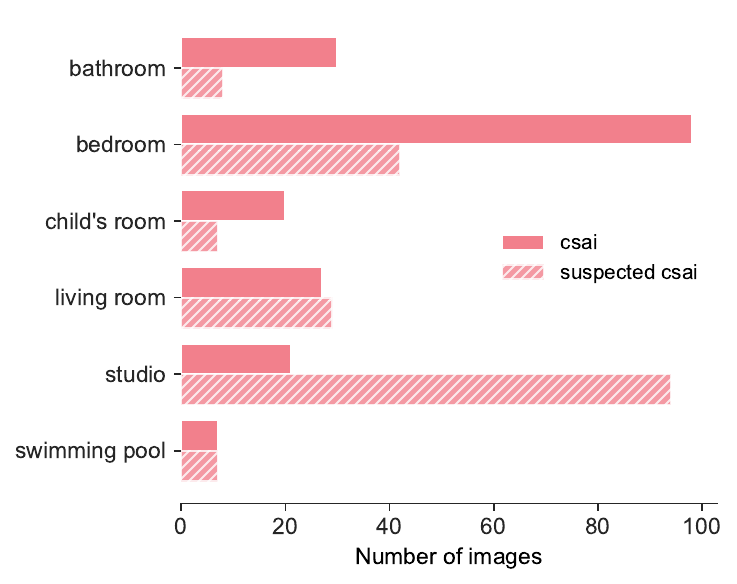}
    \caption{Histogram of ground-truth labeled scenes within CSAI and Suspected CSAI categories.}
    \label{fig:csam-hist}
\end{figure}

Moreover, this subset is grouped into two categories of child abuse material: CSAI and Suspected CSAI. The CSAI category comprises the most serious and sensitive cases, while Suspected CSAI does not show any nudity yet contains hints of eroticism. Among these, 313 images are from the CSAI category and 302 from the Suspected CSAI category. We show the distribution of samples between classes and between these two CSAI categories in Fig.~\ref{fig:csam-hist}. Classes have different frequencies per CSAI category, with ``studio'' being especially prevalent in Suspected CSAI (the agent affirms there is a correlation between this kind of scene and erotic posing). In contrast, the scenes of ``classroom'' and ``dressing room'' are not seen in any image. While this is partly due to the random selection process, it hints at a lower prevalence of these scenes, at least in the analyzed data. 

Overall, the model achieved a balanced accuracy of 36.7\% among the scene classes in Places8. This result is much lower than seen in the other data distributions and readily demonstrates the difficulties in working with CSAI. The balanced accuracies for CSAI and Suspected CSAI categories were 40.0\% and 34.1\%, respectively, showing a balanced performance despite the differences between the two groups.

\begin{figure}[t]
    \centering
    \includegraphics[width=0.45\textwidth]{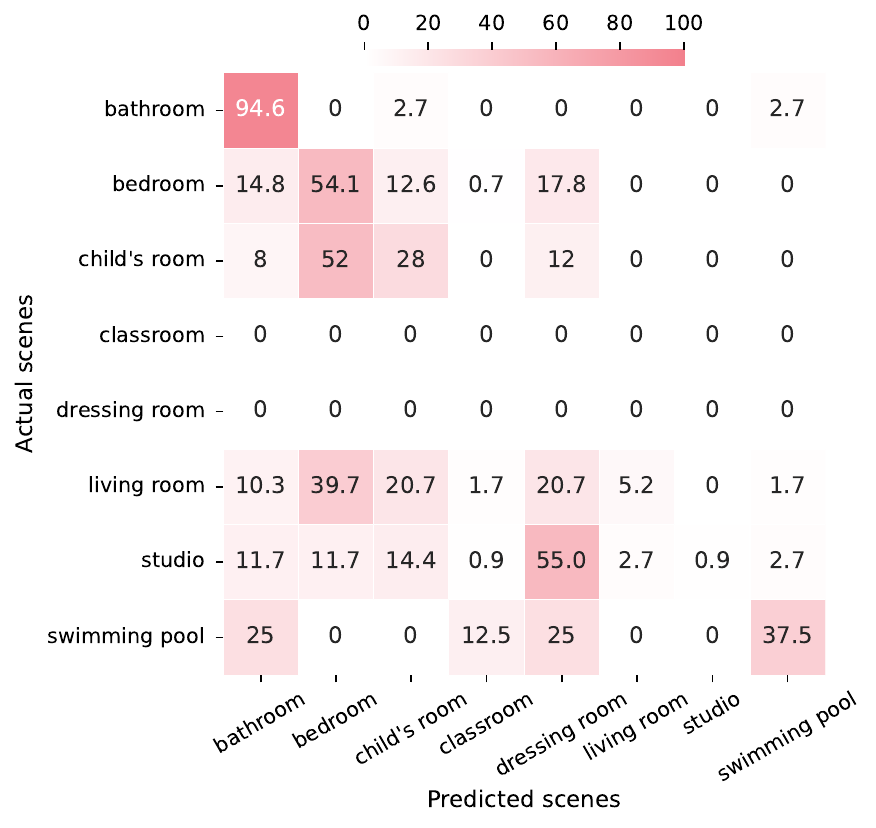}
    \caption{Confusion matrix (\%) for Places8 scenes classified in the CSAI dataset with values normalized by the number of elements in each class. ``Classroom'' and ``dressing room'' scenes are not present in this dataset.}   
    \label{fig:csam-cm}
\end{figure}

Moreover, to understand the model's weaknesses, the confusion matrix in Fig.~\ref{fig:csam-cm} shows that the prediction biases previously discussed with OOD Scenes are more consequential here. ``Child's room'', ``bedroom'' and ``living room'' are hard for the model to distinguish, with the latter confusion being more pronounced within this data. We hypothesize that the presence of people in all samples (there is no CSAI without people) may skew the model towards specific classes, with one observable effect being the distribution of samples from all classes predicted as ``dressing room'', which is one of the few classes that often include people in the training data. 

Ultimately, this test was essential to understand the model's limitations in real-world conditions, with samples that do not resemble any of the training classes' distribution of expected objects and positioning. We have found the misclassifications as ``dressing room'' to be particularly enlightening as they show a clear need to consider the presence of people as one of the known aspects of CSAI photographs in future work. 

\section{Discussion}
\label{sec:Discussion}
We have shown that combining publicly available self-supervision models with large scene classification datasets for a pretext task helps indoor classification on scenes prevalent in CSAI (but including synthetic data does not). However, the evaluation with real CSAI has highlighted that these datasets do not represent scenes recurrent in these sensitive images. Further work could analyze if adding people to the non-CSAI images affects the overall performance, as it could further introduce biases the model may leverage to drive classification.

In particular, the ambiguity among classes such as ``bedroom'', ``child's room'', and ``living room'' seems to be hard for the model to distinguish, particularly in CSAI, an ambiguity corroborated by the presence of correlation shift \cite{Ye2022OoDBenchQuantifyingUnderstanding} between the Places and CSAI data distributions. The shift is evident in a few examples: the model correlates the presence of people in the ``dressing room'' class and the presence of children in ``child's room'', but in CSAI, there are people in all images (by definition), and most of those people are children regardless of the place depicted.

This work applied self-supervision to indoor classification to aid CSAI triage and recognition, showing that SSL can leverage scene images to perform better than supervised learning. SSL performance was compared in multiple conditions, using real and synthetic images and public and sensitive data. The inference in real CSAI is vital, exposing the real distribution of scenes in this data and how a model trained with publicly available data performs. The results obtained should serve as a starting point for further investigations.

\section{Challenges and Future Work}
\label{sec:future-work}

The current solution can classify at most two of the scene classes with acceptable performance, which can be used for filtering specifically for bathrooms and bedrooms but not much else. Our analysis is, of course, limited to Indoor Scenes, one aspect of CSAI, and we know a general solution must combine this with other aspects such as object and body part detection and people; the --- expected --- appearance of the latter was shown to particularly influence the performance of a model trained on samples with few people depicted. Given our findings regarding the importance of scenery to CSAI, corroborated during the data labeling by the agents, we hope our first step to be an incentive for more research to be done within the Indoor Scene Classification field.

Additionally, all experiments used ResNet-50, which is fairly popular in SSL research, but multiple other architectures have gained much research interest after their sustained accomplishments on various tasks, such as the larger RegNets \cite{goyal2021self,radosavovic2020designing} and Vision Transformers \cite{dosovitskiy2020vit, dino, mocov3}. Also, ResNet variations have been tailored for scene classification and can perform better than its general version \cite{zeng2021scenesurvey}. 

Finally, CSAI is a challenging research topic on both experimental and psychological levels. However, society must confront this essential subject and find intelligent solutions to stop its spread, speeding up police work and keeping more children~safe.

\section*{Acknowledgments}
This work is partially funded by FAPESP~\text{2023/12086-9}, and the Serrapilheira Institute R-2011-37776. L.~S.~F.~Ribeiro is also funded by FAPESP~\text{2022/14690-8}, S.~Avila is also funded by FAPESP~\text{2020/09838-0}, \text{2013/08293-7}, H.IAAC~01245.003479/2024-1, and CNPq~316489/2023-9.
\balance

\bibliographystyle{IEEEtran}
\bibliography{mybib}
\end{document}